\documentclass{article}

\usepackage{microtype}
\usepackage{graphicx}
\usepackage{booktabs} 

\usepackage{hyperref}

\usepackage[accepted]{icml2024}

\usepackage{url}
\usepackage{amsmath}
\usepackage{amssymb}
\usepackage{mathtools}
\usepackage{amsthm}
\usepackage[linewidth=0.5pt]{mdframed}
\usepackage{caption}
\usepackage{subcaption}
\usepackage{enumitem}
\usepackage[nottoc]{tocbibind}
\usepackage{minitoc}
\usepackage{bbm}
\usepackage{algorithm}
\usepackage{algorithmic}

\usepackage[capitalize,noabbrev]{cleveref}

\theoremstyle{plain}

\theoremstyle{definition}

\theoremstyle{remark}

\usepackage[textsize=tiny]{todonotes}

\newcommand{\mytitle}{Program Machine Policy: Addressing Long-Horizon Tasks by Integrating Program Synthesis and State Machines}

\newcommand{\myrunningtitle}{
Program Machine Policy
}

\newcommand{\sun}[1]{{\color{blue}{\small\bf\sf [Sun: #1]}}}
\newcommand{\guanting}[1]{{\color{orange}{\small\bf\sf [GT: #1]}}}

\newcommand{\Skip}[1]{}

\newcommand{\method}[1]{POMP}

\newcommand{\ie}{\textit{i}.\textit{e}.,\ }
\newcommand{\eg}{\textit{e}.\textit{g}.,\ }

\newcommand{\myfig}[1]{Figure \ref{#1}}
\newcommand{\mytable}[1]{Table \ref{#1}}

\newcommand{\mysecref}[1]{Section \ref{#1}}
\newcommand{\myalgo}[1]{Algorithm \ref{#1}}

\newcommand{\dotieconcat}[2]{
  \text{\raisebox{.8ex}{$\smallfrown$}}%
}

\makeatletter
\newcommand\dslfontsize{\@setfontsize\dslfontsize\@viiipt\@viiiipt}
\makeatother

\newcommand{\myparagraph}[1]{\noindent \textbf{#1.}}
\newcommand{\vspacesection}[1]{\vspace{0.04cm}
\section{#1}
\vspace{0.04cm}}
\newcommand{\vspacesubsection}[1]{\vspace{0.04cm}
\subsection{#1}
\vspace{0.04cm}}
\newcommand{\vspacesubsubsection}[1]{\vspace{0.04cm}
\subsubsection{#1}
\vspace{0.04cm}}

\newcommand\SmallCaption[1]{%
  \captionsetup{font=normalsize}%
  \caption{#1}}

\usepackage{color}
\usepackage{xcolor}
\definecolor{codegray}{rgb}{0.5,0.5,0.5}

\usepackage{listings}

\lstdefinelanguage{DSL}{
  sensitive = true,
  keywords={DEF, WHILE, IF},
  otherkeywords={
    >, <, ==
  },
  keywords = [2]{noMarkersPresent, leftIsClear, rightIsClear, frontIsClear, markersPresent},
  keywords = [3]{move, turnLeft, turnRight, pickMarker, putMarker},
  numbersep=8pt,
  tabsize=1,
  showstringspaces=false,
  breaklines=true,
  frame=top,
  comment=[l]{//},
  morecomment=[s]{/*}{*/},
  commentstyle=\color{purple}\ttfamily,
  stringstyle=\color{red}\ttfamily,
  belowskip=0.5pt,
  aboveskip=0.2pt,
  morestring=[b]',
  morestring=[b]",
  keywordstyle=\color{blue},
  keywordstyle=[2]\color{red},
  keywordstyle=[3]\color{codegray},
  numberstyle=\tiny\color{codegray},
  basicstyle=\ttfamily\tiny,
  }
\lstloadaspects{formats}
\lstdefinestyle{numbers}
{numbers=left, stepnumber=1, numberstyle=\tiny, numbersep=10pt, numberstyle=\tiny\color{codegray}}

\icmltitlerunning{\myrunningtitle}

\begin{document}

\twocolumn[

\icmltitle{\mytitle}



\icmlsetsymbol{equal}{*}

\begin{icmlauthorlist}

\icmlauthor{Yu-An Lin}{equal,ntu}
\icmlauthor{Chen-Tao Lee}{equal,ntu}
\icmlauthor{Guan-Ting Liu}{equal,ntu}
\icmlauthor{Pu-Jen Cheng}{ntu}
\icmlauthor{Shao-Hua Sun}{ntu}
\end{icmlauthorlist}

\icmlaffiliation{ntu}{National Taiwan University, Taiwan}

\icmlcorrespondingauthor{Shao-Hua Sun}{shaohuas@ntu.edu.tw}

\icmlkeywords{Machine Learning, ICML}

\vskip 0.3in
]



\printAffiliationsAndNotice{\icmlEqualContribution} 

\begin{abstract}
Deep reinforcement learning (deep RL) excels in various domains but lacks generalizability and interpretability. 
On the other hand, programmatic RL methods~\citep{trivedi2021learning, liu2023hierarchical} reformulate RL tasks as synthesizing interpretable programs that can be executed in the environments. 
Despite encouraging results, these methods are limited to short-horizon tasks. On the other hand, representing RL policies using state machines~\citep{Inala2020SynthesizingInductGen} can inductively generalize to long-horizon tasks; however, it struggles to scale up to acquire diverse and complex behaviors.
This work proposes the Program Machine Policy (POMP), which bridges the advantages of programmatic RL and state machine policies, allowing for the representation of complex behaviors and the address of long-term tasks. Specifically, we introduce a method that can retrieve a set of effective, diverse, and compatible programs. Then, we use these programs as modes of a state machine and learn a transition function to transition among mode programs, allowing for capturing repetitive behaviors. Our proposed framework outperforms programmatic RL and deep RL baselines on various tasks and demonstrates the ability to inductively generalize to even longer horizons without any fine-tuning. Ablation studies justify the effectiveness of our proposed search algorithm for retrieving a set of programs as modes.

\end{abstract}

\vspacesection{Introduction}
\label{sec:intro}

Deep reinforcement learning (deep RL) has recently achieved tremendous success in various domains,
such as controlling robots~\citep{gu2017deep, ibarz2021train}, playing strategy board games~\citep{silver2016mastering, silver2017mastering}, and mastering video games~\citep{vinyals2019grandmaster, wurman2022outracing}. 
However, the black-box neural network policies learned by deep RL methods are not human-interpretable, posing challenges in scrutinizing model decisions and establishing user trust~\citep{lipton2018mythos, shen2020}. 
Moreover, deep RL policies often suffer from overfitting and struggle to generalize to novel scenarios~\citep{zhang2018study, cobbe2019quantifying}, limiting their applicability in the context of most real-world applications.

To address these issues,~\citet{trivedi2021learning} and~\citet{liu2023hierarchical} explored representing policies as programs, which details task-solving procedures in a formal programming language.
Such \textit{program policies} are human-readable and demonstrate significantly improved zero-shot generalizability from smaller state spaces to larger ones.
Despite encouraging results, these methods are limited to synthesizing concise programs (\ie shorter than $120$ tokens) that can only tackle short-horizon tasks (\ie less than $400$ time steps)~\citep{liu2023hierarchical}.


To solve tasks requiring generalizing to longer horizons,~\citet{Inala2020SynthesizingInductGen} proposed representing a policy using a state machine.
By learning to transfer between modes encapsulating actions corresponding to specific states, such \textit{state machine policies} can model repetitive behaviors and inductively generalize to tasks with longer horizons.
Yet, this approach is constrained by highly simplified, task-dependent grammar that can only structure constants or proportional controls as action functions.
Additionally, its teacher-student training scheme requires model-based trajectory optimization with an accurate environment model, which can often be challenging to attain in practice~\citep{polydoros2017survey}.


This work aims to bridge the best of both worlds --- 
the interpretability and scalability of \textit{program policies}
and the inductive generalizability of \textit{state machine policies}.
We propose the \textbf{P}r\textbf{o}gram \textbf{M}achine \textbf{P}olicy (\method{}),
which learns a state machine upon a set of diverse programs.
Intuitively, every mode (the inner state) of \method{} is a high-level skill described by a program instead of a single primitive action.
By transitioning between these mode programs, \method{} can reuse these skills to tackle long-horizon tasks with an arbitrary number of repeating subroutines. 

We propose a three-stage framework to learn such a Program Machine Policy. 
(1) \textbf{Constructing a program embedding space}:
To establish a program embedding space that smoothly, continuously parameterizes programs with diverse behaviors, we adopt the method proposed by~\citet{trivedi2021learning}.
(2) \textbf{Retrieving a diverse set of effective and reusable programs}: Then, we introduce a searching algorithm to retrieve a set of programs from the learned program embedding space. 
Each program can be executed in the MDP and achieve satisfactory performance; more importantly, these programs are compatible and can be sequentially executed in any order.
(3) \textbf{Learning the transition function}:
To alter between a set of programs as state machine modes, the transition function takes the current environment state and the current mode (\ie program) as input and predicts the next mode.
We propose to learn this transition function using RL via maximizing the task rewards from the MDP.

To evaluate our proposed framework POMP, we adopt the Karel domain~\citep{pattis1981karel}, which characterizes an agent that navigates a grid world and interacts with objects. 
POMP outperforms programmatic reinforcement learning and deep RL baselines on existing benchmarks proposed by~\citet{trivedi2021learning, liu2023hierarchical}.
We design a new set of tasks with long-horizon on which POMP demonstrates superior performance and the ability to generalize to even longer horizons without fine-tuning inductively.
Ablation studies justify the effectiveness of our proposed search algorithm for retrieving a set of programs as modes.

\vspacesection{Related Work}
\label{sec:related}

\myparagraph{Program Synthesis}
Program synthesis techniques revolve around program generation to convert given inputs into desired outputs. These methods have demonstrated notable successes across diverse domains such as array and tensor manipulation~\citep{balog2016deepcoder, ellis2020dreamcoder} and 
string transformation~\citep{devlin2017robustfill, hong2020latent, zhong2023hierarchical}. 
Most program synthesis methods focus on task specifications such as input/output pairs
or language descriptions; in contrast, this work aims to synthesize human-readable programs as policies to solve reinforcement learning tasks.

\myparagraph{Programmatic Reinforcement Learning}
Programmatic reinforcement learning methods~\citep{choi2005learning, distill, liu2023hierarchical} explore structured representations for representing RL policies, including decision trees~\citep{bastani2018verifiable}, state machines~\citep{Inala2020SynthesizingInductGen}, symbolic expressions~\citep{landajuela21a}, and programs~\citep{verma2018programmatically, verma2019imitation, aleixo2023show}. 
\citet{trivedi2021learning, liu2023hierarchical} attempted to produce policies described by domain-specific language programs to solve simple RL tasks.
We aim to take a step toward addressing complex, long-horizon, repetitive tasks. 

\myparagraph{State Machines for Reinforcement Learning}
Recent works adopt state machines to 
model rewards~\citep{ICML18_Icarte_UsingRM, NEURIPS19_Icarte_LearningRM, FurelosBlanco2022HierarchiesRM, xu2022JointInfRM},
or achieve inductive generalization~\citep{Inala2020SynthesizingInductGen}.
Prior works explore using symbolic programs~\citep{Inala2020SynthesizingInductGen} or 
neural networks~\citep{ICML18_Icarte_UsingRM, NEURIPS19_Icarte_LearningRM, Hasanbeig2021DeepSynthAS}
as modes (\ie states) in state machines.
On the other hand, our goal is to exploit human-readable programs for each mode of state machines so that the resulting state machine policies are more easily interpreted.

\myparagraph{Hierarchical Reinforcement Learning}
Drawing inspiration from hierarchical reinforcement learning (HRL) frameworks~\citep{barto2003recent, vezhnevets2017feudal, lee2019composing}, our Program Machine Policy shares the HRL philosophy by treating the transition function as a "high-level" policy and mode programs as "low-level" policies or skills. This aligns with the option framework~\cite{sutton1999between, bacon2017option, klissarov2021flexible}, using interpretable options as sub-policies. Diverging from option frameworks, our method retrieves a set of mode programs first and then learns a transition function to switch between modes or terminate. Further related work discussions are available in \mysecref{app:extend_related_work}.

\vspacesection{Problem Formulation}
\label{sec:problem}

Our goal is to devise a framework that can produce a Program Machine Policy (POMP),
a state machine whose modes are programs structured in a domain-specific language,
to address complex, long-horizon tasks described by Markov Decision Processes (MDPs).
To this end, 
we first synthesize a set of task-solving, diverse, compatible programs as modes,
and then learn a transition function to alter between modes.


\noindent \textbf{Domain Specific Language.} 
This work adopts the domain-specific language (DSL) of the Karel domain~\cite{bunel2018leveraging, chen2019executionguided, trivedi2021learning}, as illustrated in \myfig{fig:dsl}. 
This DSL describes the control flows as well as the perception and actions of the Karel agent.
Actions including \texttt{move}, \texttt{turnRight}, and \texttt{putMarker} define how the agent can interact with the environment. 
Perceptions, such as \texttt{frontIsClear} and \texttt{markerPresent}, formulate how the agent observes the environment.
Control flows, \eg \texttt{if}, \texttt{else}, \texttt{while}, enable representing divergent and repetitive behaviors. 
Furthermore, Boolean and logical operators like \texttt{and}, \texttt{or}, and \texttt{not} allow for composing more intricate conditions. 
This work uses programs structured in this DSL to construct the modes of a Program Machine Policy.

\begin{figure}
\centering
\begin{mdframed}[font=\scriptsize]
\vspace{-0.4cm}
{
    \begin{align*}
    \text{Program}\ \rho &\coloneqq \text{DEF}\  \text{run}\ \text{m}(\ s\ \text{m})\\
    \text{Repetition} \ n &\coloneqq \text{Number of repetitions}\\
    \text{Perception} \ h & \coloneqq \text{frontIsClear} \ | \ \text{leftIsClear} \ | \  \text{rightIsClear} \ | \ \\ 
    & \text{markerPresent} \ | \ \text{noMarkerPresent} \\
    \text{Condition} \ b &\coloneqq \text{perception h} \ | \ \text{not} \ \text{perception h} \\
    \text{Action} \ a &\coloneqq \text{move} \ | \ \text{turnLeft}
    \ | \ \text{turnRight} \ | \ \\
    & \text{putMarker} \ | \ \text{pickMarker} \\
    \text{Statement}\ s &\coloneqq \text{while}\ \text{c}(\ b\ \text{c})\ \text{w}(\ s\ \text{w}) \ | \ s_1 ; s_2 \ | \ a \ | \\ 
    & \ \text{repeat}\ \text{R=}n\ \text{r}(\ s\ \text{r}) \ | \ \text{if}\ \text{c}(\ b\ \text{c})\ \text{i}(\ s\ \text{i}) \ | \\ 
    & \ \text{ifelse}\ \text{c}(\ b\ \text{c})\ \text{i}(\ s_1\ \text{i}) \  \text{else}\ \text{e}(\ s_2\ \text{e}) \\
    \end{align*}
\vspace{-0.9cm}
}
\end{mdframed}
    \caption[]{
        \small
        \textbf{Karel Domain-Specific Language (DSL)}, designed for describing the Karel agent's behaviors. 
        \label{fig:dsl}
    }
\end{figure}

\noindent \textbf{Markov Decision Process (MDP).} 
The tasks considered in this work can be formulated as finite-horizon discounted Markov Decision Processes (MDPs).
The performance of a Program Machine Policy is evaluated based on the execution traces of a series of programs selected by the state machine transition function.
The rollout of a program $\rho$ consists of a $T$-step sequence of state-action pairs $\{(s_t , a_t)\}_{t=1,\text{ ..., }T}$ obtained from a program executor $\text{EXEC}(\cdot)$ that executes program $\rho$ to interact with an environment, 
resulting in the discounted return $\sum_{t=0}^{T} \gamma^t (r_t)$, where $r_t = \mathcal{R}(s_t, a_t)$ denotes the reward function.
We aim to maximize the total rewards by executing a series of programs following the state machine transition function.


\noindent \textbf{Program Machine Policy (POMP).} 
This work proposes a novel RL policy representation, a Program Machine Policy, which
consists of a finite set of \textit{modes} $M = \{m_k\}_{k=1, \text{ ...,} |M|}$ as internal states of the state machine and a state machine \textit{transition function} $f$ that determines how to transition among these modes. 
Each \textit{mode} $m_i$ encapsulates a human-readable program $\rho_{m_i}$ that will be executed when this mode is selected during policy execution. 
On the other hand, the \textit{transition function} $f_{\theta_{p,q}}(m_p, m_q, s)$ outputs the probability of transitioning from mode $m_p$ to mode $m_q$ given the current MDP state $s$.
To rollout a POMP, the state machine starts at initial mode $m_{\text{init}}$, which will not execute any action, and transits to the next mode $m_{i+1}$ based on the current mode $m_i$ and MDP state $s$. 
If $m_{i+1}$ equals the termination mode $m_{\text{term}}$, the Program Machine Policy will terminate and finish the rollout. Otherwise, the mode program $\rho_{m_{i+1}}$ will be executed and generates state-action pairs $\{(s^{i+1}_t , a^{i+1}_t)\}_{t=1,\text{ ..., }T^{i+1}}$ before the state machine transits to the next state $m_{i+2}$. 


\vspacesection{Approach}
\label{sec:approach}
\begin{figure*}[ht]
\centering

    \includegraphics[trim=0 0 25 0,clip,width=1.0\textwidth]{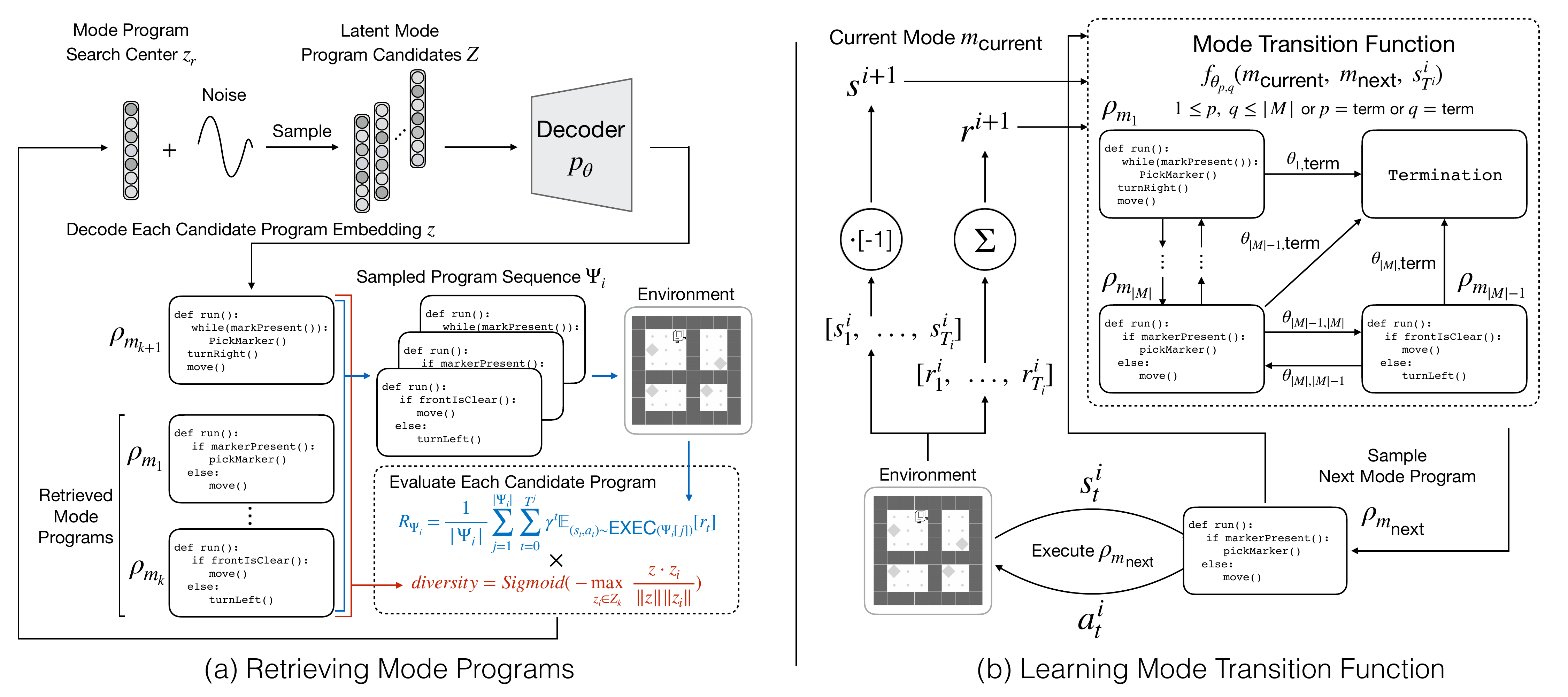}

    \caption[]{
        \small 
        \textbf{Learning Program Machine Policy.}
        \textbf{(a): Retrieving mode programs.}
        After learning the program embedding space, we propose an advanced search scheme built upon the Cross-Entropy Method (CEM) to search programs $\rho_{m_1}, ..., \rho_{m_k}, \rho_{m_{k+1}}$ of different skills. While searching for the next mode program $\rho_{m_{k+1}}$, we consider its compatibility with previously determined mode programs $\rho_{m_1}, ..., \rho_{m_k}$ by randomly sampling a sequence of mode programs. We also consider the diversity among all mode programs using the \textit{diversity multiplier}.
        \textbf{(b): Learning the mode transition function.}
        Given the current environment state $s$ and the current mode $m_\text{current}$, the mode transition function predicts the transition probability over each mode of the state machine with the aim of maximizing the total accumulative reward from the environment.

        
        \label{fig:model}
    }
\end{figure*}

We design a three-stage framework to produce 
a Program Machine Policy that can be executed and maximize the return given a task described by an MDP.
First, constructing a program embedding space that smoothly, continuously parameterizes programs with diverse behaviors is introduced in~\mysecref{sec:approach_stage1}.
Then, \mysecref{sec:approach_stage2} presents a method that retrieves a set of effective, diverse, and compatible programs as POMP modes.
Given retrieved modes, \mysecref{sec:approach_stage3} describes learning the transition function determining transition probability among the modes.
An overview of the proposed framework is illustrated in \myfig{fig:model}.

\vspacesubsection{Constructing Program Embedding Space}
\label{sec:approach_stage1}
We follow the approach and the program dataset presented by \citet{trivedi2021learning} to learn a program embedding space that smoothly and continuously parameterizes programs with diverse behaviors.
The training objectives include a VAE loss and two losses that encourage learning a behaviorally smooth program embedding space.
Once trained, we can use the learned decoder $p_\theta$ to map any program embedding $z$ to a program $\rho_z=p_\theta(z)$ consisting of a sequence of program tokens.
Details about the program dataset generation and the encoder-decoder training can be found in \mysecref{app:POMP_encoder_decoder_training}.
 

\vspacesubsection{Retrieving Mode Programs}
\label{sec:approach_stage2}

With a program embedding space,
we aim to retrieve a set of programs as modes of a Program Machine Policy given a task.
This set of programs should satisfy the following properties.

\begin{itemize}[leftmargin=3mm]
    \item \textbf{Effective}: Each program can solve the task to some extent (\ie obtain some task rewards).
    \item \textbf{Diverse}: The more behaviorally diverse the programs are, the richer behaviors can be captured by the Program Machine Policy.
    \item \textbf{Compatible}: Sequentially executing some programs with specific orders can potentially lead to improved task performance.
\end{itemize}


\vspacesubsubsection{Retrieving Effective Programs}
\label{sec:cem}
To obtain a task-solving program, 
we can apply the Cross-Entropy Method~\citep[CEM;][]{rubinstein1997optimization}, iteratively searching in a learned program embedding space~\citep{trivedi2021learning} as described below:

\begin{enumerate}
  \item[(1)] Randomly initialize a program embedding vector ${z}_r$ as the search center.
  \item[(2)] Add random noises to ${z}_r$ to generate a population of program embeddings $Z=\{z_i\}_{i=1,\text{...},n}$, where $n$ denotes the population size.
  \item[(3)] Evaluate every program embedding $z \in Z$ with the evaluation function $G$ to get a list of fitness score $[G(z_i)]_{i=1,\text{...},n}$.
  \item[(4)] Average the top k program embeddings in $Z$ according to fitness scores $[G(z_i)]_{i=1,\text{...},n}$ and assign it to the search center ${z}_r$.
  \item[(5)] Repeat (2) to (4) until the fitness score $G(z_r)$ of ${z}_r$ converges or the maximum number of steps is reached.
\end{enumerate}


Since we aim to retrieve a set of effective programs,
we can define the evaluation function as 
the program execution return of a decoded program embedding, \ie $G(z) = \sum_{t=0}^{T} \gamma^t \mathbb{E}_{(s_t,a_t) \sim \text{EXEC} (\rho_{z})}[r_{t}]$.
To retrieve a set of $|M|$ programs as the modes of a Program Machine Policy, we can 
run this CEM search $N$ times, take $|M|$ best program embeddings, and obtain the decoded program set $\{\rho_{z_{r_i}} = p_\theta(z_{r_i})\}_{i=1,\text{...},|M|}$.
\mysecref{app:cem} presents more details and the CEM search pseudocode.


\vspacesubsubsection{Retrieving Effective, Diverse Programs}
\label{sec:cem_d}
We will use the set of retrieved programs as the modes of a Program Machine Policy.
Hence, a program set with diverse behaviors can lead to a Program Machine Policy representing complex, rich behavior.
However, the program set obtained by running the CEM search for $|M|$ times can have low diversity, preventing the policy from solving tasks requiring various skills.

To address this issue, we propose the \textit{diversity multiplier} that considers previous search results to encourage diversity among the retrieved programs. The evaluation of mode programs employing the \textit{diversity multiplier} is illustrated in~\myfig{fig:model}.
Specifically, during the $(k+1)$st CEM search, each program embedding $z$ is evaluated by $G(z,Z_k) = (\sum_{t=0}^{T} \gamma^t \mathbb{E}_{(s_t,a_t) \sim \text{EXEC} (\rho_{z})}[r_{t}]) \cdot diversity(z, Z_k)$, where $diversity(z, Z_k)$ is the proposed \textit{diversity multiplier} defined as $Sigmoid(-\max_{{z}_i \in {Z}_k}\ \frac{{z} \cdot {z}_i}{\Vert {z} \Vert \Vert {z}_i \Vert})$. 
Thus, the program execution return is scaled down by $diversity(z,Z_k)$ based on the maximum cosine similarity between $z$ and the retrieved program embeddings $Z_k = \{z_i\}_{i=1,\text{...},k}$ from the previous $k$ CEM searches.
This diversity multiplier encourages searching for program embeddings different from previously retrieved programs.

To retrieve a set of $|M|$ programs as the modes of a Program Machine Policy, we can run this CEM+diversity search $N$ times, take $|M|$ best program embeddings, and obtain the decoded program set.
The procedure and the search trajectory visualization can be found in~\mysecref{app:cem_div}.

\vspacesubsubsection{Retrieving Effective, Diverse, Compatible Programs}
\label{sec:cem_d_c}

Our Program Machine Policy executes a sequence of programs by learning a transition function to select from mode programs.
Therefore, these programs need to be compatible with each other, \ie executing a program following the execution of other programs can improve task performance.
Yet, CEM+diversity discussed in~\mysecref{sec:cem_d} searches every program independently. 

In order to account for the compatibility among programs during the search, 
we propose a method, CEM+diversity+compatibility.
When evaluating the program embedding $z$, we take the decoded program $\rho_z$ as the $(k+1)$st mode. 
Then, lists of programs $\Psi_{i, i=1,\text{...}, D}$ are sampled with replacements from determined $k$ modes and the $(k+1)$st mode. Each program list $\Psi_i$ contains at least one $(k+1)$st mode program to consider the compatibility between the $(k+1)$st and previously determined $k$ modes.
We compute the return by sequentially executing these $D$ lists of programs and multiply the result with the \textit{diversity multiplier} proposed in \mysecref{sec:cem_d}. The resulting evaluation function is 
$G({z}, Z_k) = \frac{1}{D} \sum_{i=1}^{D} R_{\Psi_i} \cdot diversity({z}, Z_k)$, where $R_{\Psi_i}$ is the normalized reward obtained from executing all programs in the program list $\Psi_i$ and can be written as follows:

\begin{equation}
\label{eq:reward_rho_G}
R_{\Psi_i} = \frac{1}{|\Psi_i|} \sum_{j=1}^{|\Psi_i|} \sum_{t=0}^{T^j} \gamma^t \mathbb{E}_{(s_t,a_t) \sim \text{EXEC} (\Psi_{i}[j])}[r_t]
\end{equation}

where $|\Psi_i|$ is the number of programs in the program list $\Psi_i$, $\Psi_i[j]$ is the $j$-th program in the program list $\Psi_i$, and $\gamma$ is the discount factor.

We can run this search for $|M|$ times to obtain a set of programs that are effective, diverse, and compatible with each other, 
which can be used as mode programs for a Program Machine Policy. More details and the whole search procedure can be found in~\mysecref{app:cem_div_comp}.

\vspacesubsection{Learning Transition Function}
\label{sec:approach_stage3}

Given a set of modes (\ie programs) $M=\{m_k\}_{k=1, \text{ ..., }|M|}$, we formulate learning a transition function $f$ that determines how to transition between modes as a reinforcement learning problem aiming to maximize the task return.
In practice, we define an initial mode $m_{\text{init}}$ that initializes the Program Machine Policy at the beginning of each episode; also, we define a termination mode $m_{\text{term}}$, which terminate the episode if chosen.
Specifically, the transition function $f_{\theta_{p,q}}(m_p, m_q, s)$ outputs the probability of transitioning from mode $m_p$ to mode $m_q$, given the current environment state $s$. 

At $i$-th transition function step,
given the current state $s$ and the current mode $m_{\text{current}}$,
the transition function predicts the probability of transition to $m_{\text{term}} \cup \{m_k\}_{k=1, \text{ ..., }|M|}$.
We sample a mode $m_{\text{next}}$ based on the predicted probability distribution.
If the sampled mode is the termination mode, the episode terminates;
otherwise, we execute the corresponding program $\rho$, yielding the next state (\ie the last state $s_{T^i}^i$ of the state sequence returned by $\text{EXEC}(\rho)$, where $T^i$ denotes the horizon of the $i$-th program execution)
and the cumulative reward $r^{i+1} = \sum_{t=1}^{T^i} r^i_t$.
Note that the program execution $\text{EXEC}(\rho)$ will terminate after full execution or the number of actions emitted during $\text{EXEC}(\rho)$ reaches 200.
We assign the next state to the current state and the next mode to the current mode.
Then, we start the next ($i+1$)st transition function step. 
This process stops when the termination mode is sampled, or a maximum step is reached.
Further training details can be found in~\mysecref{app:mode_transition}.

To further enhance the explainability of the transition function, we employ the approach proposed by~\citet{koul2018learning} to extract the state machine structure from the learned transition network. 
By combining the retrieved set of mode programs and the extracted state machine, our framework is capable of solving long-horizon tasks while being self-explanatory. 
Examples of extracted state machines are shown in ~\mysecref{app:fsm_et}.

\begin{figure*}
    \begin{subfigure}[b]{0.19\textwidth}
    \centering
    \includegraphics[trim=0 120 0 120, clip,height=\textwidth]{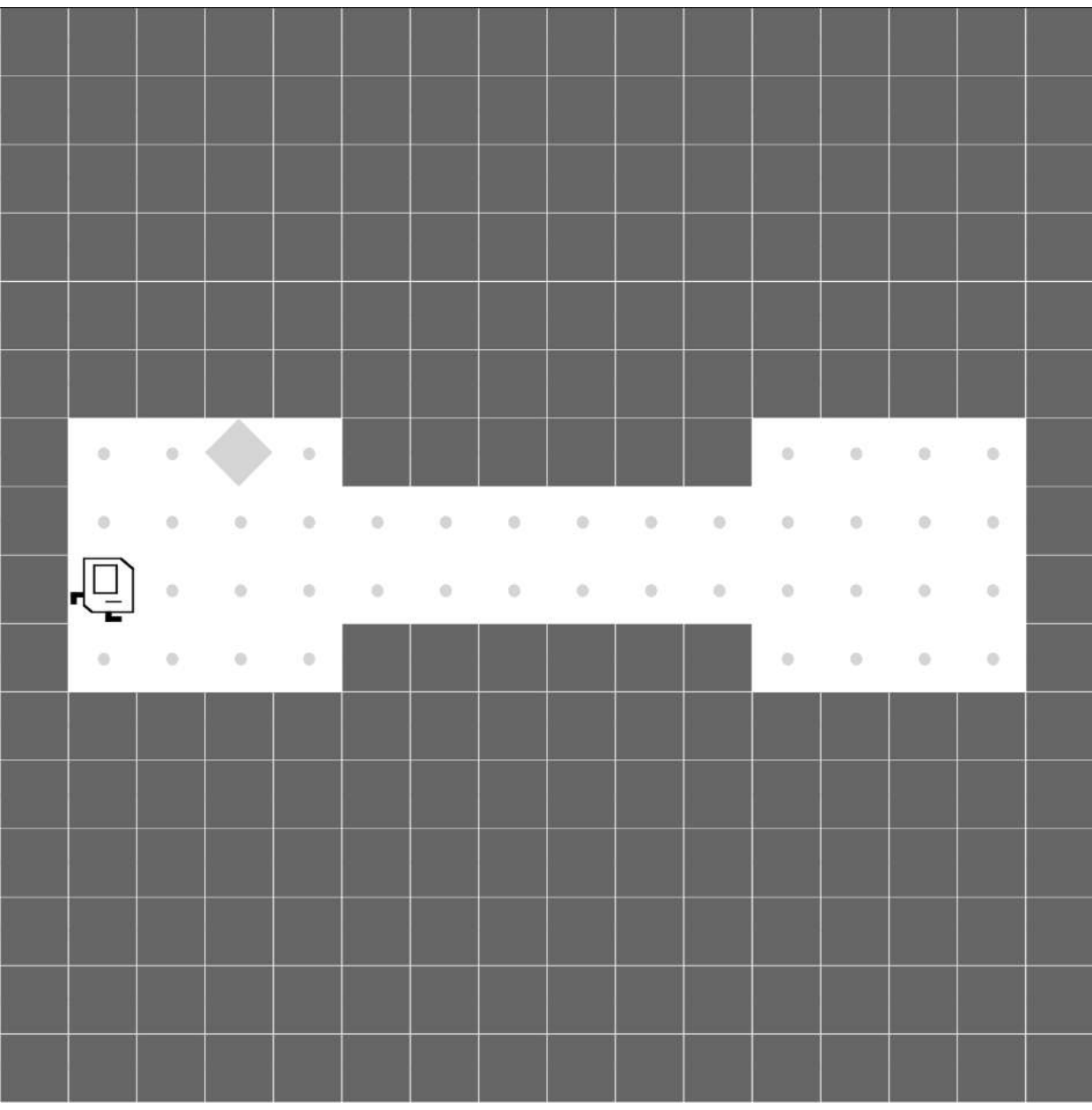}
    \SmallCaption{\scriptsize\textsc{Seesaw}}
    \end{subfigure}
    \hfill
    \begin{subfigure}[b]{0.19\textwidth}
    \centering
    \includegraphics[trim=0 120 0 120, clip,height=\textwidth]{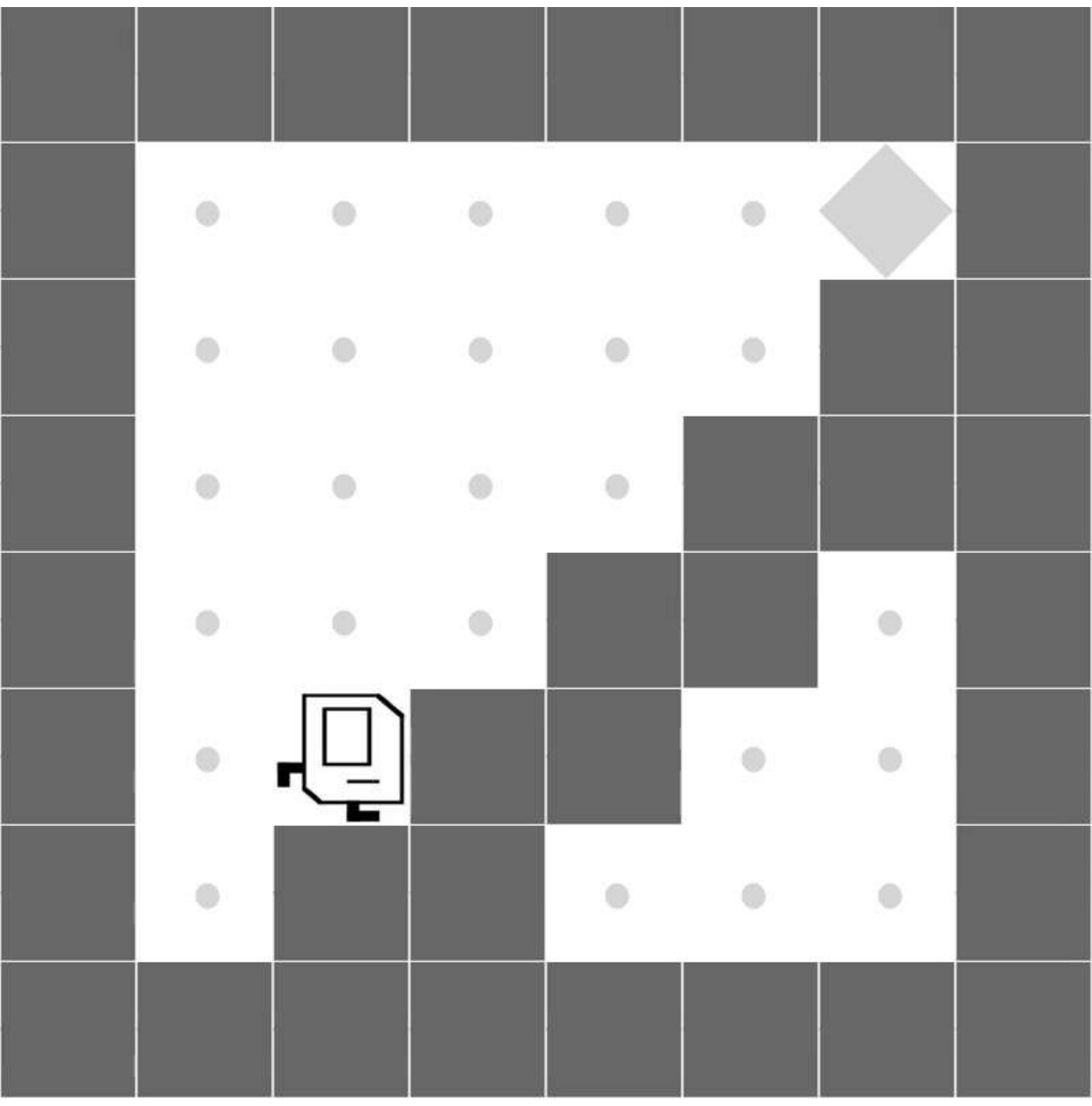}
    \SmallCaption{\scriptsize\textsc{Up-N-Down}}
    \end{subfigure}
    \hfill
    \begin{subfigure}[b]{0.19\textwidth}
    \centering
    \includegraphics[trim=0 120 0 120, clip,height=\textwidth]{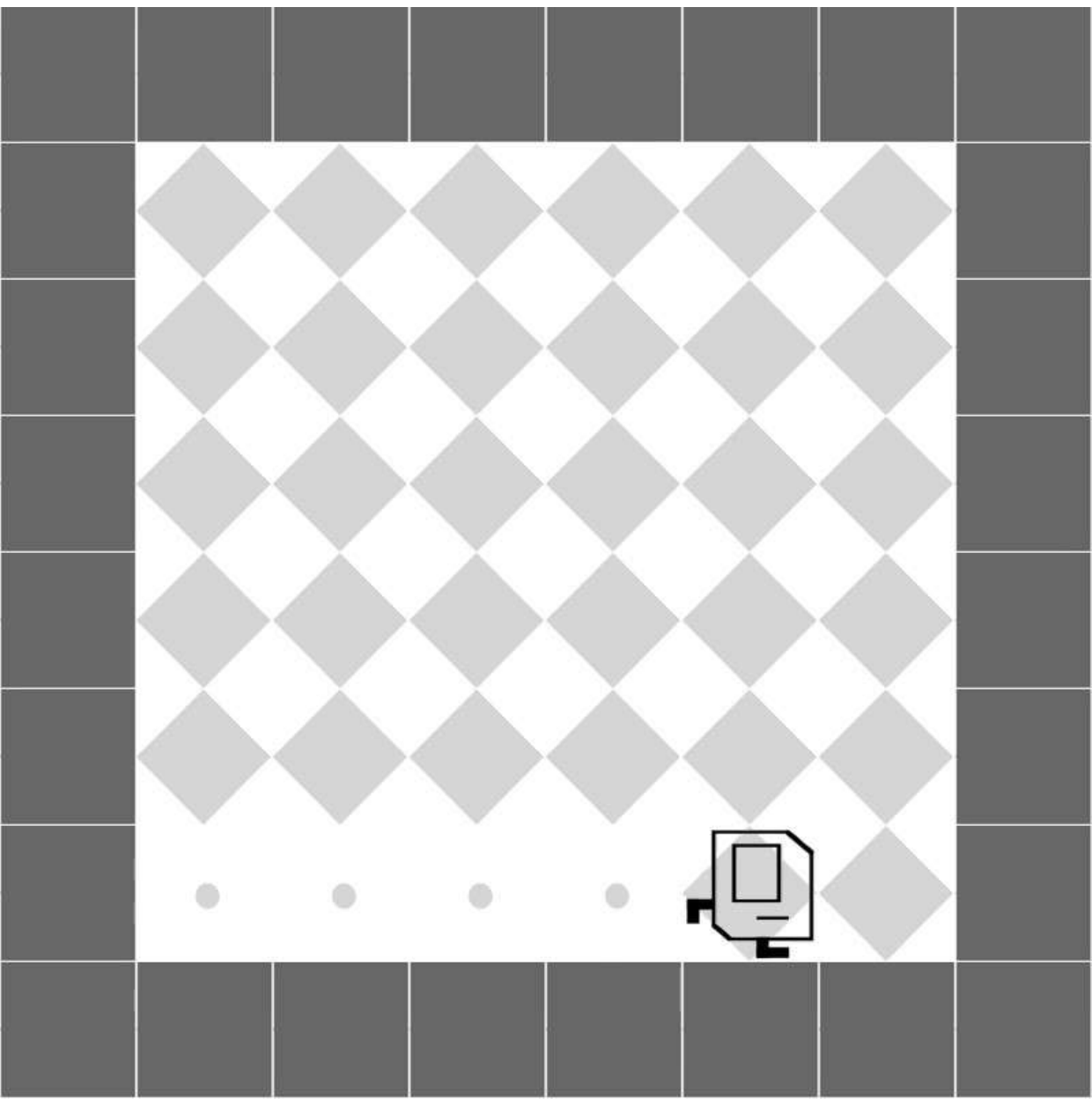}
    \SmallCaption{\scriptsize\textsc{Farmer}}
    \end{subfigure}
    \hfill
    \begin{subfigure}[b]{0.19\textwidth}
    \centering
    \includegraphics[trim=0 120 0 120, clip,height=\textwidth]{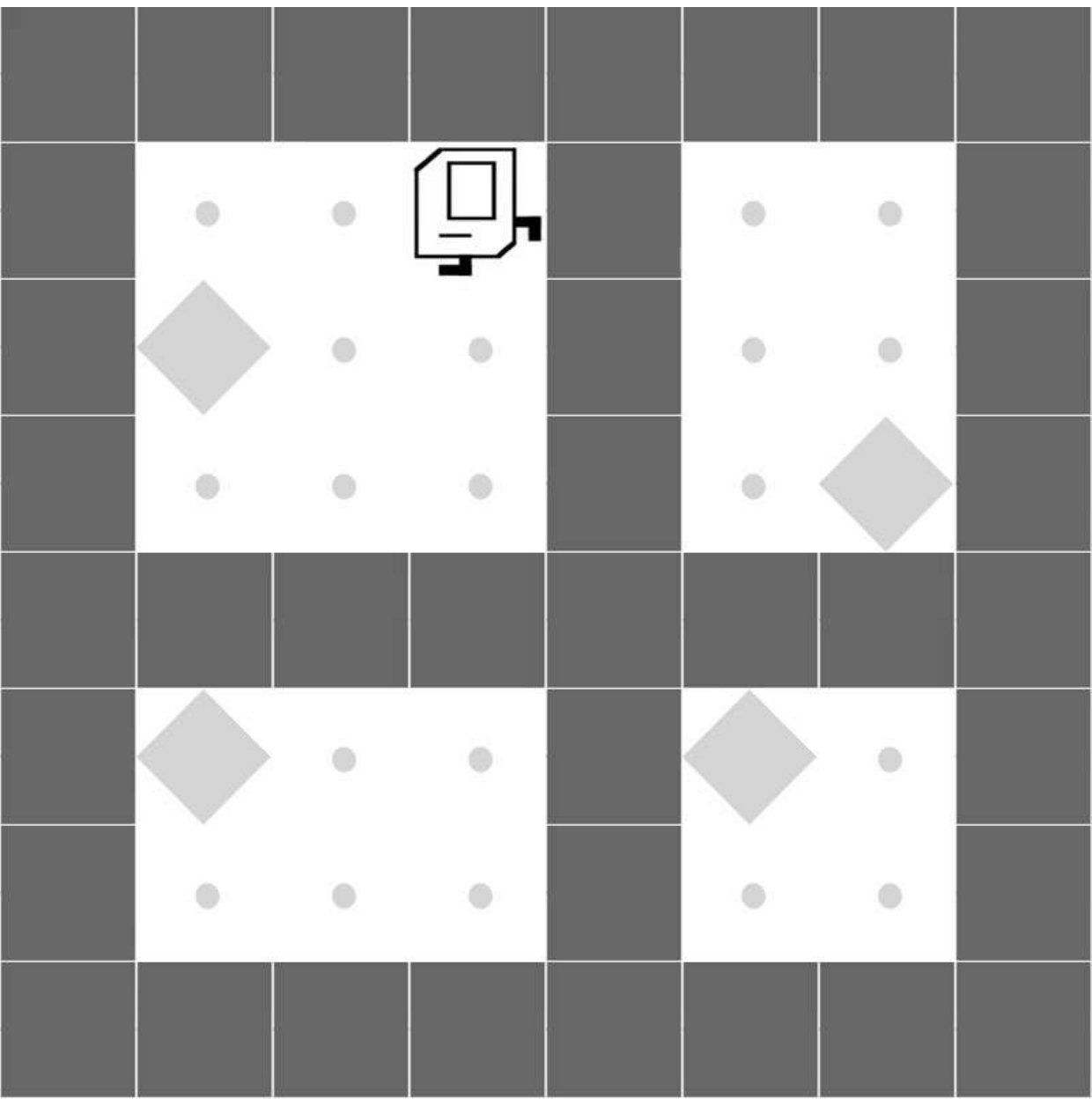}
    \SmallCaption{\scriptsize\textsc{Inf-DoorKey}}
    \end{subfigure}
    \hfill
    \begin{subfigure}[b]{0.19\textwidth}
    \centering
    \includegraphics[trim=10 140 10 130, clip,height=\textwidth]{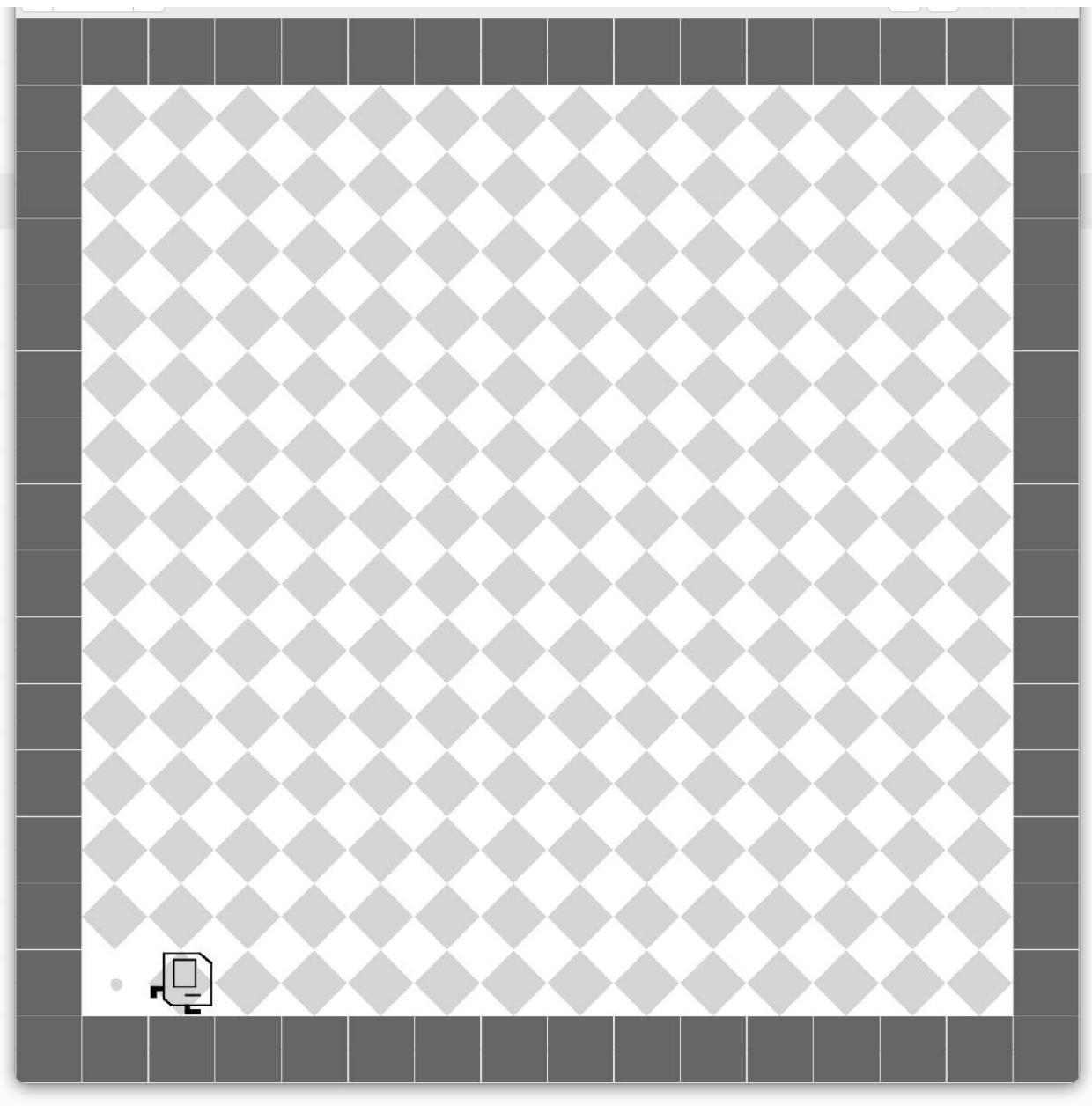}
    \SmallCaption{\scriptsize \textsc{Inf-Harvester}}
    \end{subfigure}
    \hfill
    \caption[]{
    \textbf{\textsc{Karel-Long} Problem Set}: 
    This work introduces a new set of tasks in the Karel domain.
    These tasks necessitate learning diverse, repetitive, and task-specific skills. 
    For example, in our designed \textsc{Inf-Harvester}, the agent needs to traverse the whole map and pick nearly 400 markers to solve the tasks since the environment randomly generates markers; 
    in contrast, the \textsc{Harvester} from the \textsc{Karel} problem set \citep{trivedi2021learning} can be solved by picking just 36 markers. 
    }
    \label{fig:karel_long}
\end{figure*}

\vspacesection{Experiments}
\label{sec:exp}



We aim to answer the following questions with the experiments and ablation studies.
(1) Can our proposed \textit{diversity multiplier} introduced in \mysecref{sec:cem_d} enhance CEM and yield programs with improved performance?
(2) Can our proposed CEM+diversity+compatibility introduced in \mysecref{sec:cem_d_c} retrieve a set of programs that are diverse yet compatible with each other?
(3) Can the proposed framework produce a Program Machine Policy that outperforms existing methods on long-horizon tasks?

Detailed hyperparameters for the following experiments can also be found in \mysecref{app:hyper_settings_exp}.
\vspacesubsection{Karel Problem Sets}
\label{sec:karel_problem_set}

To this end, we consider the Karel domain~\citep{pattis1981karel},
which is widely adopted in program synthesis~\citep{bunel2018leveraging,
shin2018improving, sun2018neural, chen2019executionguided} and programmatic reinforcement learning~\citep{trivedi2021learning, liu2023hierarchical}.
Specifically, we utilize 
the \textsc{Karel} problem set~\citep{trivedi2021learning} and 
the \textsc{Karel-Hard} problem set~\citep{liu2023hierarchical}.
The \textsc{Karel} problem set includes six basic tasks, each of which can be solved by a short program (less than $45$ tokens), with a horizon shorter than $200$ steps per episode.
On the other hand, the four tasks introduced in the \textsc{Karel-Hard} problem require longer, more complex programs (\ie $45$ to $120$ tokens) with longer execution horizons (\ie up to $500$ actions). 
Details about two problem sets can be found in~\mysecref{app:karel_problem_set} and~\mysecref{app:karel_hard_problem_set}.


\myparagraph{\textsc{Karel-Long} Problem Set}
Since most of the tasks in the \textsc{Karel} and \textsc{Karel-Hard} problem sets are short-horizon tasks (\ie can be finished in less than 500 timesteps), they are not suitable for evaluating long-horizon task-solving ability (\ie tasks requiring more than 3000 timesteps to finish). Hence, we introduce a newly designed \textsc{Karel-Long} problem set as a benchmark to evaluate the capability of POMP.

As illustrated in~\myfig{fig:karel_long}, 
the tasks requires the agent to fulfill extra constraints (\eg not placing multiple markers on the same spot in \textsc{Farmer}, receiving penalties imposed for not moving along stairs in \textsc{Up-N-Down}) and conduct extended exploration (\eg repetitively locating and collecting markers in \textsc{Seesaw}, \textsc{Inf-DoorKey}, and \textsc{Inf-Harvester}). More details about the \textsc{Karel-Long} tasks can be found in~\mysecref{app:karel_long_problem_set}.

\vspacesubsection{Cross-Entropy Method with Diversity Multiplier}

\begin{table*}
\centering
\caption{\textbf{Evaluation on \textsc{Karel} and \textsc{Karel-Hard} Tasks.} Mean return and standard deviation of all methods across the \textsc{Karel} and \textsc{Karel-Hard} problem set, evaluated over five random seeds. CEM+diversity outperforms CEM with significantly smaller standard deviations across 8 out of 10 tasks, highlighting the effectiveness and stability of CEM+diversity. In addition, POMP outperforms LEAPS and HPRL on eight out of ten tasks.}
\label{tab:karel_Karel_hard_POMP_main}
\resizebox{\textwidth}{!}{%
\begin{tabular}{@{}ccccccccccc@{}}
\toprule
  Method &
  \begin{tabular}[c]{@{}c@{}}\textsc{Four}\\ \textsc{Corner}\end{tabular} &
  \begin{tabular}[c]{@{}c@{}}\textsc{Top}\\ \textsc{Off}\end{tabular} &
  \begin{tabular}[c]{@{}c@{}}\textsc{Clean}\\ \textsc{House}\end{tabular} &
  \begin{tabular}[c]{@{}c@{}}\textsc{Stair}\\ \textsc{Climber}\end{tabular} &
  \textsc{Harvester} &
  \textsc{Maze} &
  \begin{tabular}[c]{@{}c@{}}\textsc{Door}\\ \textsc{Key}\end{tabular} &
  \begin{tabular}[c]{@{}c@{}}\textsc{One}\\ \textsc{Stroke}\end{tabular} &
  \textsc{Seeder} &
  \textsc{Snake} \\ 
\midrule
CEM & 0.45 $\pm$ 0.40 & 0.81 $\pm$ 0.07 & 0.18 $\pm$ 0.14 & \textbf{1.00} $\pm$ 0.00 & 0.45 $\pm$ 0.28 & \textbf{1.00} $\pm$ 0.00 & 0.50 $\pm$ 0.00 & 0.65 $\pm$ 0.19 & 0.51 $\pm$ 0.21 & 0.21 $\pm$ 0.15\\

\begin{tabular}[c]{@{}c@{}}CEM+diversity\end{tabular} & \textbf{1.00} $\pm$ 0.00 & \textbf{1.00} $\pm$ 0.00 & 0.37 $\pm$ 0.06 & \textbf{1.00} $\pm$ 0.00 & 0.80 $\pm$ 0.07 & \textbf{1.00} $\pm$ 0.00 & 0.50 $\pm$ 0.00 & 0.62 $\pm$ 0.01 & 0.69 $\pm$ 0.07 & 0.36 $\pm$ 0.02 \\ 

\midrule
DRL & 0.29 $\pm$ 0.05 & 0.32 $\pm$ 0.07 & 0.00 $\pm$ 0.00 & \textbf{1.00} $\pm$ 0.00 & 0.90 $\pm$ 0.10 & \textbf{1.00} $\pm$ 0.00 & 0.48 $\pm$ 0.03 & \textbf{0.89} $\pm$ 0.04 & 0.96 $\pm$ 0.02 & \textbf{0.67} $\pm$ 0.17 \\
LEAPS & 0.45 $\pm$ 0.40 & 0.81 $\pm$ 0.07 & 0.18 $\pm$ 0.14 & \textbf{1.00} $\pm$ 0.00 & 0.45 $\pm$ 0.28 & \textbf{1.00} $\pm$ 0.00 & 0.50 $\pm$ 0.00 & 0.65 $\pm$ 0.19 & 0.51 $\pm$ 0.21 & 0.21 $\pm$ 0.15\\
HPRL (5) & \textbf{1.00} $\pm$ 0.00 & \textbf{1.00} $\pm$ 0.00 & \textbf{1.00} $\pm$ 0.00 & \textbf{1.00} $\pm$ 0.00 & \textbf{1.00} $\pm$ 0.00 & \textbf{1.00} $\pm$ 0.00 & 0.50 $\pm$ 0.00 & 0.80 $\pm$ 0.02 & 0.58 $\pm$ 0.07 & 0.28 $\pm$ 0.11\\
\midrule
\begin{tabular}[c]{@{}c@{}}POMP (Ours)\end{tabular} & \textbf{1.00} $\pm$ 0.00 & \textbf{1.00} $\pm$ 0.00 & \textbf{1.00} $\pm$ 0.00 & \textbf{1.00} $\pm$ 0.00 & \textbf{1.00} $\pm$ 0.00 & \textbf{1.00} $\pm$ 0.00 & \textbf{1.00} $\pm$ 0.00 & 0.62 $\pm$ 0.01 & \textbf{0.97} $\pm$ 0.02 & 0.36 $\pm$ 0.02 \\ 
\bottomrule

\end{tabular}
}
\end{table*}

We aim to investigate whether our proposed \textit{diversity multiplier} can enhance CEM and yield programs with improved performance.
To this end, for each \textsc{Karel} or \textsc{Karel-Hard} task, 
we use CEM and CEM+diversity to find 10
programs.
Then, for each task, we evaluate all the programs and report the best performance in \mytable{tab:karel_Karel_hard_POMP_main}.
The results suggest that our proposed CEM+diversity achieves better performance on most of the tasks,
highlighting the improved search quality induced by covering wider regions in the search space with the \textit{diversity multiplier}.
Visualized search trajectories of CEM+diversity
 can be found in~\mysecref{app:cem_div}.


\vspacesubsection{Ablation Study}

We propose CEM+diversity+compatibility to retrieve a set of effective, diverse, compatible programs as modes of our Program Machine Policy.
This section compares a variety of implementations that consider the diversity and the compatibility of programs when retrieving them.

\begin{itemize}[leftmargin=3mm]
    \item \textbf{CEM $\times|M|$}: Conduct the CEM search described in~\mysecref{sec:cem}  $|M|$ times and take the resulting $|M|$ programs as the set of mode programs for each task.

    \item \textbf{CEM+diversity top $k$, $k=|M|$}: Conduct the CEM search with the \textit{diversity multiplier} described in~\mysecref{sec:cem_d} $N=10$ times and take the top $|M|$ results as the set of mode program embeddings for each task.
    
    \item \textbf{CEM+diversity $\times|M|$}: Conduct the CEM search with the \textit{diversity multiplier} described in~\mysecref{sec:cem_d} $N=10$ times and take the best program as the $i^{th}$ mode. Repeat this process $|M|$ times and take all $|M|$ programs as the mode program set for each task. 

    \item \textbf{POMP (Ours)}: Conduct CEM+diversity+compatibility (\ie CEM with the \textit{diversity multiplier} and $R_\Psi$ as described in~\mysecref{sec:cem_d_c}) for $N=10$ times and take the best result as the $i^{th}$ mode. Repeat the above process $|M|$ times and take all $|M|$ results as the set of mode program embeddings for each task. 
    Note that the whole procedure of retrieving programs using CEM+diversity+compatibility and learning a Program Machine Policy with retrieved mode programs is essentially our proposed framework, POMP.
\end{itemize}

Here the number of modes $|M|$ is 3 for \textsc{Seesaw, Up-N-Down, Inf-Harvester} and 5 for \textsc{Farmer} and \textsc{Inf-DoorKey}. 
We evaluate the quality of retrieved program sets according to the performance of the Program Machine Policy learned given these program sets on the \textsc{Karel-Long} tasks.
The results presented in~\mytable{tab:karel_long_ICML24} show that 
our proposed framework, POMP, outperforms its variants that ignore diversity or compatibility among modes on all the tasks.
This justifies our proposed 
CEM+diversity+compatibility 
for retrieving a set of effective, diverse, compatible programs as modes of our Program Machine Policy.



\vspacesubsection{Comparing with Deep RL and Programmatic RL Methods}

\begin{table*}[t]
\centering
\caption{\textbf{{\textsc{Karel-Long} Performance.}}
Mean return and standard deviation of all methods across the \textsc{Karel-Long} problem set, evaluated over five random seeds. 
Our proposed framework achieves the best mean reward across all tasks by learning a program machine policy with a set of effective, diverse, and compatible mode programs.
}
\vspace{-0.2cm}
\label{tab:karel_long_ICML24}
\resizebox{0.85\textwidth}{!}{%
\begin{tabular}{cccccc}
\toprule
   Method
 &
  \textsc{Seesaw} &
  \textsc{Up-N-Down} &
  \textsc{Farmer} &
  \textsc{Inf-DoorKey}  &\begin{tabular}[c]{@{}c@{}} \textsc{Inf-Harvester} \end{tabular} 
\\ 
\midrule

CEM $\times|M|$ & 0.06 $\pm$ 0.10& 0.39 $\pm$ 0.36& 0.03 $\pm$ 0.00& 0.11 $\pm$ 0.14 &0.41$\pm$ 0.17                            
\\
CEM+diversity top $k$, $k=|M|$ & 0.15 $\pm$ 0.21 & 0.25 $\pm$ 0.35 & 0.03 $\pm$ 0.00& 0.13 $\pm$ 0.16 &0.42$\pm$ 0.19
\\
CEM+diversity $\times|M|$ & 0.28 $\pm$ 0.23& 0.58 $\pm$ 0.31& \textcolor{black}{0.03 $\pm$ 0.00}& \textcolor{black}{0.36 $\pm$ 0.26} &0.47$\pm$ 0.23                 
\\ 
\midrule
DRL                                                                          & 0.00 $\pm$ 0.01& 0.00 $\pm$ 0.00& 0.38 $\pm$ 0.25& 0.17 $\pm$ 0.36 &0.74$\pm$ 0.05 
\\
Random Transition & 0.01 $\pm$ 0.00& 0.02 $\pm$ 0.01& 0.01 $\pm$ 0.00& 0.01 $\pm$ 0.01 &0.15$\pm$ 0.04                                                         
\\
PSMP                                                                         & 0.01 $\pm$ 0.01& 0.00 $\pm$ 0.00& 0.43 $\pm$ 0.23& 0.34 $\pm$ 0.45 &0.60 $\pm$ 0.04 
\\
MMN & 0.00 $\pm$ 0.01& 0.00 $\pm$ 0.00& \textcolor{black}{0.16 $\pm$ 0.27}& \textcolor{black}{0.01 $\pm$ 0.00} &\textcolor{black}{0.20 $\pm$ 0.09}                                                                           
\\
LEAPS & 0.01 $\pm$ 0.01& 0.02 $\pm$ 0.01& 0.03 $\pm$ 0.00& 0.01 $\pm$ 0.00 &0.12 $\pm$ 0.00                                                                         
\\
HPRL & \textcolor{black}{0.00 $\pm$ 0.00}& \textcolor{black}{0.00 $\pm$ 0.00}& \textcolor{black}{0.01 $\pm$ 0.00} & \textcolor{black}{0.00 $\pm$ 0.00} & 0.45 $\pm$ 0.03                                                                 
\\ 
\midrule
POMP (Ours) & \textcolor{black}{\textbf{0.53} $\pm$ 0.10}& \textcolor{black}{\textbf{0.76} $\pm$ 0.02}& \textcolor{black}{\textbf{0.62} $\pm$ 0.02}& \textcolor{black}{\textbf{0.66}$\pm$ 0.07} &\textcolor{black}{\textbf{0.79} $\pm$ 0.02} \\ 
\bottomrule

\end{tabular}%
}
\end{table*}

We compare our proposed framework and its variant to state-of-the-art deep RL and programmatic RL methods on the \textsc{Karel-Long} tasks.
\label{sec:main_result_baselines}
\begin{itemize}[leftmargin=3mm]
    \item \textbf{Random Transition} uses the same set of mode programs as POMP but with a random transition function (\ie uniformly randomly select the next mode at each step).
    The performance of this method examines the necessity to learn a transition function.

    \item \textbf{Programmatic State Machine Policy (PSMP)} learns a transition function as POMP while using primitive actions (\eg \texttt{move}, \texttt{pickMarker}) as modes. 
    Comparing POMP with this method highlights the effect of retrieving programs with higher-level behaviors as modes.

    \item \textbf{DRL} represents a policy as a neural network and is learned using PPO~\citep{schulman2017proximal}.
    The policy takes raw states (\ie Karel grids) as input and predicts the probability distribution over the set of primitive actions, (\eg \texttt{move}, \texttt{pickMarker}).

    \item \textbf{Moore Machine Network (MMN)}~\cite{koul2018learning} represents a recurrent policy with quantized memory and observations, which can be further extracted as a finite state machine.
    The policy takes raw states (\ie Karel grids) as input and predicts the probability distribution over the set of primitive actions (\eg \texttt{move}, \texttt{pickMarker}).

    \item \textbf{Learning Embeddings for Latent Program Synthesis (LEAPS)}~\cite{trivedi2021learning}
    searches for a single task-solving program using the vanilla CEM in a learned program embedding space.
    \item \textbf{Hierarchical Programmatic Reinforcement Learning (HPRL)}~\cite{liu2023hierarchical} 
    learns a meta-policy, whose action space is a learned program embedding space, to compose a series of programs to produce a program policy.
\end{itemize}

\begin{figure*}
    \begin{subfigure}[b]{0.49\textwidth}
    \centering
    \includegraphics[trim=80 0 100 0,clip,width=0.99\textwidth]{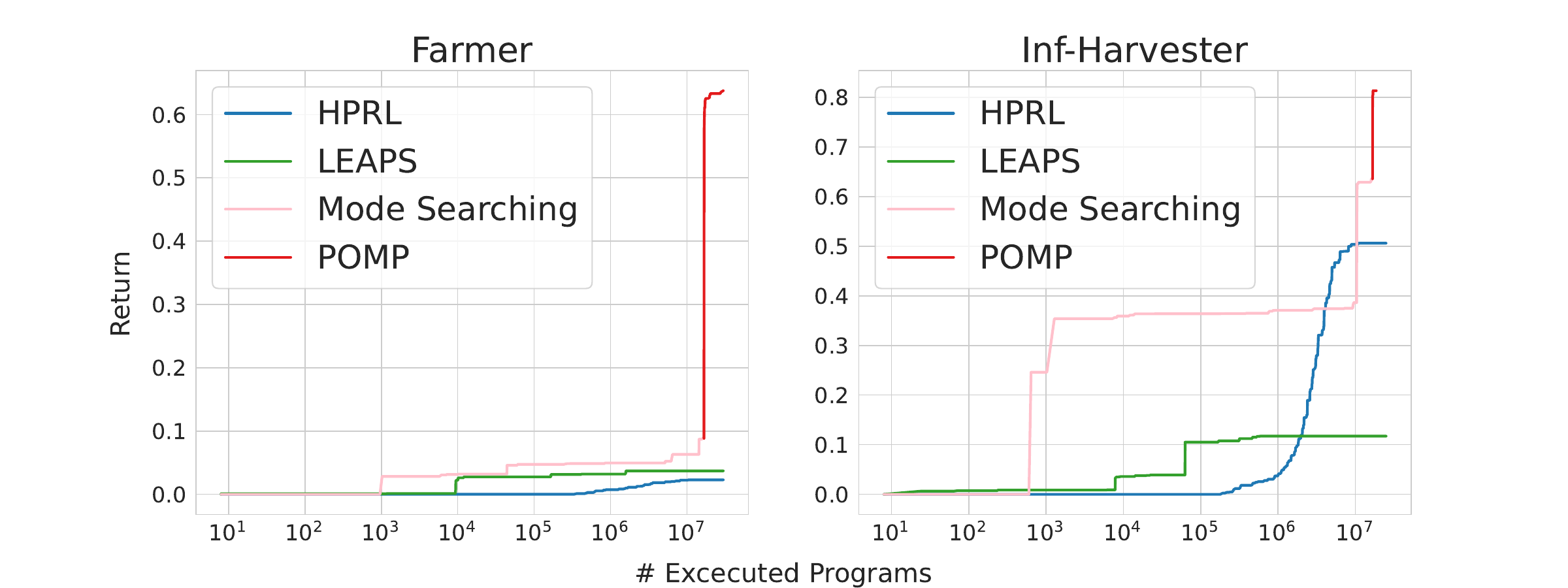}
    \caption{Sample Efficiency}
    \label{fig:se}
    \end{subfigure}
    \begin{subfigure}[b]{0.49\textwidth}
    \centering
    \includegraphics[trim=80 0 100 0,clip,width=0.99\textwidth]{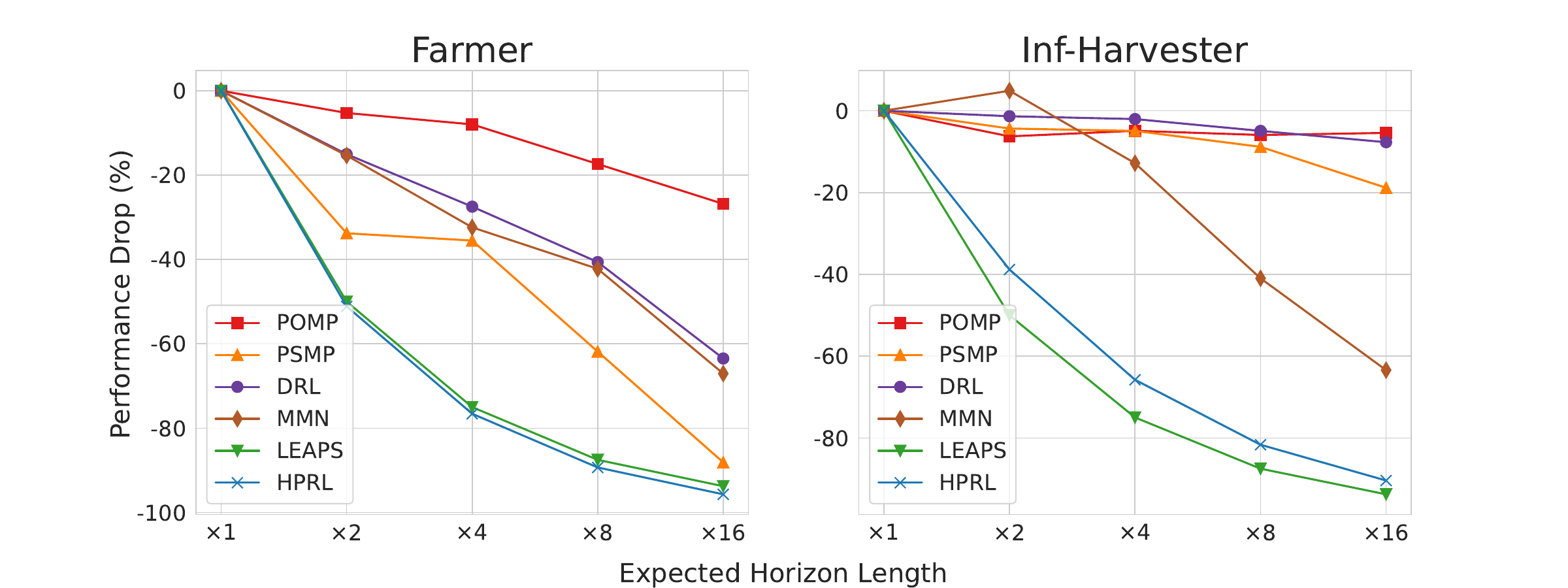}
    \caption{Inductive Generalization}
    \label{fig:ig}
    \end{subfigure}

    \vspace{-0.2cm}
    \caption[]{
        \small 
        (a) \textbf{Program sample efficiency.} The training curves of POMP and other programmatic RL approaches, where the x-axis is the total number of executed programs for interacting with the environment, and the y-axis is the maximum validation return. 
        This demonstrates that our proposed framework has better program sample efficiency and converges to better performance.
        (b) \textbf{Inductive generalization performance.}
        We evaluate and report the performance drop in the testing environments with an extended horizon, where the x-axis is the extended horizon length compared to the horizon of the training environments, and the y-axis is the performance drop in percentage.
        Our proposed framework can inductively generalize to longer horizons without any fine-tuning.
        \label{fig:inductive_curve}
    }
\end{figure*}

As \mytable{tab:karel_Karel_hard_POMP_main} shows, POMP outperforms LEAPS and HPRL on eight out of ten tasks from the \textsc{Karel} and \textsc{Karel-Hard} tasks, indicating that the retrieved mode programs are truly effective at solving short horizon tasks (\ie less than 500 actions).
For long-horizon tasks that require more than 3000 actions to solve, \mytable{tab:karel_long_ICML24} shows that POMP excels on all five tasks, with better performance on \textsc{Farmer} and \textsc{Inf-Harvester} and particular prowess in \textsc{Seesaw}, \textsc{Up-N-Down}, and \textsc{Inf-Doorkey}.

Two of these tasks require distinct skills (\eg pick and put markers in \textsc{Farmer}; go up and downstairs in \textsc{Up-N-Down}) and the capability to persistently execute one skill for an extended period before transitioning to another. 
POMP adeptly addresses this challenge due to the consideration of diversity when seeking mode programs, which ensures the acquisition of both skills concurrently. 
Furthermore, the state machine architecture of our approach provides not only the sustained execution of a singular skill but also the timely transition to another, as needed.

Unlike the other tasks, \textsc{Seesaw} and \textsc{Inf-Doorkey} demand an extended traverse to collect markers, resulting in a more sparse reward distribution.
During the search for mode programs, the emphasis on compatibility allows POMP to secure a set of mutually compatible modes that collaborate effectively to perform extended traversal.
Some retrieved programs are shown in Appendix (\myfig{fig:karel_program_examples_seesaw_upNdown}, \myfig{fig:karel_program_examples_farmer}, \myfig{fig:karel_program_examples_infDoorKey}, and \myfig{fig:karel_program_examples_harvester}).

\vspacesubsection{Program Sample Efficiency}
To accurately evaluate the sample efficiency of programmatic RL methods, we propose the concept of 
\textit{program sample efficiency}, which measures the total number of program executions required to learn a program policy.
We report the program sample efficiency of LEAPS, HPRL, and POMP on \textsc{Farmer} and \textsc{Inf-Harvester}, as shown in~\myfig{fig:se}.
POMP has better sample efficiency than LEAPS and HPRL, indicating that our framework requires fewer environmental interactions and computational costs. 
More details can be found in~\mysecref{app:se_exp}.

\vspacesubsection{Inductive Generalization}

We aim to compare the inductive generalization ability of all the methods,
which requires generalizing to out-of-distributionally (\ie unseen during training) longer task instances~\citep{Inala2020SynthesizingInductGen}.
To this end, we increase the expected horizons of \textsc{Farmer} and \textsc{Inf-Harvester} by $2\times$, $4\times$, $8\times$, and $16\times$, evaluate all the learned policies, and report the performance drop compared to the original task performance in~\myfig{fig:ig}.
More details on extending task horizons can be found in~\mysecref{app:ig_exp}.

The results show that 
POMP experiences a smaller decline in performance in these testing environments with significantly extended horizons. 
This suggests that our approach exhibits superior inductive generalization in these tasks.
The longest execution of POMP runs up to $48$k environment steps. 


\vspacesection{Conclusion}
\label{sec:conclusion}
This work aims to produce reinforcement learning policies that are human-interpretable and can inductively generalize by bridging program synthesis and state machines.
To this end, we present the Program Machine Policy (POMP) framework for representing complex behaviors and addressing long-horizon tasks.
Specifically, we introduce a method that can retrieve a set of effective, diverse, compatible programs by modifying the Cross Entropy Method (CEM).
Then, we propose to use these programs as modes of a state machine and learn a transition function to transit among mode programs using reinforcement learning.
To evaluate the ability to solve tasks with extended horizons, we design a set of tasks that requires thousands of steps in the Karel domain.
Our framework POMP outperforms various deep RL and programmatic RL methods on various tasks.
Also, POMP demonstrates superior performance in inductively generalizing to even longer horizons without fine-tuning.
Extensive ablation studies justify the effectiveness of our proposed search algorithm to retrieve mode programs and our proposed method to learn a transition function.

\section*{Acknowledgement}
\label{sec:ack}

This work was supported by the National Taiwan University and its Department of Electrical Engineering, Graduate Institute of Networking and Multimedia, Graduate Institute of Communication Engineering, and College of Electrical Engineering and Computer Science. Shao-Hua Sun was also partially supported by the Yushan Fellow Program by the Taiwan Ministry of Education.



\nocite{langley00}

\bibliography{ref}

\begin{thebibliography}{68}
\providecommand{\natexlab}[1]{#1}
\providecommand{\url}[1]{\texttt{#1}}
\expandafter\ifx\csname urlstyle\endcsname\relax
  \providecommand{\doi}[1]{doi: #1}\else
  \providecommand{\doi}{doi: \begingroup \urlstyle{rm}\Url}\fi

\bibitem[Aleixo \& Lelis(2023)Aleixo and Lelis]{aleixo2023show}
Aleixo, D.~S. and Lelis, L.~H.
\newblock Show me the way! bilevel search for synthesizing programmatic strategies.
\newblock In \emph{Association for the Advancement of Artificial Intelligence}, 2023.

\bibitem[Andreas et~al.(2017)Andreas, Klein, and Levine]{andreas2016modular}
Andreas, J., Klein, D., and Levine, S.
\newblock Modular multitask reinforcement learning with policy sketches.
\newblock In \emph{International Conference on Machine Learning}, 2017.

\bibitem[Bacon et~al.(2017)Bacon, Harb, and Precup]{bacon2017option}
Bacon, P.-L., Harb, J., and Precup, D.
\newblock The option-critic architecture.
\newblock In \emph{Association for the Advancement of Artificial Intelligence}, 2017.

\bibitem[Balog et~al.(2017)Balog, Gaunt, Brockschmidt, Nowozin, and Tarlow]{balog2016deepcoder}
Balog, M., Gaunt, A.~L., Brockschmidt, M., Nowozin, S., and Tarlow, D.
\newblock Deepcoder: Learning to write programs.
\newblock In \emph{International Conference on Learning Representations}, 2017.

\bibitem[Barto \& Mahadevan(2003)Barto and Mahadevan]{barto2003recent}
Barto, A.~G. and Mahadevan, S.
\newblock Recent advances in hierarchical reinforcement learning.
\newblock \emph{Discrete Event Dynamic Systems}, 2003.

\bibitem[Bastani et~al.(2018)Bastani, Pu, and Solar-Lezama]{bastani2018verifiable}
Bastani, O., Pu, Y., and Solar-Lezama, A.
\newblock Verifiable reinforcement learning via policy extraction.
\newblock In \emph{Neural Information Processing Systems}, 2018.

\bibitem[Bunel et~al.(2018)Bunel, Hausknecht, Devlin, Singh, and Kohli]{bunel2018leveraging}
Bunel, R.~R., Hausknecht, M., Devlin, J., Singh, R., and Kohli, P.
\newblock Leveraging grammar and reinforcement learning for neural program synthesis.
\newblock In \emph{International Conference on Learning Representations}, 2018.

\bibitem[Chen et~al.(2019)Chen, Liu, and Song]{chen2019executionguided}
Chen, X., Liu, C., and Song, D.
\newblock Execution-guided neural program synthesis.
\newblock In \emph{International Conference on Learning Representations}, 2019.

\bibitem[Cheng \& Xu(2023)Cheng and Xu]{ChengXu2023LEAGUE}
Cheng, S. and Xu, D.
\newblock {LEAGUE}: Guided skill learning and abstraction for long-horizon manipulation.
\newblock \emph{IEEE Robotics and Automation Letters}, 2023.

\bibitem[Chevalier-Boisvert et~al.(2023)Chevalier-Boisvert, Dai, Towers, de~Lazcano, Willems, Lahlou, Pal, Castro, and Terry]{chevalier2023minigrid}
Chevalier-Boisvert, M., Dai, B., Towers, M., de~Lazcano, R., Willems, L., Lahlou, S., Pal, S., Castro, P.~S., and Terry, J.
\newblock Minigrid \& miniworld: Modular \& customizable reinforcement learning environments for goal-oriented tasks.
\newblock \emph{arXiv preprint arXiv:2306.13831}, 2023.

\bibitem[Cho et~al.(2014)Cho, van Merrienboer, Gulcehre, Bahdanau, Bougares, Schwenk, and Bengio]{cho2014learning}
Cho, K., van Merrienboer, B., Gulcehre, C., Bahdanau, D., Bougares, F., Schwenk, H., and Bengio, Y.
\newblock Learning phrase representations using rnn encoder-decoder for statistical machine translation, 2014.

\bibitem[Choi \& Langley(2005)Choi and Langley]{choi2005learning}
Choi, D. and Langley, P.
\newblock Learning teleoreactive logic programs from problem solving.
\newblock In \emph{International Conference on Inductive Logic Programming}, 2005.

\bibitem[Cobbe et~al.(2019)Cobbe, Klimov, Hesse, Kim, and Schulman]{cobbe2019quantifying}
Cobbe, K., Klimov, O., Hesse, C., Kim, T., and Schulman, J.
\newblock Quantifying generalization in reinforcement learning.
\newblock In \emph{International Conference on Machine Learning}, 2019.

\bibitem[Devlin et~al.(2017)Devlin, Uesato, Bhupatiraju, Singh, Mohamed, and Kohli]{devlin2017robustfill}
Devlin, J., Uesato, J., Bhupatiraju, S., Singh, R., Mohamed, A.-r., and Kohli, P.
\newblock Robustfill: Neural program learning under noisy i/o.
\newblock In \emph{International Conference on Machine Learning}, 2017.

\bibitem[Dietterich(2000)]{dietterich1999hierarchical}
Dietterich, T.~G.
\newblock Hierarchical reinforcement learning with the {MAXQ} value function decomposition.
\newblock \emph{Journal of Artificial Intelligence Research}, 2000.

\bibitem[Ellis et~al.(2020)Ellis, Wong, Nye, Sable-Meyer, Cary, Morales, Hewitt, Solar-Lezama, and Tenenbaum]{ellis2020dreamcoder}
Ellis, K., Wong, C., Nye, M., Sable-Meyer, M., Cary, L., Morales, L., Hewitt, L., Solar-Lezama, A., and Tenenbaum, J.~B.
\newblock Dreamcoder: Growing generalizable, interpretable knowledge with wake-sleep bayesian program learning.
\newblock \emph{arXiv preprint arXiv:2006.08381}, 2020.

\bibitem[Frans et~al.(2018)Frans, Ho, Chen, Abbeel, and Schulman]{MLSH}
Frans, K., Ho, J., Chen, X., Abbeel, P., and Schulman, J.
\newblock Meta learning shared hierarchies.
\newblock In \emph{International Conference on Learning Representations}, 2018.

\bibitem[Fukushima \& Miyake(1982)Fukushima and Miyake]{fukushima1982neocognitron}
Fukushima, K. and Miyake, S.
\newblock Neocognitron: A self-organizing neural network model for a mechanism of visual pattern recognition.
\newblock In \emph{Competition and Cooperation in Neural Nets: Proceedings of the US-Japan Joint Seminar}, 1982.

\bibitem[Furelos-Blanco et~al.(2023)Furelos-Blanco, Law, Jonsson, Broda, and Russo]{FurelosBlanco2022HierarchiesRM}
Furelos-Blanco, D., Law, M., Jonsson, A., Broda, K., and Russo, A.
\newblock Hierarchies of reward machines.
\newblock In \emph{International Conference on Machine Learning}, 2023.

\bibitem[Gu et~al.(2017)Gu, Holly, Lillicrap, and Levine]{gu2017deep}
Gu, S., Holly, E., Lillicrap, T., and Levine, S.
\newblock Deep reinforcement learning for robotic manipulation with asynchronous off-policy updates.
\newblock In \emph{IEEE International Conference on Robotics and Automation}, 2017.

\bibitem[Guan et~al.(2022)Guan, Sreedharan, and Kambhampati]{Lin2022Leveraging}
Guan, L., Sreedharan, S., and Kambhampati, S.
\newblock Leveraging approximate symbolic models for reinforcement learning via skill diversity.
\newblock In \emph{International Conference on Machine Learning}, 2022.

\bibitem[Hasanbeig et~al.(2021)Hasanbeig, Jeppu, Abate, Melham, and Kroening]{Hasanbeig2021DeepSynthAS}
Hasanbeig, M., Jeppu, N.~Y., Abate, A., Melham, T.~F., and Kroening, D.
\newblock Deepsynth: Automata synthesis for automatic task segmentation in deep reinforcement learning.
\newblock In \emph{Association for the Advancement of Artificial Intelligence}, 2021.

\bibitem[Higgins et~al.(2016)Higgins, Matthey, Pal, Burgess, Glorot, Botvinick, Mohamed, and Lerchner]{higgins2016beta}
Higgins, I., Matthey, L., Pal, A., Burgess, C., Glorot, X., Botvinick, M., Mohamed, S., and Lerchner, A.
\newblock beta-vae: Learning basic visual concepts with a constrained variational framework.
\newblock In \emph{International Conference on Learning Representations}, 2016.

\bibitem[Ho \& Ermon(2016)Ho and Ermon]{ho2016generative}
Ho, J. and Ermon, S.
\newblock Generative adversarial imitation learning.
\newblock In \emph{Advances in Neural Information Processing Systems}, 2016.

\bibitem[Hong et~al.(2021)Hong, Dohan, Singh, Sutton, and Zaheer]{hong2020latent}
Hong, J., Dohan, D., Singh, R., Sutton, C., and Zaheer, M.
\newblock Latent programmer: Discrete latent codes for program synthesis.
\newblock In \emph{International Conference on Machine Learning}, 2021.

\bibitem[Ibarz et~al.(2021)Ibarz, Tan, Finn, Kalakrishnan, Pastor, and Levine]{ibarz2021train}
Ibarz, J., Tan, J., Finn, C., Kalakrishnan, M., Pastor, P., and Levine, S.
\newblock How to train your robot with deep reinforcement learning: lessons we have learned.
\newblock \emph{The International Journal of Robotics Research}, 2021.

\bibitem[Icarte et~al.(2018)Icarte, Klassen, Valenzano, and McIlraith]{ICML18_Icarte_UsingRM}
Icarte, R.~T., Klassen, T., Valenzano, R., and McIlraith, S.
\newblock Using reward machines for high-level task specification and decomposition in reinforcement learning.
\newblock In \emph{International Conference on Machine Learning}, 2018.

\bibitem[Inala et~al.(2020)Inala, Bastani, Tavares, and Solar-Lezama]{Inala2020SynthesizingInductGen}
Inala, J.~P., Bastani, O., Tavares, Z., and Solar-Lezama, A.
\newblock Synthesizing programmatic policies that inductively generalize.
\newblock In \emph{International Conference on Learning Representations}, 2020.

\bibitem[Kempka et~al.(2016)Kempka, Wydmuch, Runc, Toczek, and Ja{\'s}kowski]{kempka2016vizdoom}
Kempka, M., Wydmuch, M., Runc, G., Toczek, J., and Ja{\'s}kowski, W.
\newblock Vizdoom: A doom-based ai research platform for visual reinforcement learning.
\newblock In \emph{IEEE Conference on Computational Intelligence and Games}, 2016.

\bibitem[Klissarov \& Precup(2021)Klissarov and Precup]{klissarov2021flexible}
Klissarov, M. and Precup, D.
\newblock Flexible option learning.
\newblock In \emph{Neural Information Processing Systems}, 2021.

\bibitem[Koul et~al.(2019)Koul, Fern, and Greydanus]{koul2018learning}
Koul, A., Fern, A., and Greydanus, S.
\newblock Learning finite state representations of recurrent policy networks.
\newblock In \emph{International Conference on Learning Representations}, 2019.

\bibitem[Krizhevsky et~al.(2017)Krizhevsky, Sutskever, and Hinton]{krizhevsky2017imagenet}
Krizhevsky, A., Sutskever, I., and Hinton, G.~E.
\newblock Imagenet classification with deep convolutional neural networks.
\newblock \emph{Communications of the ACM}, 2017.

\bibitem[Landajuela et~al.(2021)Landajuela, Petersen, Kim, Santiago, Glatt, Mundhenk, Pettit, and Faissol]{landajuela21a}
Landajuela, M., Petersen, B.~K., Kim, S., Santiago, C.~P., Glatt, R., Mundhenk, N., Pettit, J.~F., and Faissol, D.
\newblock Discovering symbolic policies with deep reinforcement learning.
\newblock In \emph{International Conference on Machine Learning}, 2021.

\bibitem[Langley(2000)]{langley00}
Langley, P.
\newblock Crafting papers on machine learning.
\newblock In Langley, P. (ed.), \emph{Proceedings of the 17th International Conference on Machine Learning (ICML 2000)}, pp.\  1207--1216, Stanford, CA, 2000. Morgan Kaufmann.

\bibitem[Lee et~al.(2019)Lee, Sun, Somasundaram, Hu, and Lim]{lee2019composing}
Lee, Y., Sun, S.-H., Somasundaram, S., Hu, E., and Lim, J.~J.
\newblock Composing complex skills by learning transition policies.
\newblock In \emph{International Conference on Learning Representations}, 2019.

\bibitem[Lee et~al.(2021)Lee, Szot, Sun, and Lim]{lee2021generalizable}
Lee, Y., Szot, A., Sun, S.-H., and Lim, J.~J.
\newblock Generalizable imitation learning from observation via inferring goal proximity.
\newblock In \emph{Neural Information Processing Systems}, 2021.

\bibitem[Liang et~al.(2023)Liang, Huang, Xia, Xu, Hausman, Ichter, Florence, and Zeng]{liang2023code}
Liang, J., Huang, W., Xia, F., Xu, P., Hausman, K., Ichter, B., Florence, P., and Zeng, A.
\newblock Code as policies: Language model programs for embodied control.
\newblock In \emph{International Conference on Robotics and Automation}, 2023.

\bibitem[Lipton(2016)]{lipton2018mythos}
Lipton, Z.~C.
\newblock The mythos of model interpretability.
\newblock In \emph{ICML Workshop on Human Interpretability in Machine Learning}, 2016.

\bibitem[Liu et~al.(2023)Liu, Hu, Cheng, Lee, and Sun]{liu2023hierarchical}
Liu, G.-T., Hu, E.-P., Cheng, P.-J., Lee, H.-Y., and Sun, S.-H.
\newblock Hierarchical programmatic reinforcement learning via learning to compose programs.
\newblock In \emph{International Conference on Machine Learning}, 2023.

\bibitem[Liu et~al.(2024)Liu, Chang, Sun, and Yu]{liu2024integrating}
Liu, J.-C., Chang, C.-H., Sun, S.-H., and Yu, T.-L.
\newblock Integrating planning and deep reinforcement learning via automatic induction of task substructures.
\newblock In \emph{International Conference on Learning Representations}, 2024.

\bibitem[Nam et~al.(2022)Nam, Sun, Pertsch, Hwang, and Lim]{skill2022nam}
Nam, T., Sun, S.-H., Pertsch, K., Hwang, S.~J., and Lim, J.~J.
\newblock Skill-based meta-reinforcement learning.
\newblock In \emph{International Conference on Learning Representations}, 2022.

\bibitem[Pattis(1981)]{pattis1981karel}
Pattis, R.~E.
\newblock \emph{Karel the robot: a gentle introduction to the art of programming}.
\newblock John Wiley \& Sons, Inc., 1981.

\bibitem[Polydoros \& Nalpantidis(2017)Polydoros and Nalpantidis]{polydoros2017survey}
Polydoros, A.~S. and Nalpantidis, L.
\newblock Survey of model-based reinforcement learning: Applications on robotics.
\newblock \emph{Journal of Intelligent \& Robotic Systems}, 2017.

\bibitem[Rubinstein(1997)]{rubinstein1997optimization}
Rubinstein, R.~Y.
\newblock Optimization of computer simulation models with rare events.
\newblock \emph{European Journal of Operational Research}, 1997.

\bibitem[Schaal(1997)]{schaal1997learning}
Schaal, S.
\newblock Learning from demonstration.
\newblock In \emph{Advances in Neural Information Processing Systems}, 1997.

\bibitem[Schulman et~al.(2017)Schulman, Wolski, Dhariwal, Radford, and Klimov]{schulman2017proximal}
Schulman, J., Wolski, F., Dhariwal, P., Radford, A., and Klimov, O.
\newblock Proximal policy optimization algorithms.
\newblock \emph{arXiv preprint arXiv:1707.06347}, 2017.

\bibitem[Shen(2020)]{shen2020}
Shen, O.
\newblock Interpretability in {ML}: A broad overview, 2020.
\newblock URL \url{https://mlu.red/muse/52906366310}.

\bibitem[Shin et~al.(2018)Shin, Polosukhin, and Song]{shin2018improving}
Shin, E.~C., Polosukhin, I., and Song, D.
\newblock Improving neural program synthesis with inferred execution traces.
\newblock In \emph{Neural Information Processing Systems}, 2018.

\bibitem[Silver et~al.(2016)Silver, Huang, Maddison, Guez, Sifre, Van Den~Driessche, Schrittwieser, Antonoglou, Panneershelvam, Lanctot, et~al.]{silver2016mastering}
Silver, D., Huang, A., Maddison, C.~J., Guez, A., Sifre, L., Van Den~Driessche, G., Schrittwieser, J., Antonoglou, I., Panneershelvam, V., Lanctot, M., et~al.
\newblock Mastering the game of go with deep neural networks and tree search.
\newblock \emph{Nature}, 2016.

\bibitem[Silver et~al.(2017)Silver, Schrittwieser, Simonyan, Antonoglou, Huang, Guez, Hubert, Baker, Lai, Bolton, Chen, Lillicrap, Hui, Sifre, van~den Driessche, Graepel, and Hassabis]{silver2017mastering}
Silver, D., Schrittwieser, J., Simonyan, K., Antonoglou, I., Huang, A., Guez, A., Hubert, T., Baker, L., Lai, M., Bolton, A., Chen, Y., Lillicrap, T., Hui, F., Sifre, L., van~den Driessche, G., Graepel, T., and Hassabis, D.
\newblock Mastering the game of go without human knowledge.
\newblock \emph{Nature}, 2017.

\bibitem[Sun et~al.(2018)Sun, Noh, Somasundaram, and Lim]{sun2018neural}
Sun, S.-H., Noh, H., Somasundaram, S., and Lim, J.
\newblock Neural program synthesis from diverse demonstration videos.
\newblock In \emph{International Conference on Machine Learning}, 2018.

\bibitem[Sun et~al.(2020)Sun, Wu, and Lim]{sun2020program}
Sun, S.-H., Wu, T.-L., and Lim, J.~J.
\newblock Program guided agent.
\newblock In \emph{International Conference on Learning Representations}, 2020.

\bibitem[Sutton et~al.(1999)Sutton, Precup, and Singh]{sutton1999between}
Sutton, R.~S., Precup, D., and Singh, S.
\newblock Between mdps and semi-mdps: A framework for temporal abstraction in reinforcement learning.
\newblock \emph{Artificial intelligence}, 1999.

\bibitem[Toro~Icarte et~al.(2019)Toro~Icarte, Waldie, Klassen, Valenzano, Castro, and McIlraith]{NEURIPS19_Icarte_LearningRM}
Toro~Icarte, R., Waldie, E., Klassen, T., Valenzano, R., Castro, M., and McIlraith, S.
\newblock Learning reward machines for partially observable reinforcement learning.
\newblock In \emph{Neural Information Processing Systems}, 2019.

\bibitem[Trivedi et~al.(2021)Trivedi, Zhang, Sun, and Lim]{trivedi2021learning}
Trivedi, D., Zhang, J., Sun, S.-H., and Lim, J.~J.
\newblock Learning to synthesize programs as interpretable and generalizable policies.
\newblock In \emph{Neural Information Processing Systems}, 2021.

\bibitem[Verma et~al.(2018)Verma, Murali, Singh, Kohli, and Chaudhuri]{verma2018programmatically}
Verma, A., Murali, V., Singh, R., Kohli, P., and Chaudhuri, S.
\newblock Programmatically interpretable reinforcement learning.
\newblock In \emph{International Conference on Machine Learning}, 2018.

\bibitem[Verma et~al.(2019)Verma, Le, Yue, and Chaudhuri]{verma2019imitation}
Verma, A., Le, H., Yue, Y., and Chaudhuri, S.
\newblock Imitation-projected programmatic reinforcement learning.
\newblock In \emph{Neural Information Processing Systems}, 2019.

\bibitem[Vezhnevets et~al.(2017)Vezhnevets, Osindero, Schaul, Heess, Jaderberg, Silver, and Kavukcuoglu]{vezhnevets2017feudal}
Vezhnevets, A.~S., Osindero, S., Schaul, T., Heess, N., Jaderberg, M., Silver, D., and Kavukcuoglu, K.
\newblock Feudal networks for hierarchical reinforcement learning.
\newblock In \emph{International Conference on Machine Learning}, 2017.

\bibitem[Vinyals et~al.(2019)Vinyals, Babuschkin, Czarnecki, Mathieu, Dudzik, Chung, Choi, Powell, Ewalds, Georgiev, et~al.]{vinyals2019grandmaster}
Vinyals, O., Babuschkin, I., Czarnecki, W.~M., Mathieu, M., Dudzik, A., Chung, J., Choi, D.~H., Powell, R., Ewalds, T., Georgiev, P., et~al.
\newblock Grandmaster level in starcraft ii using multi-agent reinforcement learning.
\newblock \emph{Nature}, 2019.

\bibitem[Wang et~al.(2023{\natexlab{a}})Wang, Gonzalez-Pumariega, Sharma, and Choudhury]{wang2023demo2code}
Wang, H., Gonzalez-Pumariega, G., Sharma, Y., and Choudhury, S.
\newblock Demo2code: From summarizing demonstrations to synthesizing code via extended chain-of-thought.
\newblock \emph{arXiv preprint arXiv:2305.16744}, 2023{\natexlab{a}}.

\bibitem[Wang et~al.(2023{\natexlab{b}})Wang, Chen, and Sun]{wang2023diffusion}
Wang, H.-C., Chen, S.-F., and Sun, S.-H.
\newblock Diffusion model-augmented behavioral cloning.
\newblock \emph{arXiv preprint arXiv:2302.13335}, 2023{\natexlab{b}}.

\bibitem[Winner \& Veloso(2003)Winner and Veloso]{distill}
Winner, E. and Veloso, M.
\newblock Distill: Learning domain-specific planners by example.
\newblock In \emph{International Conference on Machine Learning}, 2003.

\bibitem[Wurman et~al.(2022)Wurman, Barrett, Kawamoto, MacGlashan, Subramanian, Walsh, Capobianco, Devlic, Eckert, Fuchs, et~al.]{wurman2022outracing}
Wurman, P.~R., Barrett, S., Kawamoto, K., MacGlashan, J., Subramanian, K., Walsh, T.~J., Capobianco, R., Devlic, A., Eckert, F., Fuchs, F., et~al.
\newblock Outracing champion gran turismo drivers with deep reinforcement learning.
\newblock \emph{Nature}, 2022.

\bibitem[Xu et~al.(2020)Xu, Gavran, Ahmad, Majumdar, Neider, Topcu, and Wu]{xu2022JointInfRM}
Xu, Z., Gavran, I., Ahmad, Y., Majumdar, R., Neider, D., Topcu, U., and Wu, B.
\newblock Joint inference of reward machines and policies for reinforcement learning.
\newblock In \emph{International Conference on Automated Planning and Scheduling}, 2020.

\bibitem[Yang et~al.(2018)Yang, Lyu, Liu, and Gustafson]{Yang2018PEORL}
Yang, F., Lyu, D., Liu, B., and Gustafson, S.
\newblock {PEORL}: Integrating symbolic planning and hierarchical reinforcement learning for robust decision-making.
\newblock In \emph{International Joint Conference on Artificial Intelligence}, 2018.

\bibitem[Zhang et~al.(2018)Zhang, Vinyals, Munos, and Bengio]{zhang2018study}
Zhang, C., Vinyals, O., Munos, R., and Bengio, S.
\newblock A study on overfitting in deep reinforcement learning.
\newblock \emph{arXiv preprint arXiv:1804.06893}, 2018.

\bibitem[Zhang et~al.(2023)Zhang, Jiahui~Zhang, Liu, Ren, Chang, Sun, and Lim]{zhang2023bootstrap}
Zhang, J., Jiahui~Zhang, K.~P., Liu, Z., Ren, X., Chang, M., Sun, S.-H., and Lim, J.~J.
\newblock Bootstrap your own skills: Learning to solve new tasks with large language model guidance.
\newblock In \emph{Conference on Robot Learning}, 2023.

\bibitem[Zhong et~al.(2023)Zhong, Lindeborg, Zhang, Lim, and Sun]{zhong2023hierarchical}
Zhong, L., Lindeborg, R., Zhang, J., Lim, J.~J., and Sun, S.-H.
\newblock Hierarchical neural program synthesis.
\newblock \emph{arXiv preprint arXiv:2303.06018}, 2023.

\end{thebibliography}
\bibliographystyle{icml2024}

\newpage
\appendix
\onecolumn

\section*{Appendix}




\section{Extended Related Work}
\label{app:extend_related_work}


\myparagraph{Hierarchical Reinforcement Learning and Semi-Markov Decision Processes}
Hierarchical Reinforcement Learning (HRL) frameworks~\citep{sutton1999between, barto2003recent, vezhnevets2017feudal, bacon2017option, lee2019composing} focus on learning and operating across different levels of temporal abstraction, enhancing the efficiency of learning and exploration, particularly in sparse-reward environments. In this work, our proposed Program Machine Policy shares the same spirit and some ideas with HRL frameworks if we view the transition function as a “high-level” policy and the set of mode programs as “low-level” policies or skills. While most HRL frameworks either pre-define and pre-learned low-level policies~\citep{lee2019composing, skill2022nam, zhang2023bootstrap} through RL or imitation learning~\citep{schaal1997learning, ho2016generative, lee2021generalizable, wang2023diffusion}, or jointly learn the high-level and low-level policies from scratch~\citep{vezhnevets2017feudal, bacon2017option, MLSH}, our proposed framework first retrieves a set of effective, diverse, and compatible low-level policies (\ie program modes) via a search method, and then learns the high-level (\ie the mode transition function).

The POMP framework also resembles the option framework~\cite{sutton1999between, bacon2017option, klissarov2021flexible}.
More specifically, one can characterize POMP as using interpretable options as sub-policies since there is a high-level neural network being used to pick among retrieved programs as described in \mysecref{sec:approach_stage3}. Besides interpretable options, our work differs from the option frameworks in the following aspects. Our work first retrieves a set of mode programs and then learns a transition function; this differs from most option frameworks that jointly learn options and a high-level policy that chooses options. Also, the transition function in our work learns to terminate, while most high-level policies in option frameworks do not.

On the other hand, based on the definition of the recursive optimality described in~\cite{dietterich1999hierarchical}, POMP can be categorized as recursively optimal since it is locally optimal given the policies of its children. Specifically, one can view the mode program retrieval process of POMP as solving a set of subtasks based on the proposed CEM-based search method that considers effectiveness, diversity, and compatibility. Then, POMP learns a transition function according to the retrieved programs, resulting in a policy as a whole. 

\myparagraph{Symbolic Planning for Long-Horizon Tasks}
Another line of research uses symbolic operators~\citep{Yang2018PEORL, Lin2022Leveraging, ChengXu2023LEAGUE, liu2024integrating} for long-horizon planning.
The major difference between POMP and~\cite{ChengXu2023LEAGUE} and~\cite{Lin2022Leveraging} is the interpretability of the skills or options. In POMP, each learned skill is represented by a human-readable program. On the other hand, neural networks used in~\cite{ChengXu2023LEAGUE} and tabular approaches used in~\cite{Lin2022Leveraging} are used to learn the skill policies. In~\cite{Yang2018PEORL}, the option set is assumed as input without learning and cannot be directly compared with~\cite{ChengXu2023LEAGUE}, ~\cite{Lin2022Leveraging} and POMP.

Another difference between the proposed POMP framework,~\cite{ChengXu2023LEAGUE},~\cite{Lin2022Leveraging}, and~\cite{Yang2018PEORL} is whether the high-level transition abstraction is provided as input. In~\cite{ChengXu2023LEAGUE}, a library of skill operators is taken as input and serves as the basis for skill learning. In~\cite{Lin2022Leveraging}, the set of “landmarks” is taken as input to decompose the task into different combinations of subgoals. In PEORL~\citep{Yang2018PEORL}, the options set is taken as input, and each option has a 1-1 mapping with each transition in the high-level planning. On the other hand, the proposed POMP framework utilized the set of retrieved programs as modes, which is conducted based on the reward from the target task without any guidance from framework input.

\section{Cross Entropy Method Details}
\label{app:cem_details}

\subsection{\textbf{CEM}}
\label{app:cem}
\begin{figure}[b]
\centering
\includegraphics[width=.45\linewidth]{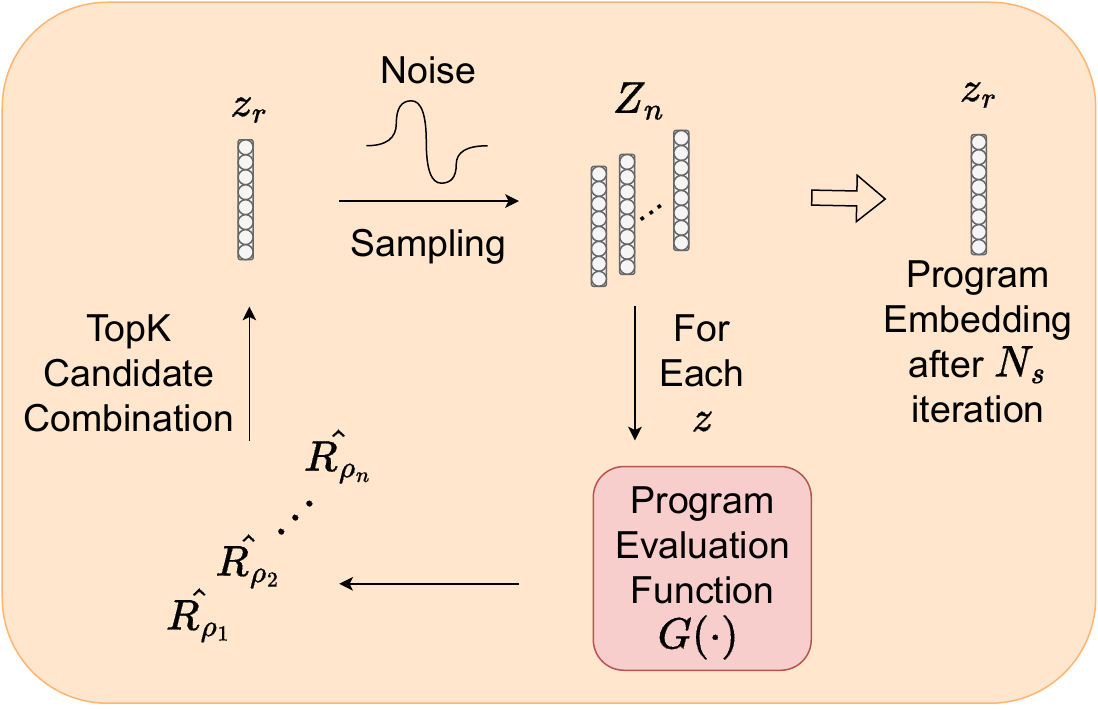}  
\caption{ 
Using the Cross-Entropy Method to search for a program with high execution reward in the learned program embedding space.
}
\label{fig:stage2_cem}
\end{figure}
\myfig{fig:stage2_cem} illustrates the workflow of the Cross Entropy Method (CEM). The corresponding pseudo-code is provided in~\myalgo{alg:cem_pseudo}. 

Detailed hyperparameters are listed below:
\begin{itemize}
    \item Population size $n$: 64
    \item Standard Deviation of Noise $\sigma$: 0.5
    \item Percent of the Population Elites $e$: 0.05
    \item Exponential $\sigma$ decay: True
    \item Maximum Iteration $N_s$: 1000
\end{itemize}


\begin{algorithm}
\caption{Cross Entropy Method}
\label{alg:cem_pseudo}
\begin{algorithmic}[1]
    \STATE Input: Evaluation Function $G$, Function Input $g$, Maximum Iteration $N_s$, Population Size $n$, Standard Deviation of Noise $\sigma$, Percent of the Population Elites $e$.
    \STATE Latent Program Search Center $z_r \gets [z_0, z_1, ..., z_i, ..., z_{255}], z_i \sim \mathcal{N}(0, 1)$
    \STATE $step \gets 0$
    \WHILE{$step < N_s$}
        \STATE Candidate Latent Programs $Z \gets [\ ]$
        \STATE Fitness Scores $L_{G} \gets [\ ]$
        \FOR{$i \gets 1$ to $n$}
            \STATE $ \varepsilon \gets [\varepsilon_0, \varepsilon_1, ..., \varepsilon_i, ..., \varepsilon_{255}], \varepsilon_i \sim \mathcal{N}(0, \sigma)$
            \STATE $Z\text{.append} (z_r + \varepsilon)$
            \STATE $L_{G}\text{.append} (G((z_r + \varepsilon), g))$
        \ENDFOR
        \STATE Elite Latent Programs $Z^{kl}$ $\gets$ Latent Programs in top $e$ percent of $Z$ ranked by $L_G$.
        \STATE $z_r \gets mean(Z^{kl})$
        \STATE $step \gets step+1$
    \ENDWHILE
\end{algorithmic}
\end{algorithm}

\subsection{\textbf{CEM+Diversity}}
\label{app:cem_div}
\begin{figure}
    \centering
    \includegraphics[trim=0 0 0 0,clip,width=0.7\textwidth]{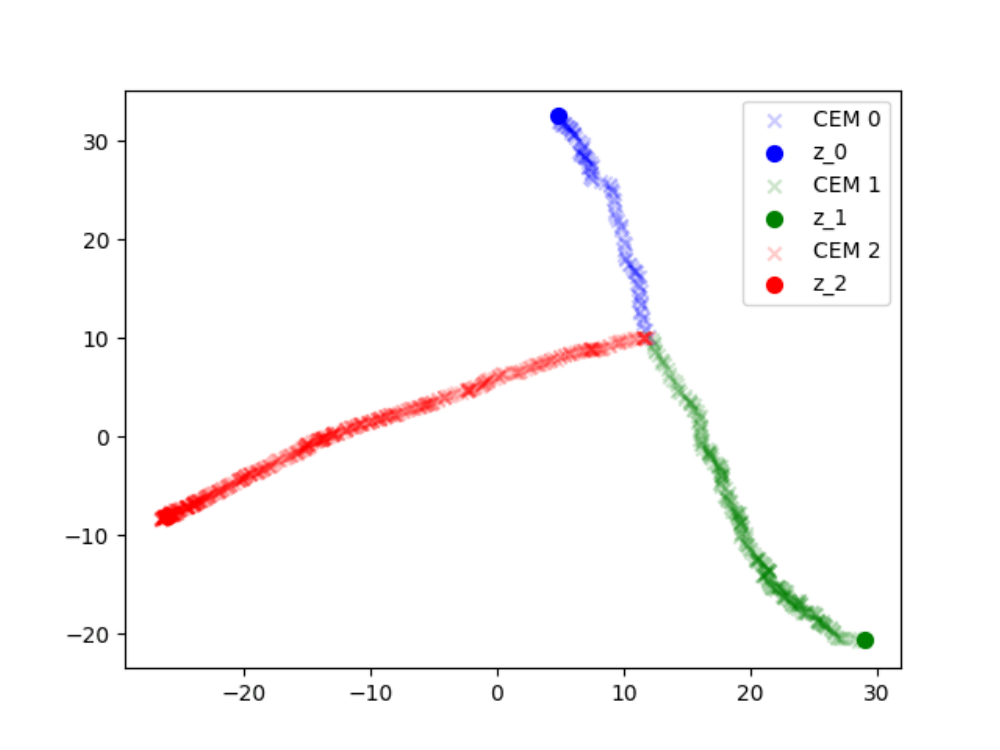}
    \caption[]{
        \small \textbf{CEM+Diversity Searching Trajectories.} It shows the trajectories of the procedure of running CEM+diversity 3 times. The program embeddings searched during the CEM are reduced to 2-dimensional embeddings using PCA. Since the diversity design, the $2nd$ CEM is forced to explore the opposite direction related to the searching path of the $1st$ CEM, and the $3rd$ CEM is urged to search a path that is perpendicular to the $1st$ and $2nd$ searching paths.
        \label{fig:mcem}
    }
\end{figure}

The procedure of running CEM+diversity N times is as follows:
\begin{enumerate}
  \item[(1)] Search the $1st$ program embedding $z_1$ by $CEM(G, g=(Z_k: \{\}))$
  \item[(2)] Search the $2nd$ program embedding $z_2$ by $CEM(G, g=(Z_k: \{z_1\}))$
  \\
   ...
  \item[(N)] Search the $Nth$ program embedding $z_N$ by $CEM(G, g=(Z_k: \{z_1, ..., z_{N-1}\}))$
\end{enumerate}
Here, $Z_k$ is the set of retrieved program embeddings $\{z_i\}_{i=1,\text{...},k-1}$ from the previous $(k-1)$ CEM searches.
The evaluation function is $G(z,Z_k) = (\sum_{t=0}^{T} \gamma^t \mathbb{E}_{(s_t,a_t) \sim \text{EXEC} (\rho_{z})}[r_{t}]) \cdot diversity(z, Z_k)$, where $diversity(z, Z_k) = Sigmoid(-\max_{{z}_i \in {Z}_k}\ \frac{{z} \cdot {z}_i}{\Vert {z} \Vert \Vert {z}_i \Vert})$. 
An example of the search trajectories can be seen in \myfig{fig:mcem}.

\subsection{\textbf{CEM+Diversity+Compatibility}}
\label{app:cem_div_comp}
\subsubsection{\textbf{Sample Program Sequence}}
In the following, we discuss the procedure for sampling a program sequence $\Psi$ from $k$ previously determined mode programs $\rho_{i, i=1,\text{...},k}$ during the search of the $(k+1)$st mode program $\rho_{k+1}$.
\begin{itemize}
\item[(1)] Uniformly sample a program $\rho_j$ from all $k+1$ programs $\{ \rho_1, \text{...}, \rho_k, \rho_{k+1} \}$, and add $\rho_j$ to $\Psi$.
\item[(2)] Repeat (1) until the $(k+1)$st program $\rho_{k+1}$ is sampled.
\item[(3)] Uniformly sample a program $\rho_j$ from $\{ \rho_1, \text{...}, \rho_k, \rho_{k+1},  \rho_{\text{term}} \}$, where $\rho_{\text{term}}$ corresponds to the termination mode $m_{\text{term}}$, and add $\rho_j$ to $\Psi$.
\item[(4)] Repeat (3) until $\rho_{\text{term}}$ is sampled.
\item[(5)] If the length of $\Psi$ is less than 10, than re-sample the $\Psi$ again. Otherwise, return $\Psi$.
\end{itemize}

\subsubsection{\textbf{The Whole Procedure}}
\label{app:whole_pro}
The procedure of running CEM+diversity+Compatibility $|M|$ times in order to retrieve $|M|$ mode programs is as follows:
\begin{itemize}[leftmargin=10mm]
  \item[(1)] Retrieve $1st$ mode program $z_1$.
  \begin{itemize}
  \item[a.] Sample $\Psi_{i=1}$ with $k=0$ 
  \item[b.]
  Run CEM+diversity N times with $Z_k = \{\}$ to get N program embeddings.
  \item[c.]
  Choose the program embedding with the highest $G(z, Z_k=\{\})$ among the N program embeddings as $z_1$.
  \end{itemize}
  
  \item[(2)] Retrieve $2nd$ mode program $z_2$.
  \begin{itemize}
  \item[a.]
  Sample $\Psi_{i=1,2}$ from $k=1$ previously determined mode program.
  \item[b.]
  Run CEM+diversity N times with $Z_k = \{z_1\}$, to get N program embeddings.
  \item[c.]
  Choose the program embedding with the highest $G(z, Z_k=\{z_1\})$ among the N program embeddings as $z_2$.
  \end{itemize}
   ...
  \item[($|M|$)] Retrieve $|M|th$ mode program $z_{|M|}$.
  \begin{itemize}
  \item[a.]
  Sample $\Psi_{i, i=1,...,2^{|M|-1}}$ from $k=|M|-1$ previously determined mode programs.
  \item[b.]
  Run CEM+diversity N times with $Z_k = \{z_1, z_2, ..., z_{|M|-1}\}$, to get N program embeddings.
  \item[c.]
  Choose the program embedding with the highest $G(z, Z_k=\{z_1, z_2, ..., z_{|M|-1}\})$ among the N program embeddings as $z_{|M|}$.
  \end{itemize}
  
\end{itemize}

The evaluation function for \textbf{CEM+Diversity+Compatibility} is 
$G({z}, Z_k) = \frac{1}{D} \sum_{i=1}^{D} R_{\Psi_i} \cdot diversity({z}, Z_k)$, and $R_{\Psi_i}$ can be written as:

\begin{equation}
R_{\Psi_i} = \frac{1}{|\Psi_i|} \sum_{j=1}^{|\Psi_i|} \sum_{t=0}^{T^j} \gamma^t \mathbb{E}_{(s_t,a_t) \sim \text{EXEC} (\Psi_{i}[j])}[r_t]
\end{equation}

where $|\Psi_i|$ is the number of programs in the program list $\Psi_i$, $\Psi_i[j]$ is the $j$-th program in the program list $\Psi_i$, and $\gamma$ is the discount factor.

\section{Program Sample Efficiency}
\label{app:se_exp}
\begin{figure}[h]
    \centering
    \includegraphics[trim=0 0 0 0,clip,width=1\textwidth]{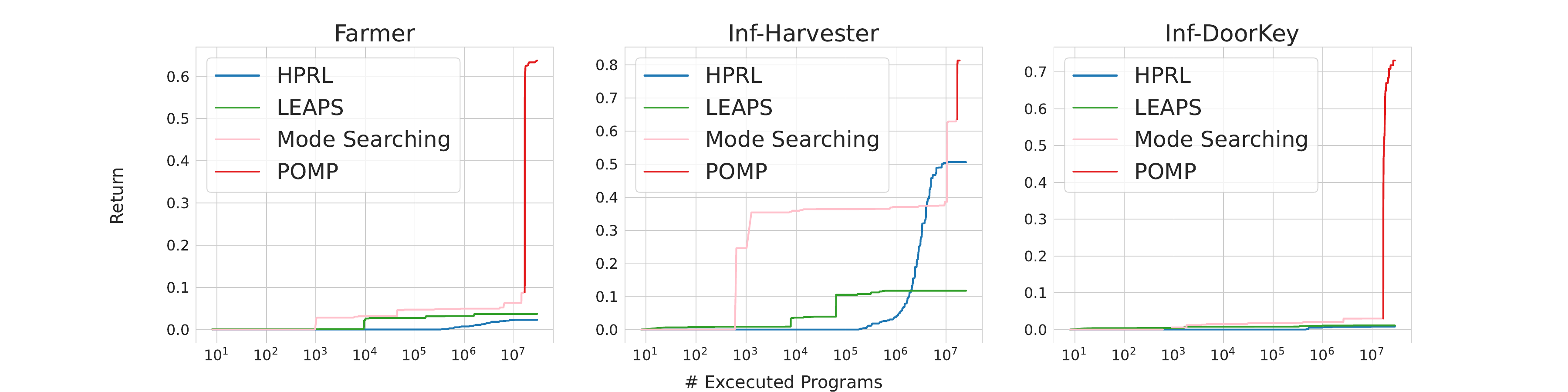}
    \caption[]{
        \small 
        \textbf{Program Sample Efficiency.} Results of different programmatic RL approaches in \textsc{Farmer}, \textsc{Inf-Harvester}, \textsc{Inf-DoorKey}.
        \label{fig:seall}
    }
\end{figure}
Generally, during the training phases of programmatic RL approaches, programs will be synthesized first and then executed in the environment of a given task to evaluate whether these programs are applicable to solve the given task. This three-step procedure (synthesis, execution, and evaluation) will be repeatedly done until the return converges or the maximum training steps are reached. 
The sample efficiency of programmatic RL approaches with programs being the basic step units, dubbed as "Program Sample Efficiency" by us, is therefore important when it comes to measuring the efficiency of the three-step procedure. 
To be more clear, the purpose of this analysis is to figure out how many times the three-step procedure needs to be done to achieve a certain return. As shown in \myfig{fig:seall}, 
POMP has the best program sample efficiencies in \textsc{Farmer} and \textsc{Inf-DoorKey}. 
The details of the return calculation for each approach are described below. 

\subsection{\textbf{POMP}}
During the mode program searching process of POMP, at most 50 CEMs are done to search mode programs. In each CEM, a maximum of 1000 iterations will be carried out, and in each iteration of the CEM, n (population size of the CEM) times of the three-step procedure are done. The return of a certain number of executed programs in the first half of the figure is recorded as the maximum return obtained from executing the previously searched programs in the order that sampled random sequences indicated.

During the transition function training process, the three-step procedure is done once in each PPO training step. The return of a certain number of executed programs in the remainder of the figure is recorded as the maximum validation return obtained by POMP.

\subsection{\textbf{LEAPS}}
During the program searching process of LEAPS, the CEM is used to search the targeted program, and the hyperparameters of the CEM are tuned. A total of 216 CEMs are done, as we follow the procedure proposed in \cite{trivedi2021learning}. In each CEM, a maximum of 1000 iterations will be carried out, and in each iteration of the CEM, n (population size of the CEM) times of the three-step procedure are done. The return of a certain number of executed programs in the figure is recorded as the maximum return obtained from executing the previously searched programs solely.

\subsection{\textbf{HPRL}}
During the meta-policy training process of HPRL, the three-step procedure is done once in each PPO training step. Therefore, with the settings of the experiment described in ~\mysecref{app:hprl_exp}, the three-step procedure will be done 25M times when the training process is finished. The return of a certain number of executed programs in the figure is recorded as the maximum return obtained from the cascaded execution of 5 programs, which are decoded from latent programs output by the meta-policy.

\section{Inductive Generalization}
\label{app:ig_exp}
\begin{figure}
    \centering
    \includegraphics[trim=0 0 0 0,clip,width=1\textwidth]{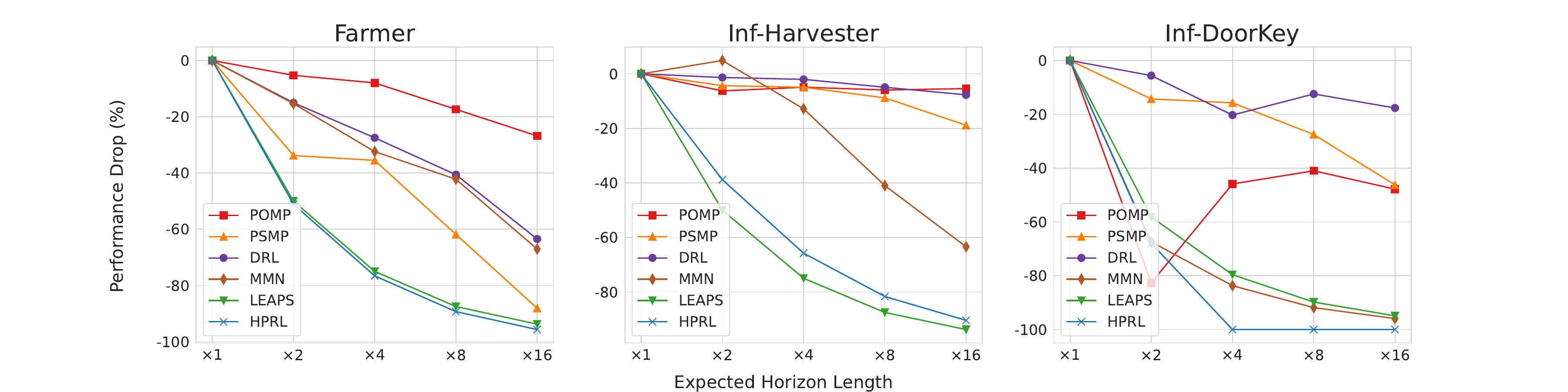}
    \caption[]{
        \small 
        \textbf{Inductive Generalization.} Experiment Results on different baselines in \textsc{Farmer}, \textsc{Inf-Harvester}, and  \textsc{Inf-DoorKey}.
        \label{fig:igall}
    }
\end{figure}



To test the ability of each method when it comes to inductively generalizing, we scale up the expected horizon of the environment by increasing the upper limit of the target for each \textsc{Karel-Long} task solely during the testing phase. To elaborate more clearly, using \textsc{Farmer} as an example, the upper limit number in this task is essentially the maximum iteration number of the filling-and-collecting rounds that we expect the agent to accomplish (more details of the definition of the maximum iteration number can be found in \mysecref{app:karel_long_problem_set}). In the original task settings, the agent is asked to continuously placing and then picking markers in a total of 10 rounds of the filling-and-collecting process. That means, all policies across every method are trained to terminate after 10 rounds of the placing-and-picking markers process is finished. Nevertheless, the upper limit number is set to 20, 40, etc, in the testing environment to test the generalization ability of each policy.

Since most of the baselines don't perform well on \textsc{Seesaw}, \textsc{Up-N-Down}, and \textsc{Inf-Doorkey} (\ie more than half of baseline approaches have mean return close to $0.0$ on these tasks), we do the inductive generalization experiments mainly on 
\textsc{Farmer} and \textsc{Inf-Harvester}.
The expected horizon lengths of the testing environments will be 2, 4, 8, and 16 times longer compared to the training environment settings, respectively. Also, rewards gained from picking or placing markers and the penalty for actions are divided by 2, 4, 8, and 16, respectively, so that the maximum total reward of each task is normalized to 1.
The detailed setting and the experiment results for each of these three tasks are shown as follows.

\subsection{\textsc{\textbf{Farmer}}}
During the training phases of our and other baseline methods, we set the maximum iteration number to 10. However, we adjusted this number for the testing phase to 20, 40, 80, and 160. As shown in \myfig{fig:ig}, when the expected horizon length grows, the performances of all the baselines except POMP drop dramatically, indicating that our method has a much better inductive generalization property on this task. More details of the definition of the maximum iteration number can be found in \mysecref{app:karel_long_problem_set}.

\subsection{\textsc{\textbf{Inf-Harvester}}}
During the training phases of our and other baseline methods, we set the emerging probability
to $\frac{1}{2}$. However, we adjusted this number for the testing phase to $\frac{3}{4}$, $\frac{7}{8}$, $\frac{15}{16}$ and $\frac{31}{32}$. As shown in \myfig{fig:ig}, when the expected horizon length grows, the performances of POMP, PSMP, and DRL drop slightly, but the performances of MMN, LEAPS, and HPRL drop extensively. More details of the definition of the emerging probability can be found in \mysecref{app:karel_long_problem_set}.


\section{State Machine Extraction}
\label{app:fsm_et}
In our approach, since we employ the neural network transition function, the proposed Program Machine Policy is only partially or locally interpretable -- once the transition function selects a mode program, human users can read and understand the following execution of the program.

\begin{figure}[b]
    \centering
    \includegraphics[trim=0 0 0 0,clip,width=1\textwidth]{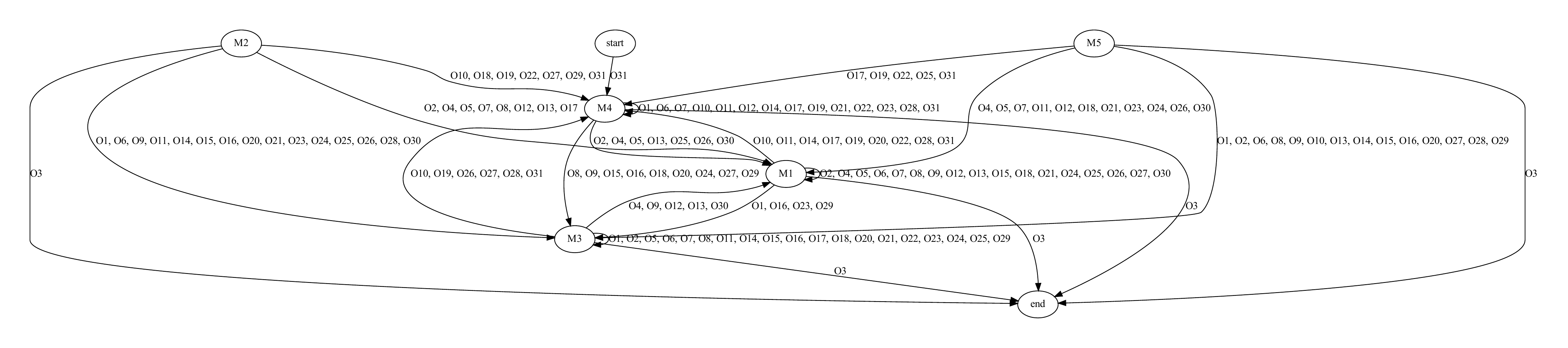}
    \caption[]{
        \small 
        \textbf{Example of extracted state machine on \textsc{Farmer}}. $O1$ to $O31$ represent the unique quantized vectors encoded from observations. The corresponding mode programs of $M1$ to $M5$ are displayed in \myfig{fig:karel_program_examples_farmer}.
        \label{fig:fsm_fm}
    }
\end{figure}

\begin{figure}
    \centering
    \includegraphics[trim=0 0 0 0,clip,width=0.4\textwidth]{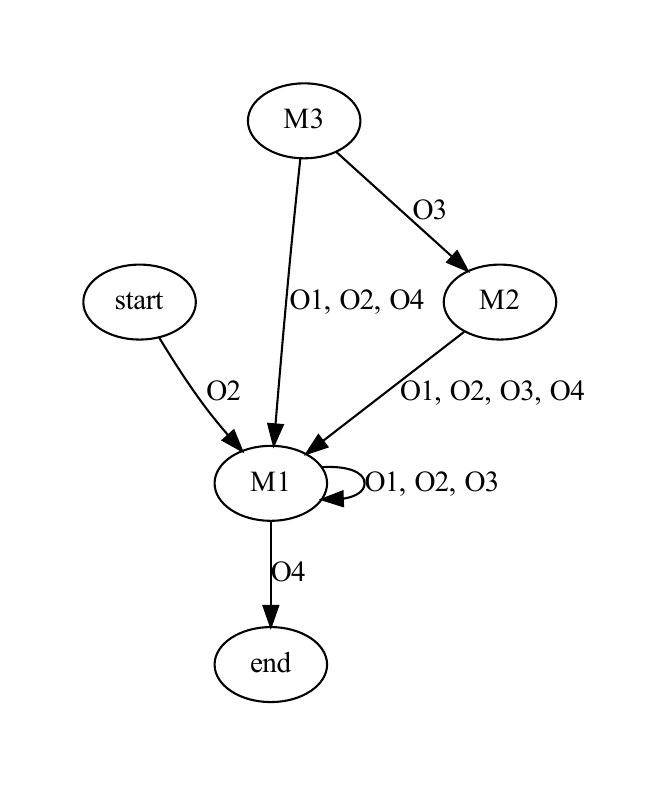}
    \caption[]{
        \small 
        \textbf{Example of extracted state machine on \textsc{Inf-Harvester}}. $O1$ to $O4$ represent the unique quantized vectors encoded from observations. The corresponding mode programs of $M1$ to $M3$ are displayed in \myfig{fig:karel_program_examples_harvester}.
        \label{fig:fsm_ih}
    }
\end{figure}

\begin{figure}
    \centering
    \includegraphics[trim=0 0 0 0,clip,width=0.7\textwidth]{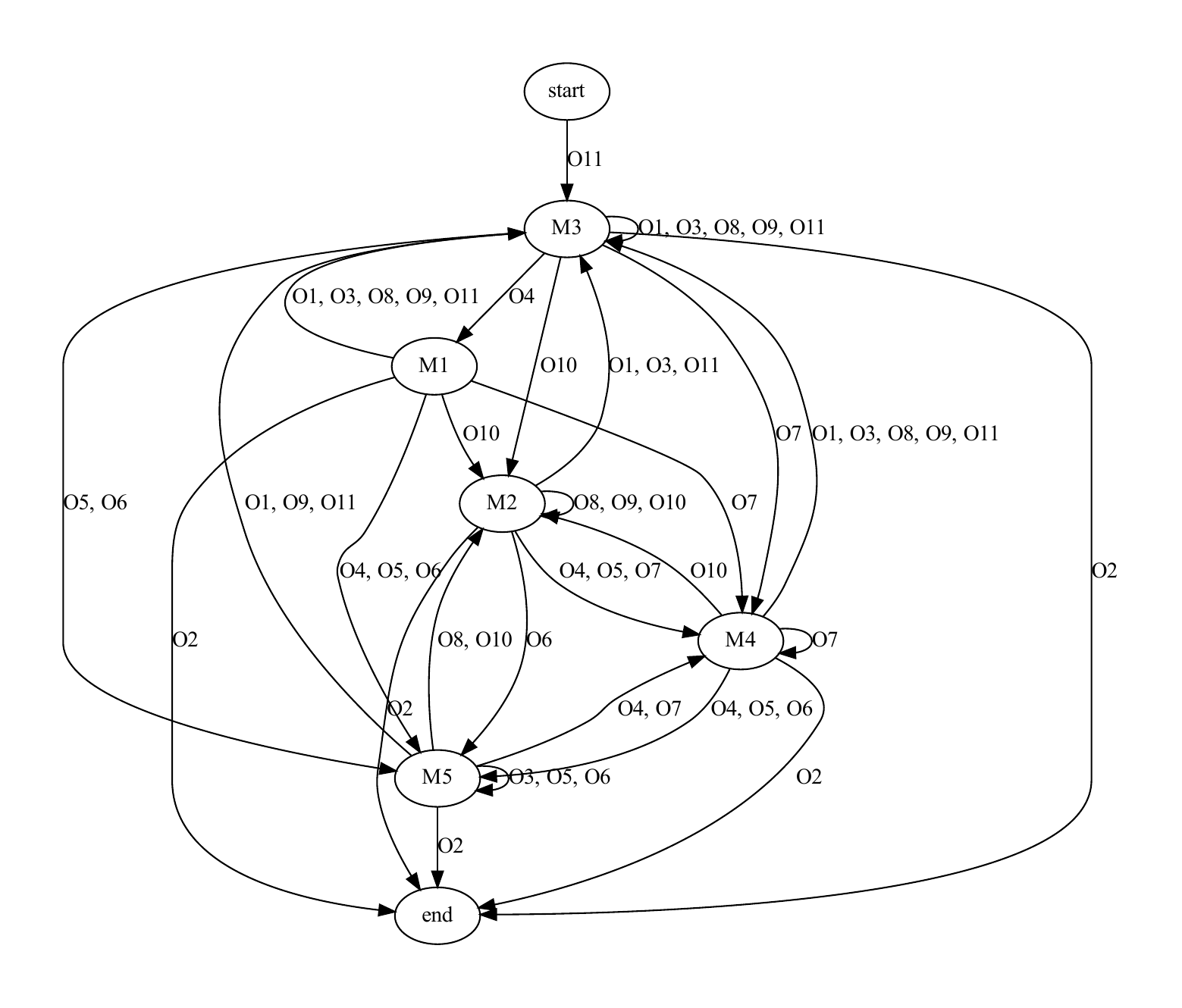}
    \caption[]{
        \small 
        \textbf{Example of extracted state machine on \textsc{Inf-Doorkey}}. $O1$ to $O11$ represent the unique quantized vectors encoded from observations. The corresponding mode programs of $M1$ to $M5$ are displayed in \myfig{fig:karel_program_examples_infDoorKey}.
        \label{fig:fsm_id}
    }
\end{figure}

To further increase the interpretability of the trained mode transition function $f$, we extracted the state machine structure by the approach proposed in ~\citep{koul2018learning}.  In this setup, since POMP utilizes the previous mode as one of the inputs and predicts the next mode, we focus solely on encoding the state observations. Each state observation is extracted by convolutional neural networks and fully connected layers to a $1 \times 32$ vector, which is then quantized into a $1 \times h$ vector, where $h$ is a hyperparameter to balance between finite state machine simplicity and performance drop. We can construct a state-transition table using these quantized vectors and modes. The final step involves minimizing these quantized vectors, which allows us to represent the structure of the state machine effectively. Examples of extracted state machine are shown in \myfig{fig:fsm_fm}, \myfig{fig:fsm_id} and \myfig{fig:fsm_ih}.


\section{Hyperparameters and Settings of Experiments}
\label{app:hyper_settings_exp}
\subsection{\textbf{POMP}}
\label{app:pomp_exp}

\subsubsection{Encoder \& Decoder} 
\label{app:POMP_encoder_decoder_training}
We follow the training procedure and the model structure proposed in \cite{trivedi2021learning}, which uses the GRU~\cite{cho2014learning} network to implement both the encoder $q_{\phi}$ and the decoder $p_{\theta}$ with hidden dimensions of 256. The encoder $q_{\phi}$ and decoder $p_{\theta}$ are trained on programs randomly sampled from the Karel DSL. The loss function for training the encoder-decoder model includes the $\beta$-VAE~\cite{higgins2016beta} loss, the program behavior reconstruction loss~\cite{trivedi2021learning}, and the latent behavior reconstruction loss~\cite{trivedi2021learning}.

The program dataset used to train $q_{\phi}$ and $p_{\theta}$ consists of 35,000 programs for training and 7,500 programs for validation and testing. We sequentially sample program tokens for each program based on defined probabilities until an ending token or when the maximum program length of 40 is reached. The defined probability of each kind of token is listed below:
\begin{itemize}
    \item \texttt{WHILE}: 0.15
    \item \texttt{REPEAT}: 0.03
    \item \texttt{STMT\_STMT}: 0.5
    \item \texttt{ACTION}: 0.2
    \item \texttt{IF}: 0.08
    \item \texttt{IFELSE}: 0.04 
\end{itemize}

Note that, the token \texttt{STMT\_STMT} represents dividing the current token into two separate tokens, each chosen based on the same probabilities defined above. This token primarily dictates the lengths of the programs, as well as the quantity and complexity of nested loops and statements.

\subsubsection{Mode Program Synthesis}
We conduct the procedure described in ~\mysecref{app:whole_pro} to search modes for each Program Machine Policy. For tasks that only need skills for picking markers and traversing, i.e., \textsc{Inf-Harvester}, \textsc{Seesaw}, and \textsc{Up-N-Down}, we set $|M|=3$. For the rest, we set $|M|=5$.


\begin{figure}
\centering
    \includegraphics[trim=0 0 10 0,clip,width=0.6\columnwidth]{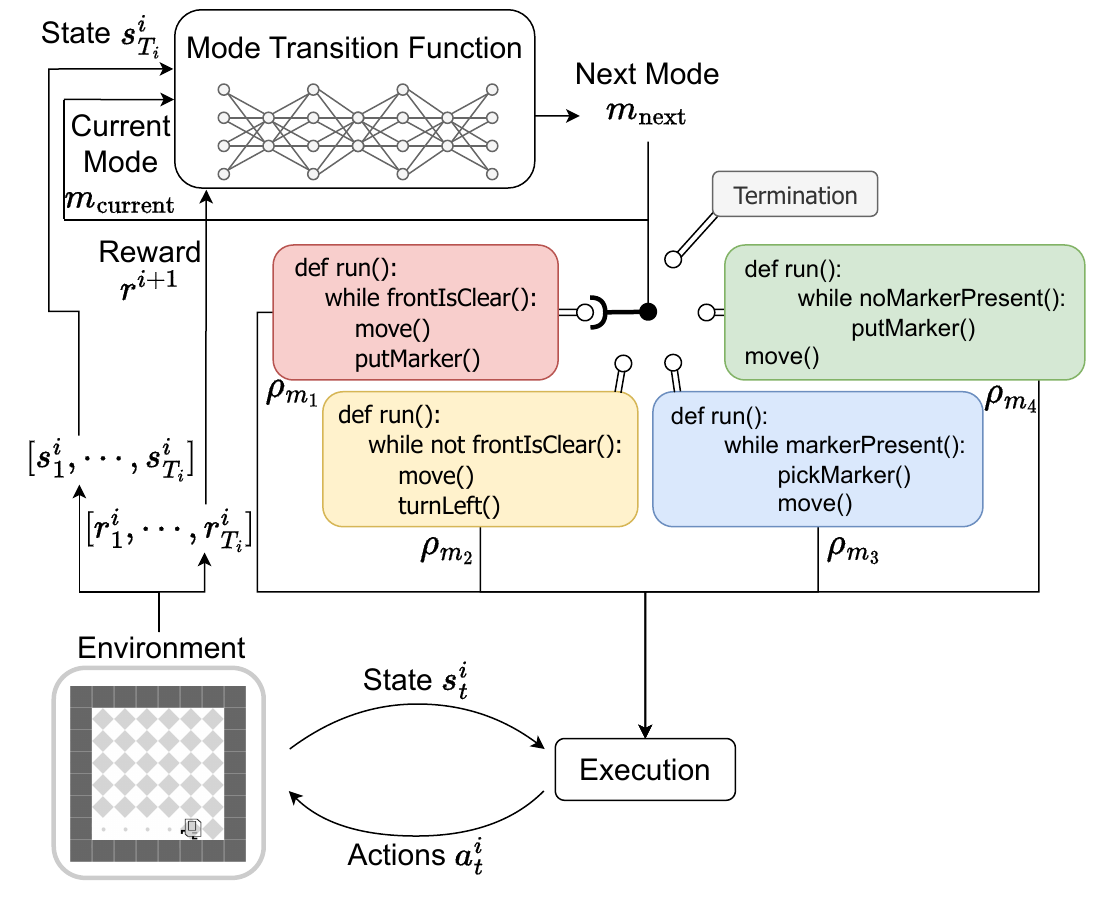}
    \vspace{-0.2cm}
    \caption[]{
        \textbf{Program Machine Policy Execution.}
        \label{fig:stage2_mode_trans}
    }
    \vspace{-0.2cm}
\end{figure}

\subsubsection{Mode Transition Function}
\label{app:mode_transition}
The mode transition function $f$ 
consists of convolutional layers~\citep{fukushima1982neocognitron, krizhevsky2017imagenet} to derive features from the Karel states and the fully connected layers to predict the transition probabilities among each mode. Meanwhile, we utilize one-hot encoding to represent the current mode index of the Program Machine Policy. The detail setting of the convolutional layers is the same as those described in ~\mysecref{app:drl_exp}. The training process of the mode transition function $f$ is illustrated in \myfig{fig:stage2_mode_trans} and it can be optimized using the PPO~\citep{schulman2017proximal} algorithm. 
The hyperparameters are listed below:
\begin{itemize}
    \item Maximum program number: 1000
    \item Batch size : 32
    \item Clipping: 0.05
    \item $\alpha$: 0.99
    \item $\gamma$: 0.99
    \item GAE lambda: 0.95
    \item Value function coefficient: 0.5
    \item Entropy coefficient: 0.1
    \item Number of updates per training iteration: 4
    \item Number of environment steps per set of training iterations: 32
    \item Number of parallel actors: 32
    \item Optimizer : Adam
    \item Learning rate: \{0.1, 0.01, 0.001, 0.0001, 0.00001\}
\end{itemize}

\subsection{\textbf{PSMP}}
\label{app:pomp_prim}
It resembles the setting in ~\mysecref{app:pomp_exp}. The input, output, and structure of the mode transition function $f$ remain the same. However, the 5 modes programs are replaced by 5 primitive actions (\texttt{move}, \texttt{turnLeft}, \texttt{turnRight}, \texttt{putMarker}, \texttt{pickMarker}).

\subsection{\textbf{DRL}}
\label{app:drl_exp}
DRL training on the Karel environment uses the PPO~\citep{schulman2017proximal} algorithm with 20 million timesteps. Both the policies and value networks share a convolutional encoder that interprets the state of the grid world. This encoder comprises two layers: the initial layer has 32 filters, a kernel size of 4, and a stride of 1, while the subsequent layer has 32 filters, a kernel size of 2, and maintains the same stride of 1. The policies will predict the probability distribution of primitive actions (move, turnLeft, turnRight, putMarker, pickMarker) and termination. During our experiments with DRL on \textsc{Karel-Long} tasks, we fixed most of the hyperparameters and did hyperparameters grid search over learning rates. The hyperparameters are listed below:
\begin{itemize}
    \item Maximum horizon: 50000
    \item Batch size : 32
    \item Clipping: 0.05
    \item $\alpha$: 0.99
    \item $\gamma$: 0.99
    \item GAE lambda: 0.95
    \item Value function coefficient: 0.5
    \item Entropy coefficient: 0.1
    \item Number of updates per training iteration: 4
    \item Number of environment steps per set of training iterations: 128
    \item Number of parallel actors: 32
    \item Optimizer : Adam
    \item Learning rate: \{0.1, 0.01, 0.001, 0.0001, 0.00001\}
\end{itemize}

\subsection{\textbf{MMN}}
\label{app:mmn_exp}
Aligned with the approach described in~\cite{koul2018learning}, we trained and quantized a recurrent policy with a GRU cell and convolutional neural network layers to extract information from gird world states. During our experiments with MMN on \textsc{Karel-Long} tasks, we fixed most of the hyperparameters and did a hyperparameter grid search over learning
rates. The hyperparameters are listed below:
\begin{itemize}
    \item Hidden size of GRU cell: 32
    \item Number of quantized bottleneck units for observation: 128
    \item Number of quantized bottleneck units for hidden state: 16
    \item Maximum horizon: 50000
    \item Batch size : 32
    \item Clipping: 0.05
    \item $\alpha$: 0.99
    \item $\gamma$: 0.99
    \item GAE lambda: 0.95
    \item Value function coefficient: 0.5
    \item Entropy coefficient: 0.1
    \item Number of updates per training iteration: 4
    \item Number of environment steps per set of training iterations: 128
    \item Number of parallel actors: 32
    \item Optimizer : Adam
    \item Learning rate: \{0.1, 0.01, 0.001, 0.0001, 0.00001\}
\end{itemize}

\subsection{\textbf{LEAPS}}
\label{app:leaps_exp}
In line with the setup detailed in ~\cite{trivedi2021learning}, we conducted experiments over various hyperparameters of the CEM to optimize rewards for LEAPS. The hyperparameters are listed below:
\begin{itemize}
    \item Population size (n): \{8, 16, 32, 64\}
    \item $\sigma$: \{0.1, 0.25, 0.5\}
    \item $e$: \{0.05, 0.1, 0.2\}
    \item Exponential $\sigma$ decay: \{True, False\}
    \item Initial distribution $\mathcal{P}$ : \{$\mathcal{N} (1, 0)$, $\mathcal{N} (0, \sigma)$, $\mathcal{N} (0, 0.1\sigma)$\}
\end{itemize}

\subsection{\textbf{HPRL}}
\label{app:hprl_exp}
In alignment with the approach described in~\cite{liu2023hierarchical}, we trained the meta policy for each task to predict a program sequence. The hyperparameters are listed below:
\begin{itemize}
    \item Max subprogram: 5
    \item Max subprogram Length: 40
    \item Batch size : 128
    \item Clipping: 0.05
    \item $\alpha$: 0.99
    \item $\gamma$: 0.99
    \item GAE lambda: 0.95
    \item Value function coefficient: 0.5
    \item Entropy coefficient: 0.1
    \item Number of updates per training iteration: 4
    \item Number of environment steps per set of training iterations: 32
    \item Number of parallel actors: 32
    \item Optimizer : Adam
    \item Learning rate: 0.00001
    \item Training steps: 25M
\end{itemize}

\section{Details of \textsc{Karel} Problem Set}
\label{app:karel_problem_set}
\begin{figure*}[ht]
    \centering
    \begin{subfigure}[b]{0.95\textwidth}
    \centering
    \includegraphics[trim=0 0 0 0, clip, width=\textwidth]{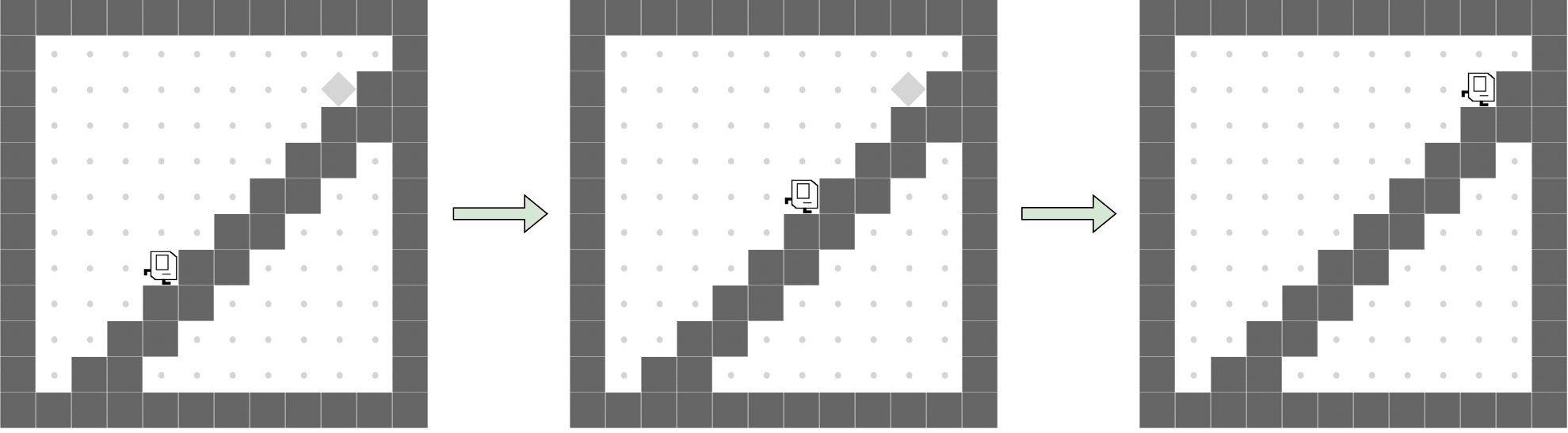}
    \caption{\textsc{StairClimber}}
    \end{subfigure}
    \hspace{0.5cm}
    \begin{subfigure}[b]{0.95\textwidth}
    \centering
    \includegraphics[trim=0 0 0 0, clip,width=\textwidth]{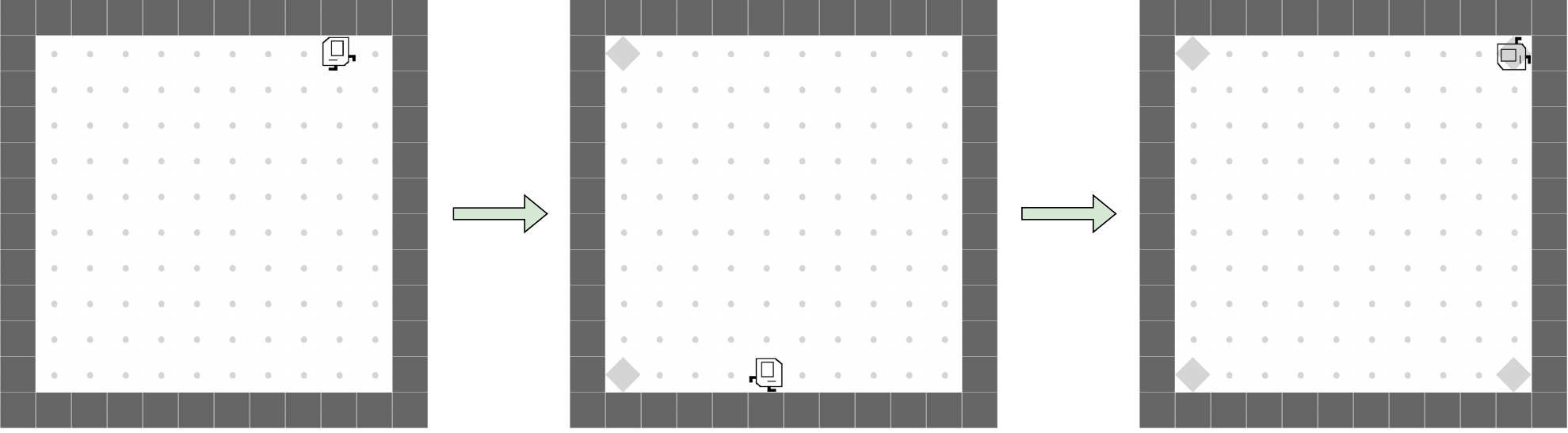}
    \caption{\textsc{FourCorner}}
    \end{subfigure}
    \hspace{0.5cm}
    \begin{subfigure}[b]{0.95\textwidth}
    \centering
    \includegraphics[trim=0 0 0 0, clip,width=\textwidth]{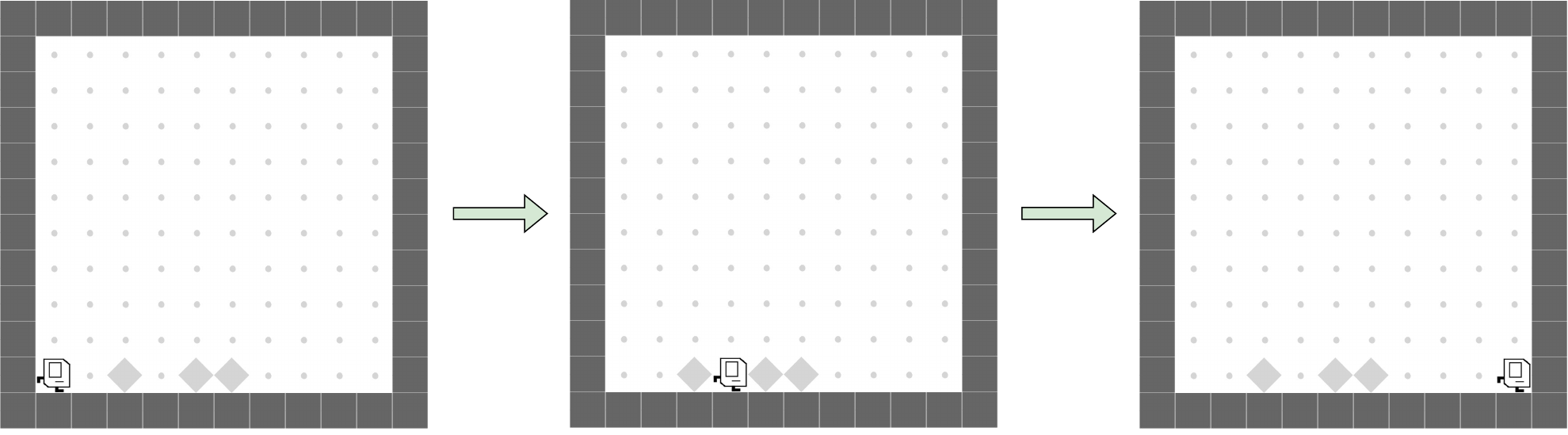}
    \caption{\textsc{TopOff}}
    \end{subfigure}
    \hspace{0.5cm}
    \begin{subfigure}[b]{0.95\textwidth}
    \centering
    \includegraphics[trim=0 0 0 0, clip,width=\textwidth]{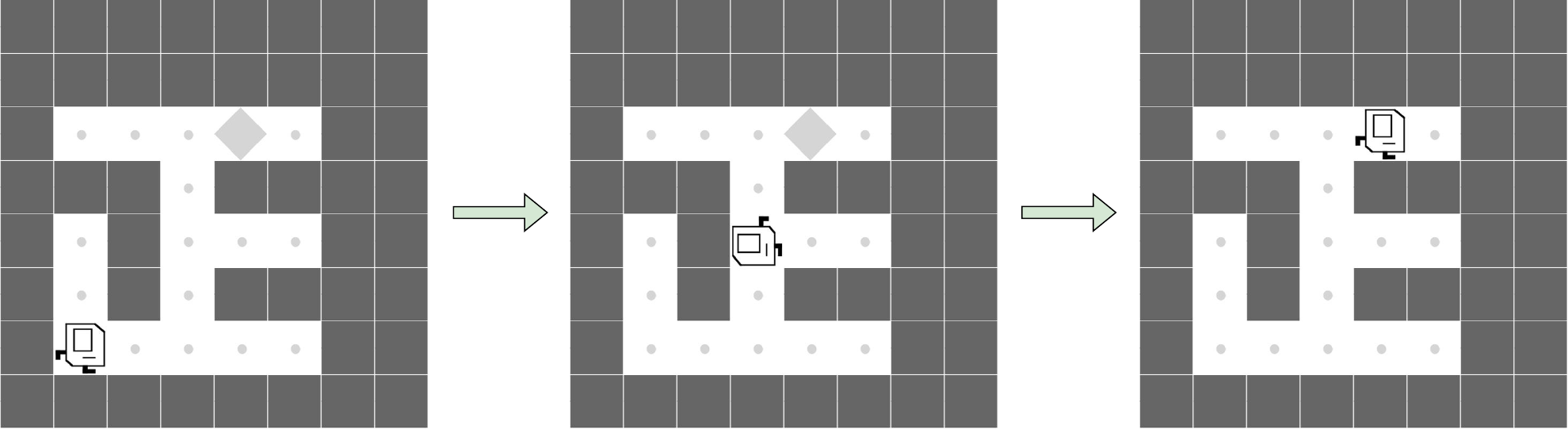}
    \caption{\textsc{Maze}}
    \end{subfigure}
    \caption[]{
    Visualization of \textsc{StairClimber}, \textsc{FourCorner}, \textsc{TopOff}, and \textsc{Maze} in the \textsc{Karel} problem set presented in \citet{trivedi2021learning}. For each task, a random initial state, a legitimate internal state, and the ideal end state are shown. In most tasks, the position of markers and the initial location of the Karel agent are randomized. More details of the \textsc{Karel} problem set can be found in \mysecref{app:karel_problem_set}.
    }
    \label{fig:KarelFig1}
\end{figure*}

\begin{figure*}[ht]
    \centering
    \begin{subfigure}[b]{1\textwidth}
    \centering
    \includegraphics[trim=0 0 0 0, clip,width=\textwidth]{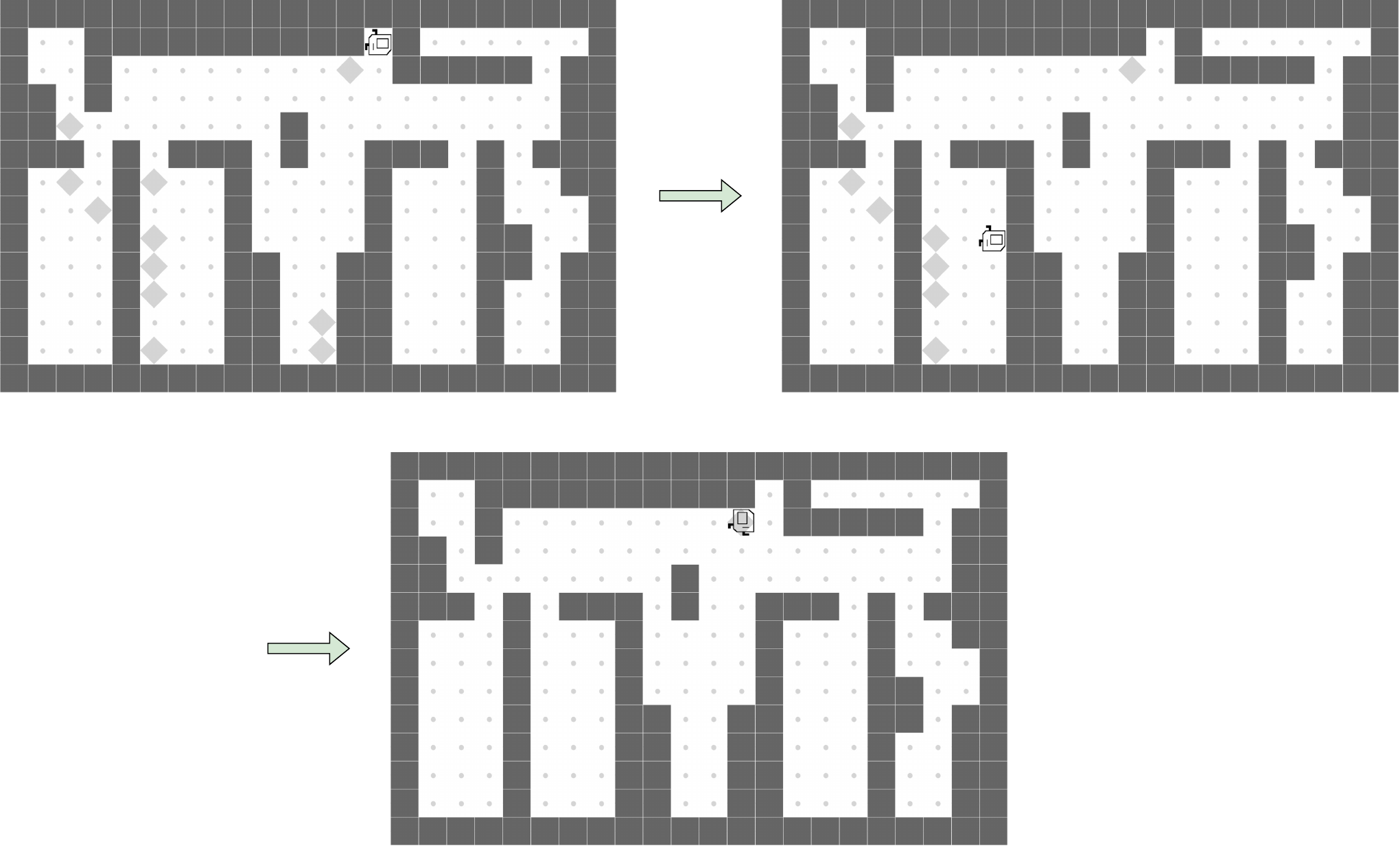}
    \caption{\textsc{CleanHouse}}
    \end{subfigure}
    \hspace{0.5cm}
    \begin{subfigure}[b]{1\textwidth}
    \centering
    \includegraphics[trim=0 0 0 0, clip,width=\textwidth]{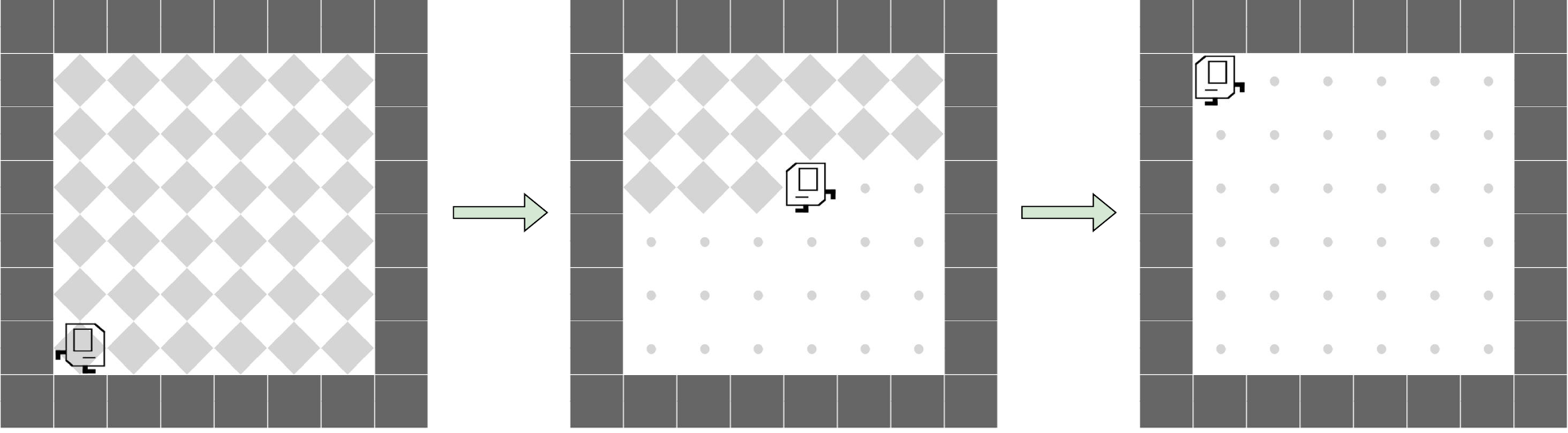}
    \caption{\textsc{Harvester}}
    \end{subfigure}
    \caption[]{
    Visualization of \textsc{CleanHouse} and \textsc{Harvester} in the \textsc{Karel} problem set presented in \citet{trivedi2021learning}. For each task, a random initial state, a legitimate internal state, and the ideal end state are shown. More details of the \textsc{Karel} problem set can be found in \mysecref{app:karel_problem_set}.
    }
    \label{fig:KarelFig2}
\end{figure*}

The \textsc{Karel} problem set is presented in \citet{trivedi2021learning}, consisting of the following tasks: \textsc{StairClimber}, \textsc{FourCorner}, \textsc{TopOff}, \textsc{Maze}, \textsc{CleanHouse} and \textsc{Harvester}. \myfig{fig:KarelFig1} and \myfig{fig:KarelFig2} provide visual depictions of a randomly generated initial state, an internal state sampled from a legitimate trajectory, and the desired final state for each task. The experiment results presented in \mytable{tab:karel_Karel_hard_POMP_main} are evaluated by averaging the rewards obtained from $32$ randomly generated initial configurations of the environment. 


\subsection{\textsc{\textbf{StairClimber}}}
This task takes place in a $12\times12$ grid environment, where the agent's objective is to successfully climb the stairs and reach the marked grid. The marked grid and the agent's initial location are both randomized at certain positions on the stairs, with the marked grid being placed on the higher end of the stairs. The reward is defined as $1$ if the agent reaches the goal in the environment, $-1$ if the agent moves off the stairs, and $0$ otherwise.



\subsection{\textsc{\textbf{FourCorner}}}

This task takes place in a $12\times12$ grid environment, where the agent's objective is to place a marker at each of the four corners. The reward received by the agent will be $0$ if any marker is placed on the grid other than the four corners. Otherwise, the reward is calculated by multiplying $0.25$ by the number of corners where a marker is successfully placed.

\subsection{\textsc{\textbf{TopOff}}}
This task takes place in a $12\times12$ grid environment, where the agent's objective is to place a marker on every spot where there's already a marker in the environment's bottom row. The agent should end up in the rightmost square of this row when the rollout concludes. The agent is rewarded for each consecutive correct placement until it either misses placing a marker where one already exists or places a marker in an empty grid on the bottom row.

\subsection{\textsc{\textbf{Maze}}}
This task takes place in an $8\times8$ grid environment, where the agent's objective is to find a marker by navigating the grid environment. The location of the marker, the initial location of the agent, and the configuration of the maze itself are all randomized. The reward is defined as $1$ if the agent successfully finds the marker in the environment, $0$ otherwise.

\subsection{\textsc{\textbf{CleanHouse}}}
This task takes place in a $14\times22$ grid environment, where the agent's objective is to collect as many scattered markers as possible. The initial location of the agent is fixed, and the positions of the scattered markers are randomized, with the additional condition that they will only randomly be scattered adjacent to some wall in the environment. The reward is defined as the ratio of the collected markers to the total number of markers initially placed in the grid environment.

\subsection{\textsc{\textbf{Harvester}}}
This task takes place in an $8\times8$ grid environment, where the environment is initially populated with markers appearing in all grids. The agent's objective is to pick up a marker from each location within this grid environment. The reward is defined as the ratio of the picked markers to the total markers in the initial environment.


\section{Details of \textsc{Karel-Hard} Problem Set}
\label{app:karel_hard_problem_set}
\begin{figure*}[ht]
    \centering
    \begin{subfigure}[b]{1\textwidth}
    \centering
    \includegraphics[trim=0 0 0 0, clip, width=\textwidth]{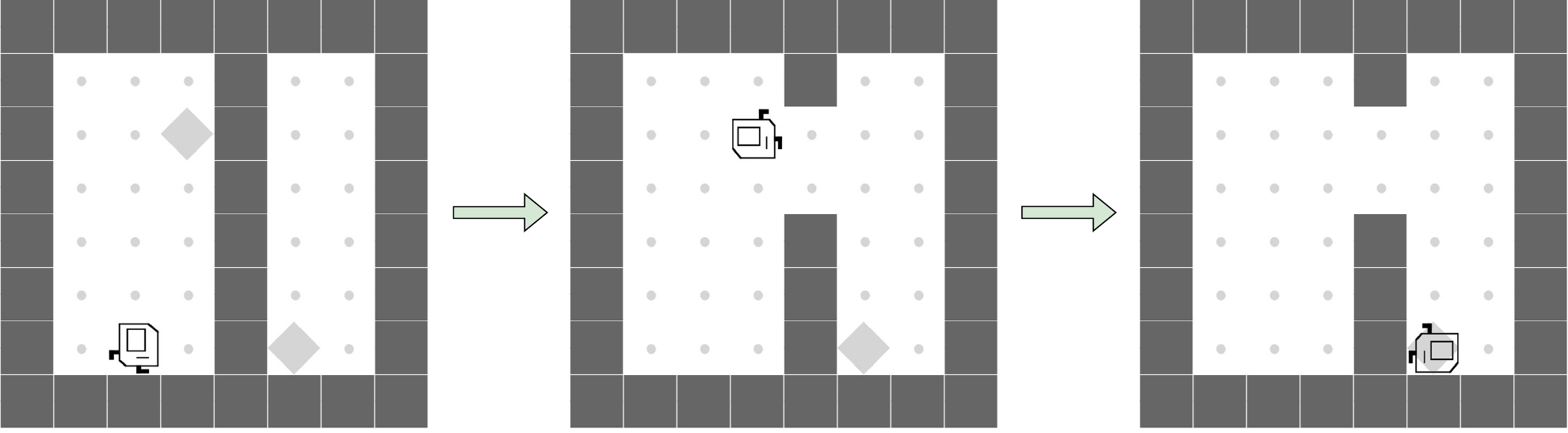}
    \caption{\textsc{DoorKey}}
    \end{subfigure}
    \hspace{0.5cm}
    \begin{subfigure}[b]{1\textwidth}
    \centering
    \includegraphics[trim=0 0 0 0, clip,width=\textwidth]{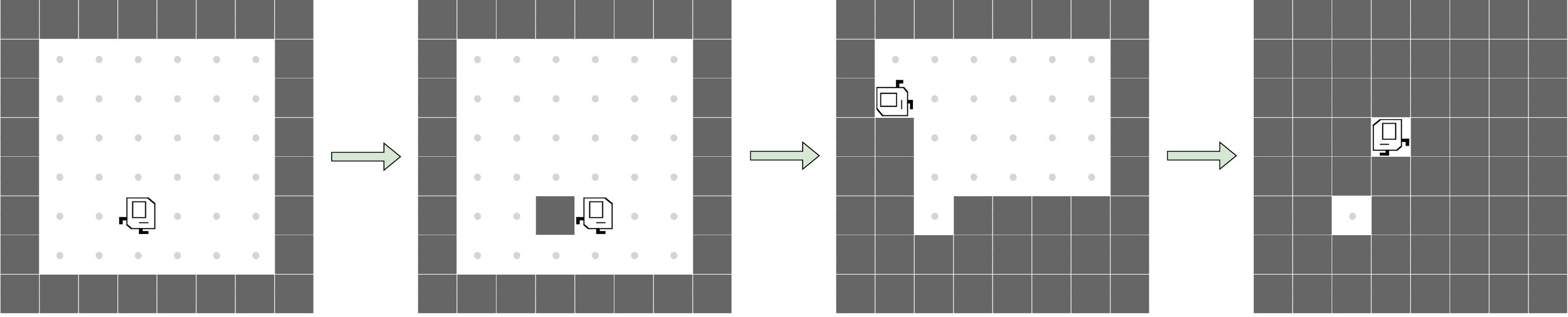}
    \caption{\textsc{OneStroke}}
    \end{subfigure}
    \hspace{0.5cm}
    \begin{subfigure}[b]{1\textwidth}
    \centering
    \includegraphics[trim=0 0 0 0, clip,width=\textwidth]{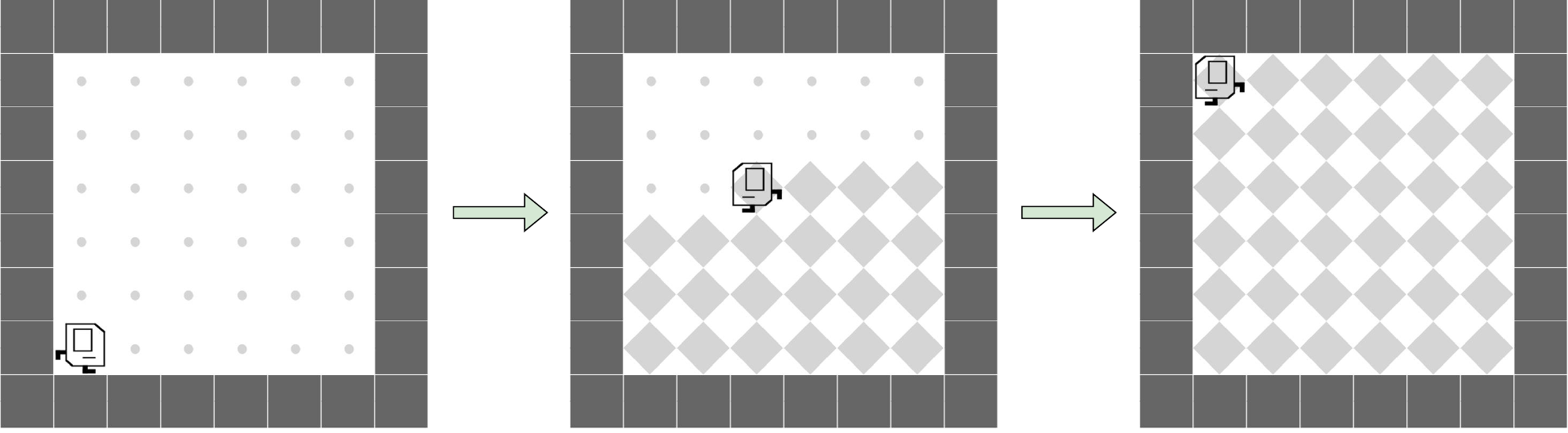}
    \caption{\textsc{Seeder}}
    \end{subfigure}
    \hspace{0.5cm}
    \begin{subfigure}[b]{1\textwidth}
    \centering
    \includegraphics[trim=0 0 0 0, clip,width=\textwidth]{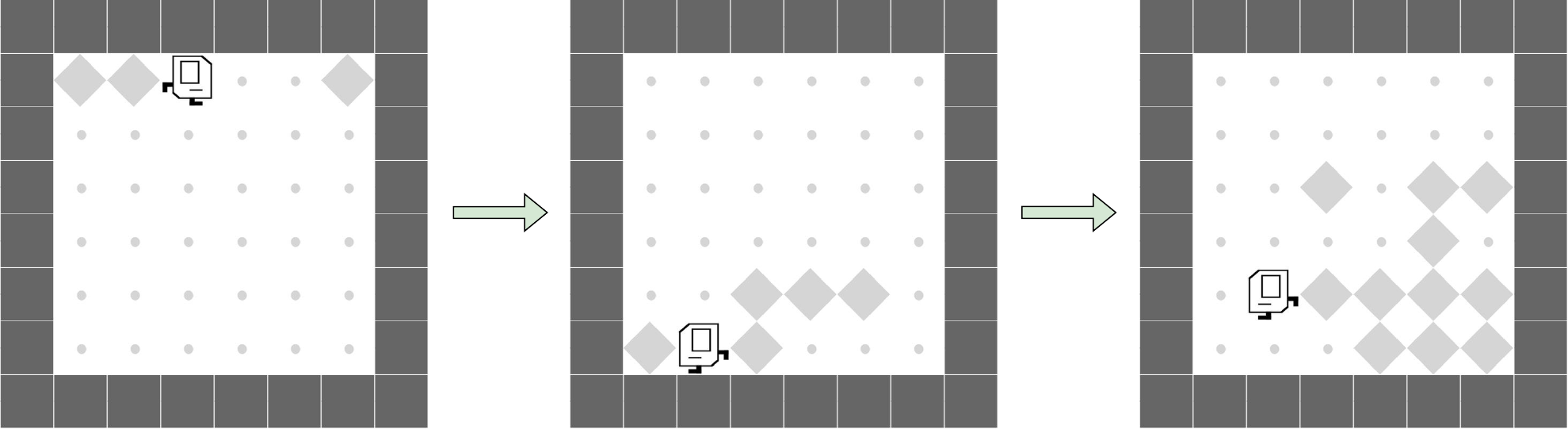}
    \caption{\textsc{Snake}}
    \end{subfigure}
    \caption[]{
    Visualization of each task in the \textsc{Karel-Hard} problem set proposed by \citet{liu2023hierarchical}. For each task, a random initial state, some legitimate internal state(s), and the ideal end state are shown. More details of the \textsc{Karel-Hard} problem set can be found in \mysecref{app:karel_hard_problem_set}.
    }
    \label{fig:KarelHardFig1}
\end{figure*}

The \textsc{Karel-Hard} problem set proposed by \citet{liu2023hierarchical} consists of the following tasks: \textsc{DoorKey}, \textsc{OneStroke}, \textsc{Seeder} and \textsc{Snake}. Each task in this benchmark is designed to have more constraints and be more structurally complex than tasks in the \textsc{Karel} problem set. \myfig{fig:KarelHardFig1} provides a visual depiction of a randomly generated initial state, some internal state(s) sampled from a legitimate trajectory, and a final state for each task. The experiment results presented in \mytable{tab:karel_Karel_hard_POMP_main} are evaluated by averaging the rewards obtained from $32$ randomly generated initial configurations of the environment. 


\subsection{\textsc{\textbf{DoorKey}}}
This task takes place in an $8\times8$ grid environment, where the grid is partitioned into a $6\times3$ left room and a $6\times2$ right room. Initially, these two rooms are not connected. The agent's objective is to collect a key (marker) within the left room to unlock a door (make the two rooms connected) and subsequently position the collected key atop a target (marker) situated in the right room. The agent's initial location, the key's location, and the target's location are all randomized. The agent receives a $0.5$ reward for collecting the key and another $0.5$ reward for putting the key on top of the target.

\subsection{\textsc{\textbf{OneStroke}}}
This task takes place in an $8\times8$ grid environment, where the agent's objective is to navigate through all grid cells without revisiting any of them. Once a grid cell is visited, it transforms into a wall. If the agent ever collides with these walls, the episode ends. The reward is defined as the ratio of grids visited to the total number of empty grids in the initial environment.

\subsection{\textsc{\textbf{Seeder}}}
This task takes place in an $8\times8$ grid environment, where the agent's objective is to place a marker on every single grid. If the agent repeatedly puts markers on the same grid, the episode will then terminate. The reward is defined as the ratio of the number of markers successfully placed to the total number of empty grids in the initial environment.

\subsection{\textsc{\textbf{Snake}}}
This task takes place in an $8\times8$ grid environment, where the agent plays the role of the snake's head and aims to consume (pass-through) as much food (markers) as possible while avoiding colliding with its own body. Each time the agent consumes a marker, the snake's body length grows by $1$, and a new marker emerges at a different location. Before the agent successfully consumes $20$ markers, exactly one marker will consistently exist in the environment. The reward is defined as the ratio of the number of markers consumed by the agent to $20$.

\section{Details of \textsc{Karel-Long} Problem Set}
\label{app:karel_long_problem_set}


\begin{figure}
    \centering
    \includegraphics[trim=0 0 0 0,clip,width=1\textwidth]{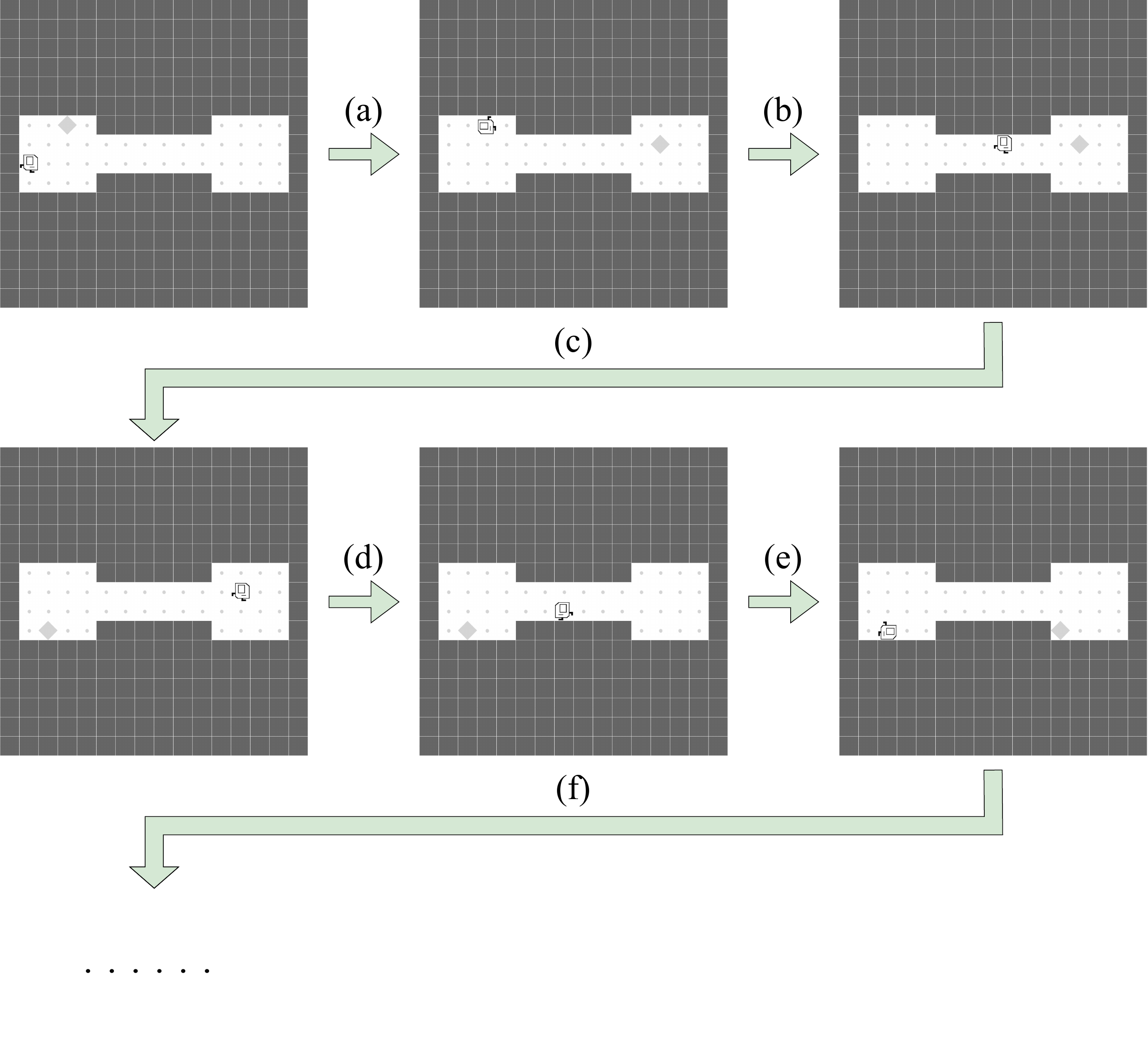}
    \caption[]{
        \textbf{Visualization of \textsc{Seesaw} in the \textsc{Karel-Long} problem set.} This partially shows a typical trajectory of the Karel agent during the task \textsc{Seesaw}. 
        (a): Once the Karel agent collects a marker in the left chamber, a new marker will appear in the right chamber.
        (b): The agent must navigate through a middle corridor to collect the marker in the right chamber.
        (c): Upon the Karel agent collecting a marker in the right chamber, a new marker further appears in the left chamber.
        (d): Once again, the agent is traversing through the corridor to the left chamber.
        (e): A new marker appears in the right chamber again after the agent picks up the marker in the left chamber.
        (f): The agent will move back and forth between the two chambers to collect the emerging markers continuously. Note that the locations of all the emerging markers are randomized. Also, note that we have set the number of emerging markers to 64 during the training phase, meaning the agent has to pick up 64 markers to fully complete the task.
         More details of the task \textsc{Seesaw} can be found in \mysecref{app:karel_long_problem_set}.
        \label{fig:Karel_Long_Seesaw_Example}
    }
\end{figure}

\begin{figure}
    \centering
    \includegraphics[trim=0 0 0 0,clip,width=0.95\textwidth]{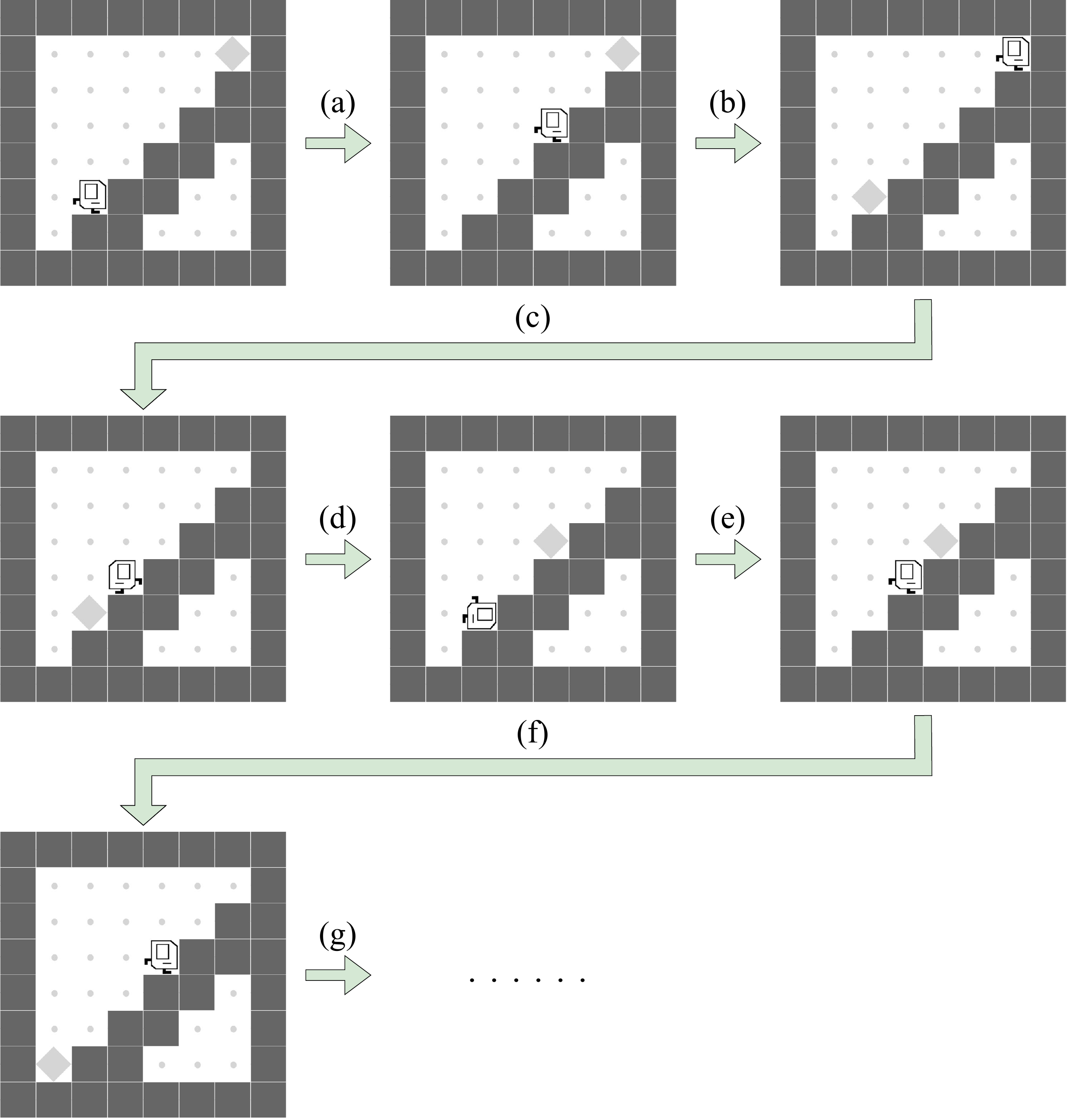}
    \caption[]{
        \textbf{Visualization of \textsc{Up-N-Down} in the \textsc{Karel-Long} problem set.} This partially shows a typical trajectory of the Karel agent during the task \textsc{Up-N-Down}. 
        (a): The Karel agent is ascending the stairs to collect a load located above the stairs. Note that the agent can theoretically collect the load without directly climbing up the stairs, but it will receive some penalties if it does so.
        (b): Once the agent collects the load, a new load appears below the stairs.
        (c): The agent is descending the stairs to collect a load located below the stairs. Still, note that the agent can theoretically collect the load without directly climbing down the stairs, but it will receive some penalties if it does so.
        (d): Upon the agent collecting the load, a new load appears above the stairs.
        (e): The agent is again ascending the stairs to collect a load.
        (f): A new load appears below the stairs again after the agent collects the load located above the stairs.
        (g): The agent will then descend and ascend the stairs repeatedly to collect the emerging loads. Note that the locations of all the emerging loads are randomized right near the stairs, and they will always appear above or below the stairs, depending on the position of the agent. Also, note that we have set the number of emerging loads to 100 during the training phase, meaning the agent has to collect 100 loads to complete the task fully.
         More details of the task \textsc{Up-N-Down} can be found in \mysecref{app:karel_long_problem_set}.
        \label{fig:Karel_Long_upNdown_Example}
    }
\end{figure}

\begin{figure}
    \centering
    \includegraphics[trim=0 0 0 0,clip,width=1\textwidth]{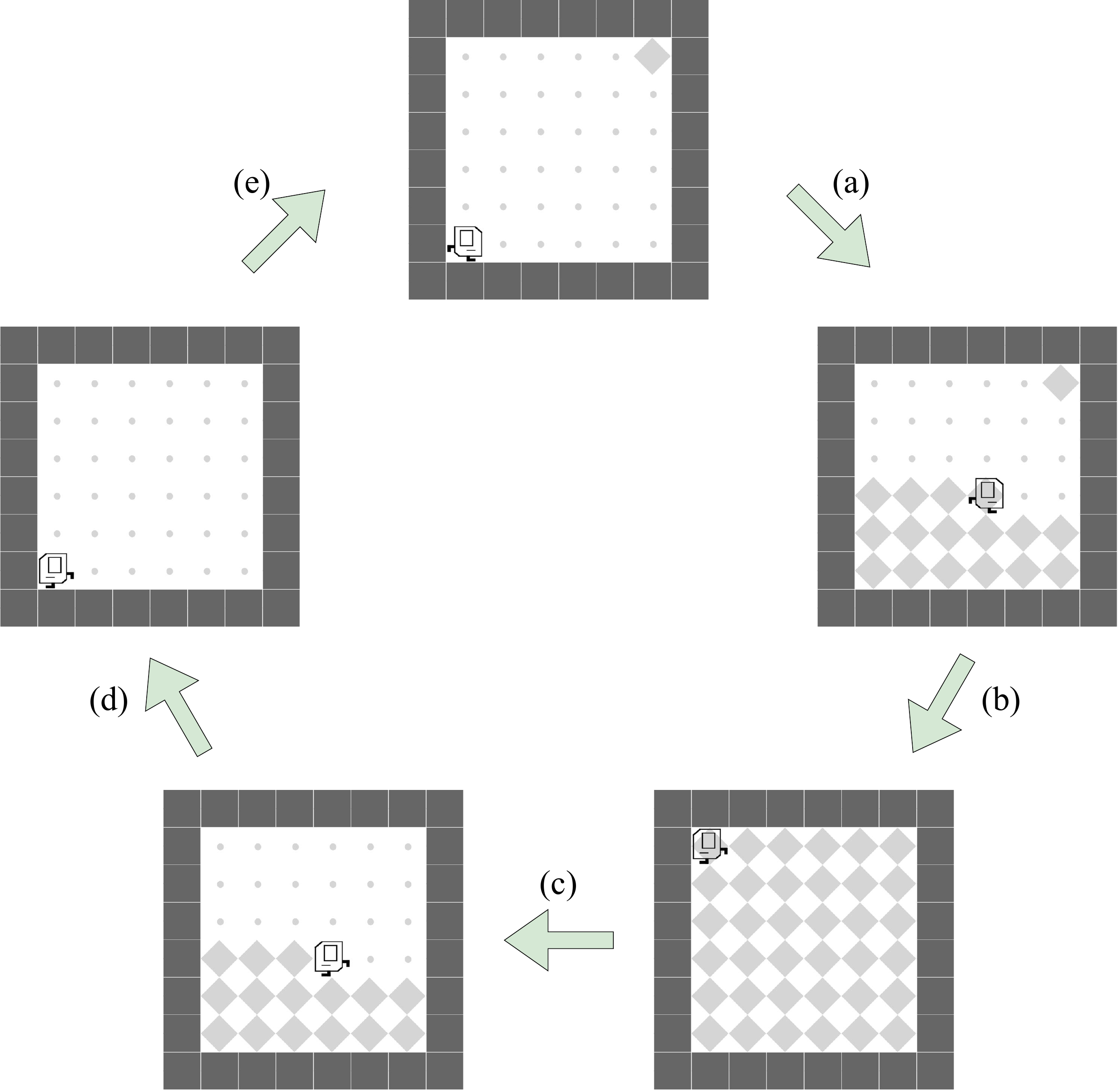}
    \caption[]{
        \textbf{Visualization of \textsc{Farmer} in the \textsc{Karel-Long} problem set.} This partially shows a typical trajectory of the Karel agent during the task \textsc{Farmer}. 
        (a): The Karel agent is filling (placing) the entire environment layout with markers. Note that, in the initial state, there is a single marker located in the upper-right corner. The marker is designed to be a signal indicating the agent to start filling the environment layout.
        (b): The agent successfully populates the entire environment. 
        (c): The agent is then asked to pick up markers as much as possible.
        (d): The agent successfully picks all markers up, leaving the environment empty.
        (e): If there is another filling-and-collecting round, a marker will appear in the upper-right corner to indicate that the agent should start the filling process again. Otherwise, the agent completes the entire task, and no further marker will appear. For simplicity, here, we only show the former case. Note that we have set the number of max markers to 720 during the training phase, meaning the agent has to fill the entire environment layout with markers and pick up all markers twice to fully complete the task. 
        More details of the task \textsc{Farmer} can be found in \mysecref{app:karel_long_problem_set}.
        \label{fig:Karel_Long_Farmer_Example}
    }
\end{figure}

\begin{figure}
    \centering
    \includegraphics[trim=0 0 0 0,clip,width=0.9\textwidth]{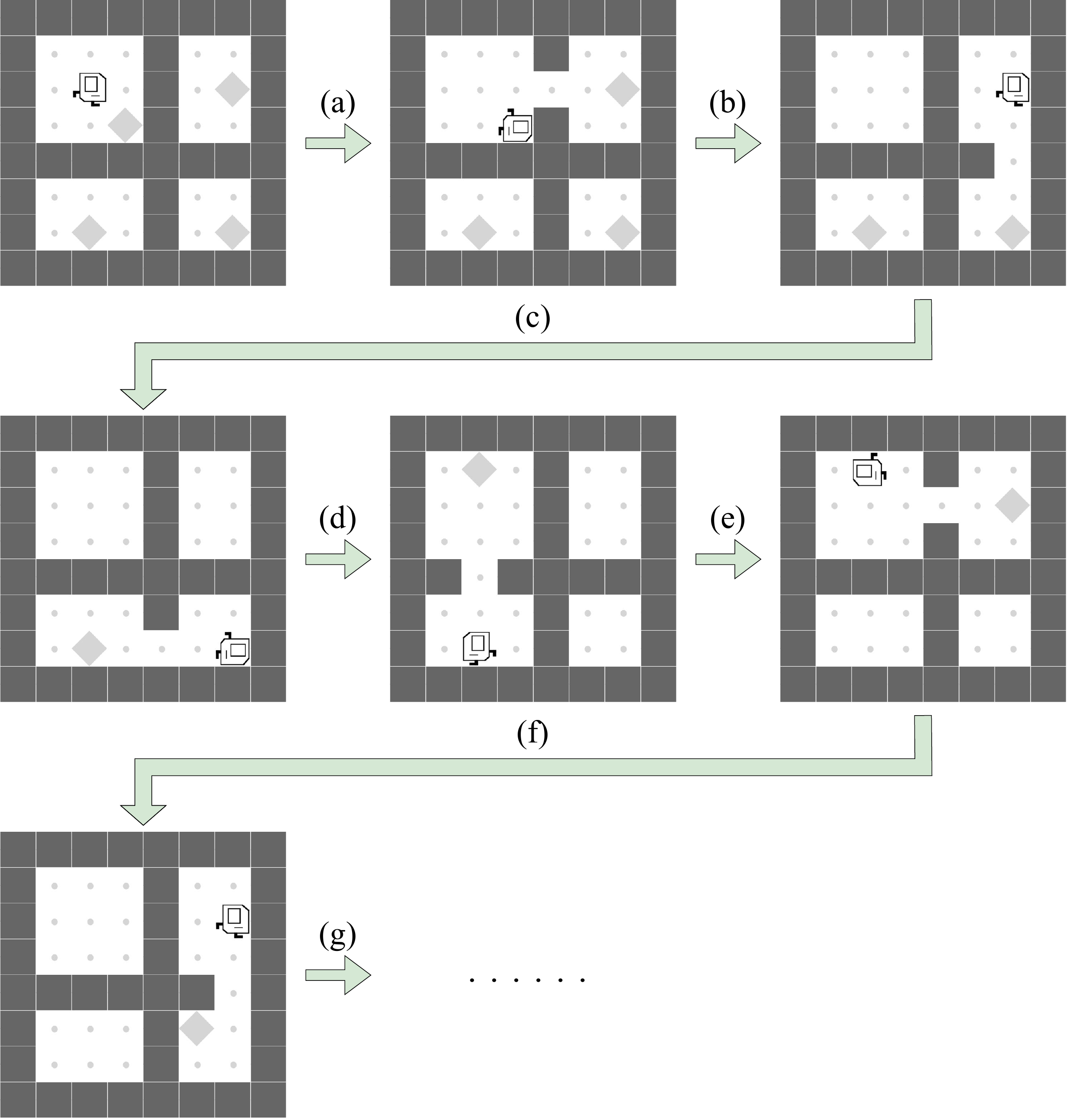}
    \caption[]{
        \textbf{Visualization of \textsc{Inf-Doorkey} in the \textsc{Karel-Long} problem set.} This partially shows a typical trajectory of the Karel agent during the task \textsc{Inf-Doorkey}. 
        (a): The Karel agent picks up a marker in the upper-left chamber. Then, a passage to the upper-right chamber opens, allowing the agent to traverse through.
        (b): The agent successfully places a marker at a marked grid located in the upper-right chamber. Subsequently, a passage to the lower-right chamber opens, allowing the agent to traverse through.
        (c): After the agent collects a marker in the lower-right chamber, a passage to the lower-left chamber opens, allowing the agent to traverse through.
        (d): The agent properly places a marker at a marked grid located in the lower-left chamber. After that, a passage to the upper-left chamber opens, and a new marker appears in the upper-left chamber.
        (e): Upon the agent picking up a marker in the upper-left chamber, the passage to the upper-right chamber opens again, and a grid is marked randomly in the upper-right chamber.
        (f): The agent accurately places a marker at a marked grid located in the upper-right chamber. Afterward, the passage to the lower-right chamber opens again, and a new marker emerges in the lower-right chamber.
        (g): The agent will repeatedly pick up and place markers in this fashion until the number of max keys is reached. We have set the number of max keys to 100 during the training phase, meaning the agent has to pick up and place 100 markers in total to fully complete the task.
         More details of the task \textsc{Inf-Doorkey} can be found in \mysecref{app:karel_long_problem_set}.
        \label{fig:Karel_Long_InfDoorkey_Example}
    }
\end{figure}

\begin{figure}
    \centering
    \includegraphics[trim=0 0 0 0,clip,width=1\textwidth]{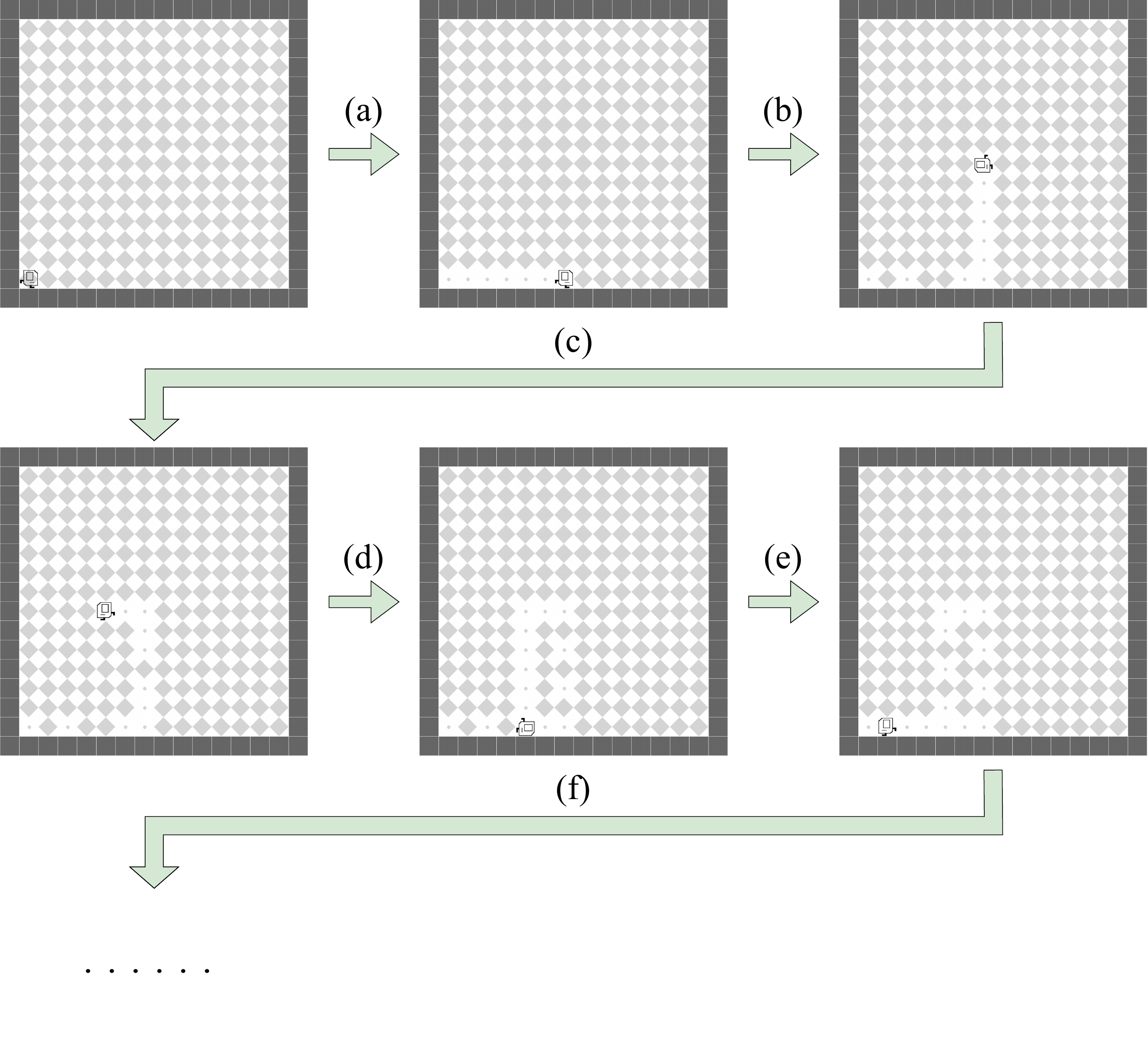}
    \caption[]{
        \textbf{Visualization of \textsc{Inf-Harvester} in the \textsc{Karel-Long} problem set.} This partially shows a legitimate trajectory of the Karel agent during the task \textsc{Inf-Harvester}. 
        (a): The Karel agent is picking up markers in the last row. Meanwhile, no new markers are popped out in the last row.
        (b): The agent turns left and picks up 6 markers in the $7^{th}$ column. During this picking-up process, 3 markers appeared in 3 previously empty grids in the last row.
        (c): The agent is collecting markers in the $8^{th}$ row. During this picking-up process, 1 marker appeared in a previously empty grid in the $7^{th}$ column.
        (d): The agent picks up 6 markers in the $5^{th}$ column. During this picking-up process, 2 markers appeared in 2 previously empty grids in the $7^{th}$ column.
        (e): The agent picks up 2 more markers in the last row. During this picking-up process, 2 markers appeared in 2 previously empty grids in the $5^{th}$ column.
        (f): Since markers will appear in previously empty grids based on the emerging probability, the agent will continuously and indefinitely collect markers until no markers are left, and no more new markers are further popped out in the environment. Note that we have set the emerging probability to $\frac{1}{2}$ during the training phase.
         More details of the task \textsc{Inf-Harvester} can be found in \mysecref{app:karel_long_problem_set}.
        \label{fig:Karel_Long_InfHarvester_Example}
    }
\end{figure}

We introduce a newly designed \textsc{Karel-Long} problem set as a benchmark to evaluate the capability of POMP. Each task is designed to possess long-horizon properties based on the Karel states. Besides, we design the tasks in our \textsc{Karel-Long} benchmark to have a constant per-action cost (i.e., $0.0001$). \myfig{fig:Karel_Long_Seesaw_Example},  \myfig{fig:Karel_Long_upNdown_Example}, \myfig{fig:Karel_Long_Farmer_Example}, \myfig{fig:Karel_Long_InfDoorkey_Example}, and \myfig{fig:Karel_Long_InfHarvester_Example} provide visual depictions of all the tasks within the \textsc{Karel-Long} problem set. For each task, a randomly generated initial state and some internal states sampled from a legitimate trajectory are provided.




\subsection{\textsc{\textbf{Seesaw}}}
This task takes place in a $16\times16$ grid environment, where the agent's objective is to move back and forth between two $4\times4$ chambers, namely the left chamber and the right chamber, to continuously collect markers. To facilitate movement between the left and right chambers, the agent must traverse through a middle $2\times6$ corridor. Initially, exactly one marker is randomly located in the left chamber, awaiting the agent to collect. Once a marker is picked in a particular chamber, another marker is then randomly popped out in the other chamber, further waiting for the agent to collect. Hence, the agent must navigate between the two chambers to pick markers continuously. The reward is defined as the ratio of the number of markers picked by the agent to the total number of markers that the environment is able to generate (dubbed as "emerging markers").


\subsection{\textsc{\textbf{Up-N-Down}}}
This task takes place in an $8\times8$ grid environment, where the agent's objective is to ascend and descend the stairs repeatedly to collect markers (loads) appearing both above and below the stairs. Once a marker below (above) the stairs is picked up, another marker will appear above (below) the stairs, enabling the agent to continuously collect markers. If the agent moves to a grid other than those right near the stairs, the agent will receive a constant penalty (i.e., $0.005$). The reward is defined as the ratio of the number of markers picked by the agent to the total number of markers that the environment is able to generate (dubbed as "emerging loads").

\subsection{\textsc{\textbf{Farmer}}}
This task takes place in an $8\times8$ grid environment, where the agent's objective is to repeatedly fill the entire environment layout with markers and subsequently collect all of these markers. Initially, all grids in the environment are empty except for the one in the upper-right corner. The marker in the upper-right corner is designed to be a signal that indicates the agent needs to start populating the environment layout with markers (analogous to the process of a farmer sowing seeds). After most of the grids are placed with markers, the agent is then asked to pick up markers as much as possible (analogous to the process of a farmer harvesting crops). Then, the agent is further asked to fill the environment again, and the whole process continues in this back-and-forth manner. We have set a maximum iteration number to represent the number of the filling-and-collecting rounds that we expect the agent to accomplish. The reward is defined as the ratio of the number of markers picked and placed by the agent to the total number of markers that the agent is theoretically able to pick and place (dubbed as "max markers").

\subsection{\textsc{\textbf{Inf-Doorkey}}}
This task takes place in an $8\times8$ grid environment, where the agent's objective is to pick up a marker in certain chambers, place a marker in others, and continuously traverse between chambers until a predetermined upper limit number of marker-picking and marker-placing that we have set is reached. The entire environment is divided into four chambers, and the agent can only pick up (place) markers in one of these chambers. Once the agent does so, the passage to the next chamber opens, allowing the agent to proceed to the next chamber to conduct another placement (pick-up) action. The reward is defined as the ratio of markers picked and placed by the agent to the total number of markers that the agent can theoretically pick and place (dubbed as "max keys").


\subsection{\textsc{\textbf{Inf-Harvester}}}
This task takes place in a $16\times16$ grid environment, where the agent's objective is to continuously pick up markers until no markers are left, and no more new markers are further popped out in the environment. Initially, the environment is entirely populated with markers. Whenever the agent picks up a marker from the environment, there is a certain probability (dubbed as "emerging probability") that a new marker will appear in a previously empty grid within the environment, allowing the agent to collect markers both continuously and indefinitely. The reward is defined as the ratio of the number of markers picked by the agent to the expected number of total markers that the environment can generate at a certain probability.

\begin{figure*}[ht]
\centering
\begin{mdframed}[frametitle=Karel Programs]
\begin{subfigure}[t]{0.90\textwidth}

\centering
\textbf{\textsc{Seesaw}}

{
\begin{subfigure}[t]{0.32\textwidth}
Mode 1
\begin{lstlisting}
DEF run m( 
    move 
    move 
    pickMarker 
    turnLeft 
    WHILE c( frontIsClear c) w( 
        move 
        pickMarker 
        w) 
    WHILE c( frontIsClear c) w( 
        move 
        pickMarker 
        w) 
    m)
\end{lstlisting}
\end{subfigure}
\begin{subfigure}[t]{0.32\textwidth}
Mode 2
\begin{lstlisting}
DEF run m( 
    WHILE c( markersPresent c) w( 
        pickMarker 
        turnRight 
        w) 
    WHILE c( markersPresent c) w( 
        pickMarker 
        w) 
    WHILE c( markersPresent c) w( 
        pickMarker 
        w) 
    m)
\end{lstlisting}
\end{subfigure}
\begin{subfigure}[t]{0.32\textwidth}
Mode 3
\begin{lstlisting}
DEF run m( 
    IF c( frontIsClear c) i( 
        turnLeft 
        move 
        move 
        pickMarker 
        turnLeft 
        IF c( frontIsClear c) i( 
            move 
            pickMarker 
            move 
            pickMarker 
            i) 
        move 
        i) 
    IF c( frontIsClear c) i( 
        pickMarker 
        move 
        pickMarker 
        i) 
    move 
    pickMarker 
    m)
\end{lstlisting}
\end{subfigure}
}



\end{subfigure}

\begin{subfigure}[t]{0.90\textwidth}

\centering
\textbf{\textsc{UP-N-Down}}

{
\begin{subfigure}[t]{0.32\textwidth}
Mode 1
\begin{lstlisting}
DEF run m( 
    IF c( frontIsClear c) i( 
        move 
        move 
        i) 
    IF c( not c( leftIsClear c) c) i( 
        move 
        move 
        i) 
    m)
\end{lstlisting}
\end{subfigure}
\begin{subfigure}[t]{0.32\textwidth}
Mode 2
\begin{lstlisting}
DEF run m( 
    turnLeft 
    move 
    turnRight 
    move 
    m)
\end{lstlisting}
\end{subfigure}
\begin{subfigure}[t]{0.32\textwidth}
Mode 3
\begin{lstlisting}
DEF run m( 
    turnLeft 
    move 
    turnRight 
    move 
    turnLeft 
    move 
    turnRight 
    move 
    turnLeft 
    move 
    turnRight 
    move 
    turnLeft 
    move 
    turnRight 
    move 
    turnLeft 
    move 
    turnRight 
    move 
    turnLeft 
    move 
    turnRight 
    move 
    turnLeft 
    move 
    turnRight 
    move 
    turnLeft 
    move 
    turnRight 
    move 
    turnLeft 
    move 
    turnRight 
    move 
    turnLeft 
    move 
    turnRight 
    move 
    m)
\end{lstlisting}
\end{subfigure}
}



\end{subfigure}


\end{mdframed}
\caption[\method\\ Karel Tasks Synthesized Programs]{\textbf{Example programs on Karel-Long tasks: \textsc{Seesaw} and \textsc{Up-N-Down}.} The programs with best rewards out of all random seeds are shown. $|M|=3$ for \textsc{Seesaw} and \textsc{Up-N-Down}.
}
\label{fig:karel_program_examples_seesaw_upNdown}
\end{figure*}

\begin{figure*}[ht]
\centering
\begin{mdframed}[frametitle=Karel Programs]
\begin{subfigure}[t]{0.90\textwidth}

\centering
\textbf{\textsc{Farmer}}

{
\begin{subfigure}[t]{0.32\textwidth}
Mode 1
\begin{lstlisting}
DEF run m( 
    pickMarker 
    REPEAT R=0 r( 
        pickMarker 
        move 
        turnRight 
        move 
        move 
        turnLeft 
        move 
        move 
        turnRight 
        move 
        turnLeft 
        move 
        move 
        pickMarker 
        move 
        turnLeft 
        move 
        move 
        pickMarker 
        move 
        turnLeft 
        move 
        move 
        pickMarker 
        move 
        r) 
    m)


\end{lstlisting}
\end{subfigure}
\begin{subfigure}[t]{0.32\textwidth}
Mode 2
\begin{lstlisting}
DEF run m( 
    turnRight 
    move 
    putMarker 
    turnRight 
    move 
    pickMarker 
    putMarker 
    move 
    pickMarker 
    putMarker 
    move 
    pickMarker 
    putMarker 
    move 
    pickMarker 
    putMarker 
    move 
    pickMarker 
    putMarker 
    move 
    m)

\end{lstlisting}
\end{subfigure}
\begin{subfigure}[t]{0.32\textwidth}
Mode 3
\begin{lstlisting}
DEF run m( 
    turnRight 
    move 
    turnRight 
    putMarker 
    pickMarker 
    move 
    pickMarker 
    putMarker 
    move 
    pickMarker 
    putMarker 
    move 
    pickMarker 
    putMarker 
    move 
    pickMarker 
    putMarker 
    move 
    pickMarker 
    putMarker 
    move 
    pickMarker 
    m)

\end{lstlisting}
\end{subfigure}
}

{
\begin{subfigure}[t]{0.32\textwidth}
Mode 4
\begin{lstlisting}
DEF run m( 
    REPEAT R=17 r( 
        IF c( not c( rightIsClear c) c) i( 
            putMarker 
            move 
            i) 
        r) 
    m)
\end{lstlisting}
\end{subfigure}
\begin{subfigure}[t]{0.32\textwidth}
Mode 5
\begin{lstlisting}
DEF run m( 
    putMarker 
    pickMarker 
    turnRight 
    move 
    pickMarker 
    turnRight 
    move 
    pickMarker 
    putMarker 
    move 
    pickMarker 
    putMarker 
    move 
    pickMarker 
    putMarker 
    move 
    pickMarker 
    putMarker 
    move 
    pickMarker 
    putMarker 
    move 
    m)
\end{lstlisting}
\end{subfigure}

}
\end{subfigure}

\end{mdframed}
\caption[\method\\ Karel Tasks Synthesized Programs]{\textbf{Example programs on Karel-Long tasks: \textsc{Farmer}.} The programs with best rewards out of all random seeds are shown. $|M|=5$ for \textsc{Farmer}.
}
\label{fig:karel_program_examples_farmer}
\end{figure*}

\begin{figure*}[ht]
\centering
\begin{mdframed}[frametitle=Karel Programs]

\begin{subfigure}[t]{0.90\textwidth}

\centering
\textbf{\textsc{Inf-Doorkey}}

{
\begin{subfigure}[t]{0.32\textwidth}
Mode 1
\begin{lstlisting}
DEF run m( 
    IFELSE c( noMarkersPresent c) i( 
        move 
        putMarker 
        i) 
    ELSE e( 
        pickMarker 
        putMarker 
        e) 
    m)

\end{lstlisting}
\end{subfigure}
\begin{subfigure}[t]{0.32\textwidth}
Mode 2
\begin{lstlisting}
DEF run m( 
    IF c( noMarkersPresent c) i( 
        move 
        move 
        move 
        IF c( noMarkersPresent c) i( 
            move 
            REPEAT R=1 r( 
                move 
                turnLeft 
                r) 
            i) 
        i) 
    IF c( noMarkersPresent c) i( 
        move 
        REPEAT R=1 r( 
            move 
            REPEAT R=1 r( 
                move 
                r) 
            r) 
        i) 
    m)
\end{lstlisting}
\end{subfigure}
\begin{subfigure}[t]{0.32\textwidth}
Mode 3
\begin{lstlisting}
DEF run m( 
    putMarker 
    pickMarker 
    IF c( not c( markersPresent c) c) i( 
        move 
        i) 
    WHILE c( noMarkersPresent c) w( 
        IFELSE c( markersPresent c) i( 
            turnLeft 
            i) 
        ELSE e( 
            move 
            turnRight 
            e) 
        w) 
    m)

\end{lstlisting}
\end{subfigure}
}

{
\begin{subfigure}[t]{0.32\textwidth}
Mode 4
\begin{lstlisting}
DEF run m( 
    putMarker 
    turnLeft 
    pickMarker 
    move 
    turnLeft 
    IFELSE c( markersPresent c) i( 
        move 
        turnLeft 
        i) 
    ELSE e( 
        move 
        e) 
    m)
\end{lstlisting}
\end{subfigure}
\begin{subfigure}[t]{0.32\textwidth}
Mode 5
\begin{lstlisting}
DEF run m( 
    turnRight 
    IF c( leftIsClear c) i( 
        turnLeft 
        i) 
    putMarker 
    IF c( leftIsClear c) i( 
        turnLeft 
        i) 
    pickMarker 
    move 
    IF c( leftIsClear c) i( 
        turnLeft 
        i) 
    putMarker 
    move 
    IF c( leftIsClear c) i( 
        turnLeft 
        i) 
    pickMarker 
    move 
    m)
\end{lstlisting}
\end{subfigure}

}

\end{subfigure}


\end{mdframed}
\caption[\method\\ Karel Tasks Synthesized Programs]{\textbf{Example programs on Karel-Long tasks: \textsc{Inf-Doorkey}.} The programs with best rewards out of all random seeds are shown. $|M|=5$ for \textsc{Inf-Doorkey}.
}
\label{fig:karel_program_examples_infDoorKey}
\end{figure*}

\begin{figure*}[ht]
\centering
\begin{mdframed}[frametitle=Karel Programs]
\begin{subfigure}[t]{0.90\textwidth}

\centering
\textbf{\textsc{Inf-Harvester}}

{
\begin{subfigure}[t]{0.32\textwidth}
Mode 1
\begin{lstlisting}
DEF run m( 
    REPEAT R=7 r( 
        turnLeft 
        move 
        REPEAT R=8 r( 
            turnLeft 
            WHILE c( frontIsClear c) w( 
                pickMarker 
                move 
                w) 
            r) 
        turnLeft 
        REPEAT R=8 r( 
            pickMarker 
            r) 
        r) 
    m)

\end{lstlisting}
\end{subfigure}
\begin{subfigure}[t]{0.32\textwidth}
Mode 2
\begin{lstlisting}
DEF run m( 
    WHILE c( frontIsClear c) w( 
        IF c( frontIsClear c) i( 
            IF c( frontIsClear c) i( 
                pickMarker 
                move 
                i) 
            IF c( frontIsClear c) i( 
                pickMarker 
                i) 
            i) 
        w) 
    IF c( frontIsClear c) i( 
        pickMarker 
        i) 
    pickMarker 
    m)
\end{lstlisting}
\end{subfigure}
\begin{subfigure}[t]{0.32\textwidth}
Mode 3
\begin{lstlisting}
DEF run m( 
    REPEAT R=5 r( 
        turnLeft 
        move 
        pickMarker 
        move 
        turnLeft 
        pickMarker 
        move 
        pickMarker 
        move 
        pickMarker 
        move 
        pickMarker 
        move 
        pickMarker 
        move 
        pickMarker 
        move 
        pickMarker 
        move 
        pickMarker 
        move 
        pickMarker 
        move 
        pickMarker 
        move 
        pickMarker 
        move 
        pickMarker
        move 
        pickMarker 
        move 
        pickMarker 
        move 
        pickMarker 
        r) 
    m)

\end{lstlisting}
\end{subfigure}
}


\end{subfigure}




\end{mdframed}
\caption[\method\\ Karel Tasks Synthesized Programs]{\textbf{Example programs on Karel-Long tasks: \textsc{Inf-Harvester}.} The programs with best rewards out of all random seeds are shown. $|M|=3$ for \textsc{Inf-Harvester}.
}
\label{fig:karel_program_examples_harvester}
\end{figure*}

\section{Designing Domain-Specific Languages}
Our program policies are designed to describe high-level task-solving procedures or decision-making logics of an agent. Therefore, our principle of designing domain-specific languages (DSLs) considers a general setting where an agent can perceive and interact with the environment to fulfill some tasks. DSLs consist of control flows, perceptions, and actions. While control flows are domain-independent, perceptions and actions can be designed based on the domain of interest, requiring specific expertise and domain knowledge. 

Such DSLs are proposed and utilized in various domains, including ViZDoom~\citep{kempka2016vizdoom}, 2D MineCraft~\citep{andreas2016modular, sun2020program}, and gym-minigrid~\citep{chevalier2023minigrid}. Recent works~\citep{liang2023code, wang2023demo2code} also explore describing agents’ behaviors using programs with functions taking arguments.


\end{document}